\documentclass[12pt,oneside,reqno,a4paper]{article}

\usepackage[utf8]{inputenc}
\usepackage{amsmath,amsthm,amssymb}     

\usepackage{lmodern}                    
\usepackage[T1]{fontenc}                
\usepackage{microtype}                  

\usepackage{typearea}                   
   \areaset[0mm]{135mm}{210mm}          
\usepackage{graphicx}
\usepackage{url}

\usepackage{ifpdf}
\ifpdf
  \pdfoutput=1
  \usepackage[pdfusetitle,colorlinks=true,linktoc=all]{hyperref}
\else
  \usepackage[dvips,ps2pdf,linktoc=all]{hyperref}
\fi

\ifdefined\hypersetup
  \hypersetup{
    pdfkeywords={}, linkcolor=black, citecolor=black, filecolor=black, urlcolor=black,
  }
\fi

\usepackage{fancyhdr}
\usepackage[margin=2cm]{geometry}
\usepackage{shapepar,tabularx}
\usepackage{longtable}
\usepackage{multicol}
\usepackage{multirow}
\usepackage{subcaption}                 
\usepackage{aligned-overset}

\usepackage{mathtools}                  
\usepackage{braket}
\usepackage{latexsym,amscd}

\usepackage{enumerate}
\usepackage{paralist}

\usepackage{tikz,pgfplots, pgfmath}
\usepackage{tikz-cd}
\usepackage{tikz-3dplot}
\usetikzlibrary{arrows.meta,calc,decorations.markings,math,cd,shadings}

\usepackage[all,2cell, color]{xy} 
\UseAllTwocells 
\SilentMatrices

\usepackage[capitalise]{cleveref}
\usepackage[backend=biber]{biblatex}
\usepackage{imakeidx}                   

\usepackage{color,epic}
\usepackage{pifont}
\usepackage{verbatim}
\usepackage{epsdice}
\usepackage{blindtext}                  

\makeindex
\newcommand{\defind}[1]{{\bf{#1}\index{#1}}}

\newcounter{Def}[subsection]

\newtheoremstyle{upright}
{1.5em}
{1.5em}
{\upshape} 
{}
{\bfseries}
{:}
{1em}
{}
\theoremstyle{upright}

\newtheorem{Theorem}[Def]{Theorem}
\newtheorem{Example}[Def]{Example}
\newtheorem{Remark}[Def]{Remark}
\newtheorem{Corollary}[Def]{Corollary}

\newtheorem{Definition}[Def]{Definition}
\newtheorem{Lemma}[Def]{Lemma}
\newtheorem{Proposition}[Def]{Proposition}
\newtheorem{Notation}[Def]{Notation}

\newenvironment{herefigure}[1][]{
  \begin{center}
    \begin{minipage}{\textwidth}
      \captionsetup{type=figure} 
    }{
    \end{minipage}
  \end{center}
}

\newenvironment{Proof}{\paragraph{Proof:}}{\hfill$\square$}

\newcommand{\curry}{\mathit{curry}}
\newcommand{\uncurry}{\mathit{uncurry}}

\newcommand{\Id}[1]{\mathit{Id}_{#1}}
\newcommand{\id}[1]{\mathit{id}_{#1}}

\newcommand{\Hom}[3]{\mathit{Hom}_{#1}\left(#2,#3\right)}
\newcommand{\Ob}[1]{\mathit{Obj}{#1}}

\newcommand{\op}{\mathit{op}}

\newcommand{\cocone}[2]{\mathit{cocone}\left(#1,#2\right)}
\newcommand{\colim}[1]{\mathit{colim}\left(#1\right)}
\newcommand{\colimNotScaling}[1]{\mathit{colim}(#1)}
\newcommand{\coeq}[1]{\mathit{coeq}\left(#1\right)}

\newcommand{\Lan}[2]{\mathit{Lan}_{#1}#2}

\newcommand{\Sheaf}[1]{\mathit{Shv}\left({#1}\right)}
\newcommand{\SheafSt}[1]{\mathit{Shv^*}\left({#1}\right)}
\newcommand{\SheafFunctor}{\mathit{Shv}}

\newcommand{\Val}[1]{\mathit{Val}\left({#1}\right)}
\newcommand{\SVal}[1]{\mathit{USVal}\left({#1}\right)}

\newcommand{\CVal}[1]{\mathit{CVal}\left({#1}\right)}

\newcommand{\Fuz}{\mathit{Fuz}}
\newcommand{\CFuz}{\mathit{CFuz}}
\newcommand{\SFuz}{\mathit{USFuz}}

\newcommand{\C}{\mathit{C}}
\newcommand{\M}{\mathit{M}}

\newcommand{\SC}{\mathit{SC}}
\newcommand{\SM}{\mathit{SM}}

\newcommand{\NSet}{\mathit{NSet}}
\newcommand{\CNSet}{\mathit{CNSet}}
\newcommand{\SNSet}{\mathit{USNSet}}

\newcommand{\Met}{\mathit{Met}}
\newcommand{\EPMet}{\mathit{EPMet}}

\newcommand{\Top}{\mathit{Top}}
\newcommand{\SET}{\mathit{Set}}
\newcommand{\Lip}{\mathit{Lip}}
\newcommand{\Loc}{\mathit{Loc}}
\newcommand{\Cat}{\mathit{Cat}}

\newcommand{\TopRe}{\mathit{Geom}}

\newcommand{\MetRe}{\mathit{MetRe}}
\newcommand{\Sing}{\mathit{Sing}}
\newcommand{\y}{\mathit{y}}

\newcommand{\CMetRe}{\mathit{CMetRe}}
\newcommand{\CSing}{\mathit{CSing}}
\newcommand{\Cy}{\mathit{Cy}}

\newcommand{\Fin}[1]{\mathit{Fin-}{#1}}

\newcommand{\norm}[1]{\left\lVert#1\right\rVert}

\newcommand{\pathlen}[1]{\lvert #1 \rvert}

\newcommand{\numberthis}[1]{%
  \refstepcounter{equation}%
  \tag{\theequation}\label{#1}%
}

\usepackage[capitalise]{cleveref}
\usepackage[backend=biber]{biblatex}
\crefname{Def}{Definition}{Definitions}
\Crefname{Def}{Definition}{Definitions}

\crefalias{Theorem}{Theorem}
\crefalias{Example}{Example}
\crefalias{Remark}{Remark}
\crefalias{Corollary}{Corollary}
\crefalias{Application}{Application}
\crefalias{Definition}{Def}
\crefalias{Lemma}{Lemma}
\crefalias{Proposition}{Proposition}
\crefalias{Notation}{Notation}

\crefname{Theorem}{Theorem}{Theorems}
\Crefname{Theorem}{Theorem}{Theorems}
\crefname{Example}{Example}{Examples}
\Crefname{Example}{Example}{Examples}
\crefname{Remark}{Remark}{Remarks}
\Crefname{Remark}{Remark}{Remarks}
\crefname{Corollary}{Corollary}{Corollaries}
\Crefname{Corollary}{Corollary}{Corollaries}
\crefname{Application}{Application}{Applications}
\Crefname{Application}{Application}{Applications}
\crefname{Lemma}{Lemma}{Lemmas}
\Crefname{Lemma}{Lemma}{Lemmas}
\crefname{Proposition}{Proposition}{Propositions}
\Crefname{Proposition}{Proposition}{Propositions}
\crefname{Notation}{Notation}{Notations}
\Crefname{Notation}{Notation}{Notations}

\usepackage{fancyhdr}
\title{The Theory behind UMAP?}
\author{David Wegmann\\
    University of Erlangen-Nuremberg\\
    \href{mailto:david.wegmann@fau.de}{david.wegmann@fau.de}}

\begin{document}

\maketitle

\pagenumbering{Alph}
\renewcommand{\thepage}{C-\Roman{page}}

\pagenumbering{Roman}

\begin{abstract}
In 2018, McInnes et al. introduced a dimensionality reduction algorithm called UMAP in \cite{UMAP}, which enjoys wide popularity among data scientists. The article \cite{UMAP} introduces a finite variant of a functor called the metric realization from an unpublished draft \cite{Spivak2009METRICRO} by Spivak. This draft \cite{Spivak2009METRICRO} contains many errors, most of which are reproduced by McInnes et al. \cite{UMAP} and subsequent publications. This thesis aims to repair these errors and provide a self-contained document with the full derivation of Spivak's functors and McInnes et al.'s finite variant. We contribute an explicit description of the metric realization and related functors. At the end of this thesis, we discuss the UMAP algorithm, as well as claims about properties of the algorithm and the correspondence of McInnes et al.'s finite variant to the UMAP algorithm.
\end{abstract}

\section{Introduction}

In 2018, McInnes et al. introduced a dimensionality reduction algorithm called UMAP \cite{UMAP}, which enjoys wide popularity among data scientists. As of August 28, 2025, Google Scholar reports 19,058 citations of \cite{UMAP}. In their article, McInnes et al. include a theory section that aims to justify this algorithm. It attempts to construct a finite variant of a functor called the metric realization, which first appeared in the unpublished draft \cite{Spivak2009METRICRO} by Spivak. This draft in turn uses a sheaf-theoretic perspective of fuzzy sets from Barr \cite{BarrFuzzy}. Spivak's draft \cite{Spivak2009METRICRO} contains several minor mistakes and gaps, but most results seem to work out if one tries to fill in the missing pieces.

Spivak never published the draft and added disclaimers warning of potential errors. He only put the draft on his website (with another disclaimer about errors) after the article by McInnes et al. \cite{UMAP} received attention. This article \cite{UMAP} also cites Barr \cite{BarrFuzzy}, reproduces many of Spivak's errors, and incorrectly redefines concepts that were correctly defined by Barr \cite{BarrFuzzy} in 1986. Some examples of mistakes and gaps in \cite{UMAP} and \cite{Spivak2009METRICRO} are listed in \cref{Sec:Issues}.

The goal of this thesis is to resolve these issues and provide a self-contained document which contains an explicit construction of Spivak's metric realization, McInnes et al.'s finite metric realization, as well as all non-standard structures that this construction depends on.

We begin by summarizing the necessary categorical background, as well as adapting some established results to our applications in \cref{Sec:Preliminaries}.

We then summarize all the relevant definitions and results for fuzzy sets from Barr in \cref{Sec:ValuedSets} and repair the associated issues in \cite{UMAP} and \cite{Spivak2009METRICRO} described in \cref{Sec:Issues}, \cref{Issue1} and \ref{Issue2}. We also contribute a more explicit alternative proof for the main result of Barr \cite{BarrFuzzy}, as explained at the beginning of \cref{Sec:ValuedSets}.

In \cref{Sec:EPMetAndMetRe}, we summarize the target category of the metric realization from \cite{Spivak2009METRICRO}. We then derive the metric realization similarly to Spivak \cite{Spivak2009METRICRO} while repairing the associated issues described in \cref{Sec:Issues}, \cref{IssueDefFunctor} and \ref{IssueKan}. Our main contribution is the explicit description of the metric realization and associated functors in \cref{Subsection:ClassicalMetricRealization}, as explained in detail at the beginning of \cref{Sec:EPMetAndMetRe}. These results rely on the other contribution in \cref{Sec:ValuedSets}.

In section \cref{Sec:FinMetRe}, we provide a construction of McInnes et al.'s finite metric realization, but this requires, in particular, a precise interpretation of the vague definitions described in \cref{Sec:Issues}, \cref{IssuesFin}.

Finally, we discuss the main ideas behind the UMAP algorithm from \cite{UMAP} in \cref{Sec:UMAP}. We comment on claims about properties of the algorithm and how they relate to our work in \cref{Subsection:Discussion}.

\subsection{Issues in McInnes et al. \cite{UMAP} and Spivak \cite{Spivak2009METRICRO}} \label{Sec:Issues}

The metric realization functor $\text{Real} : \text{sFuzz} \rightarrow \text{EPMet}$ that appears in \cite{UMAP} is a generalization of the geometric realization functor $\TopRe : \SET^{\Delta^\op} \rightarrow \Top$. This functor was originally introduced in Spivak's unpublished draft \cite[Section~3]{Spivak2009METRICRO}.

Here, the category of simplicial sets $\SET^{\Delta^\op}$ is replaced by the category of simplicial fuzzy sets $\text{sFuzz}$ where each element has an associated membership degree in (0,1]. The category of topological spaces $\Top$ is replaced by the category of extended pseudo-metric spaces $\text{EPMet}$. Extended pseudo-metric spaces are similar to metric spaces, but allow the distance between different points to be 0 or $\infty$. The functor $\text{Real}$ then realizes a simplicial fuzzy set $S$ as an extended pseudo-metric space similar to the geometric realization. Additionally, the values of elements in $S$ influence the size of corresponding simplices.

McInnes et al. build upon Spivak's result by introducing a finite variant of this metric realization in \cite{UMAP}. However, McInnes et al. reproduce several of the issues present in Spivak's original draft \cite{Spivak2009METRICRO}. Below we identify examples of these issues that appear in \cite{UMAP}, noting when they originate from \cite{Spivak2009METRICRO}.

The notation for categories in Spivak's draft \cite{Spivak2009METRICRO} differs from that in McInnes et al. \cite{UMAP}, so we provide a translation table in the index of this thesis.

\begin{enumerate}
\item \label{Issue1} In the paragraphs before \cite[Definition~3]{UMAP}, Spivak and Barr \cite{BarrFuzzy} are cited, followed by an attempt at defining a topology on $I = (0,1]$ which is missing the empty set. This mirrors an error in Spivak's draft \cite[Section~1]{Spivak2009METRICRO}. In \cite[Definition~3]{UMAP}, fuzzy sets are defined to be presheaves on this supposed topological space $I$ such that all restrictions are injections. This is incorrect, as it does not lead to the desired equivalence of categories. The correct definition can be found in Barr \cite[Section~2]{BarrFuzzy}. In \cite[Definition~4]{UMAP}, the category of fuzzy sets, called Fuzz, is then suddenly defined as a category of sheaves, not of presheaves.

\item \label{Issue2} The article \cite{UMAP} then defines simplicial objects over this category. It provides an equivalent alternative definition that also mirrors an error in Spivak's draft \cite[Section~1]{Spivak2009METRICRO}. Definition~5 in \cite{UMAP} defines simplicial fuzzy sets as functors from $\Delta^\op$ to Fuzz, where $\Delta$ is the simplex category and Fuzz is the category of fuzzy sets discussed in issue \ref{Issue1}. This is just a special case of the general definition of simplicial objects in a category $\mathcal{C}$ and is correct. As Fuzz is defined as a subcategory of the category of functors from $I^\op$ to $\SET$, the simplicial objects in \cite[Definition~5]{UMAP} form a subcategory of the functor category $(\SET^{I})^{\Delta^\op}$.

  The next paragraph then attempts to identify this category with an equivalent subcategory of the functor category $\SET^{(\Delta \times I)^\op}$. This is commonly referred to as {\it currying} by computer scientists, see \cref{Prelim:Currying}. However, to obtain an equivalence of categories here, the additional conditions on functors in Fuzz must be properly translated. The sheaf condition is correctly translated by endowing $\Delta$ with the trivial Grothendieck topology and then considering only sheaves on the product Grothendieck topology $\Delta \times I$.

  If Fuzz were defined as the category of all sheaves on $I$, this would indeed be an equivalent category. But Fuzz also requires that most restriction maps are injective, as discussed in issue \ref{Issue1}. This condition is missing throughout \cite{UMAP}, but it can be restored by considering the subcategory of sheaves $F : \Delta \times I \rightarrow \SET$ such that all restrictions $F(id_{[n]},-)$ are injective.

  However, this is still cumbersome as it requires dealing with a product Grothendieck topology. It is much easier to forgo Grothendieck topologies entirely and simply define this category as functors $F : \Delta \times I \rightarrow \SET$ such that $F([n],-) : I \rightarrow \SET$ is a fuzzy set as in \cref{Def:SimpValSet}.
  
\item \label{IssueDefFunctor} In the paragraphs leading up to the derivation of the metric realization \cite[Definition~7]{UMAP}, a functor called Real is defined. This definition contains three major unrelated errors.
  \begin{enumerate}
  \item \label{IssueIntervalIncluding0} They involve logarithms of parameters $a$ and $b$ where $0 \le a \le b \le 1$ but including 0 breaks this definition as $\log(0)$ is undefined. Excluding 0 leads to further issues: as $I$ is supposed to be a topology, the empty set must map to something, but the empty set is not of the form $[0,a)$ for $a > 0$. This problem propagates in the literature. Barth et al. \cite{barth} attempt to clear up the theory behind UMAP and explicitly include 0 in the definition of $I$ \cite[Remark~3.1]{barth} but later define the metric realization in the same way. Assuming $\log(0) = -\infty$ does not address this problem, as the metric simplices of size 1 are then defined as the empty set but the action map collapses the input simplex into a point since we scale by a factor divided by infinity. That article also uses an incorrect definition for sheaf-theoretic fuzzy sets \cite[Definition~3.1]{barth}. If the definition requires all restrictions to be injective, fuzzy sets cannot have more than one element, see \cref{Rem:SheafEmptyCover}.
  \item Independent of issue 3(a), the scaling factor $\frac{\log(b)}{\log(a)}$ may result in division by 0, because $a$ can take the value 1. We repair this by defining all metric simplices with the same underlying set and instead scaling the metric by a factor in \cref{Def:PseudoMetricSimplicies}.
    
  \item Independent of issue 3(a) and 3(b), the action of Real on morphisms is required to result in non-expansive maps, but the article \cite{UMAP} fails to show this.
    
    First, the article \cite{UMAP} fails to clarify which metric is used in this context, the article states ``The metric on Real($\Delta^n_{<a}$) is simply inherited from $\mathbb{R}^{n+1}$'', but does not specify a metric on $\mathbb{R}^{n+1}$. However, the article \cite{UMAP} cites Spivak's unpublished draft which uses the Euclidean metric \cite[Section~3]{Spivak2009METRICRO}.

    The article \cite{UMAP} then states that their definition is ``clearly non-expansive'' with a reference to the scaling factor $\frac{\log{b}}{\log{a}}$ which is smaller than or equal to one for $a \le b$. However, if one evaluates the action on a non-scaling degeneracy, the action is clearly not non-expansive, as we show in \cref{Rem:L1}.

    This error first appears in Spivak's unpublished draft \cite[Section~3]{Spivak2009METRICRO} and is, apart from \cite{UMAP}, also reproduced by Barth et al. \cite[Prop 5.1 (28)]{barth}. Barth et al. denote this functor as Re \cite[Prop 5.1 (28)]{barth} and cite Spivak, but do not mention which metric they use in this context.

    To resolve this, we equip the simplices with the $\ell_1$ metric in \cref{Def:PseudoMetricSimplicies}. This is the only $\ell_p$ metric that makes this definition non-expansive, as shown in \cref{Rem:L1}.
  \end{enumerate}
\item \label{IssueKan} The article \cite{UMAP} constructs both Spivak's metric realization and his finite metric realization as a left Kan extension along the Yoneda embedding. This Yoneda embedding is supposed to map from $\Delta \times I$ to sFuzz or Fin-sFuzz in the finite case, but functors in the image of the Yoneda embedding are not sheaves as demanded in \cite[Definition~4]{UMAP}. Again due to \cref{Rem:SheafEmptyCover}: $y([n],0)$ must be a singleton set. This is not guaranteed as the involved hom sets in $\Delta$ may have more than one element. There is also no proof that the objects in the image of the respective nerves of both functors are sheaves. This is easy to see for the finite metric realization, but more difficult for Spivak's metric realization. We repair this in \cref{Cor:SimplicialKanExt} for Spivak's metric realization.
  
\item \label{IssuesFin} To define a finite variant of Spivak's metric realization, McInnes et al. \cite{UMAP} introduce finite variants of the categories that appear in Spivak's metric realization. These definitions leave room for interpretation.
  \begin{enumerate}
  \item In the paragraph just after \cite[Definition~6]{UMAP}, the article states ``We denote the subcategory of finite extended-pseudo-metric spaces FinEPMet.'' but finiteness is not specified further. We interpret this as demanding that the underlying set has finitely many elements in \cref{Def:FinEpMet}.
  \item In the paragraph just before \cite[Definition~7]{UMAP}, the article states ``For our case we are only interested in finite metric spaces. To correspond with this we consider the subcategory of bounded fuzzy simplicial
    sets Fin-sFuzz.''. Again, boundedness is not specified further. Here, it is much more difficult to come up with a working interpretation, because an object in sFuzz is a functor that maps a pair of a natural number and a real number in the interval $(0,1]$ to a set. Our interpretation of this definition can be found in \cref{Def:FinSFuz}.
  \end{enumerate}
  Both of our interpretations have been chosen, because they seem consistent with McInnes et al.'s results.

\end{enumerate}
In addition to the already listed issues with \cite{UMAP}, there is a claim in Spivak's draft \cite{Spivak2009METRICRO} worth criticizing. This claim does not appear in \cite{UMAP} and is not used to derive any other results. If it turns out to be false, nothing else would break.
\begin{enumerate}
\item In Spivak's draft \cite{Spivak2009METRICRO}, the category Fuz denotes the category discussed in issue \ref{Issue1}, a subcategory of the category $\Sheaf{I}$ that contains only the sheaves where certain restrictions are injective. In \cite[Lemma 1.3]{Spivak2009METRICRO} Spivak attempts to construct a left adjoint $m : \Sheaf{I} \rightarrow \text{Fuz}$ to the forgetful functor $U : \text{Fuz} \rightarrow \Sheaf{I}$ by simply identifying the elements that violate the additional injectivity condition. For $m$ to be a well-defined functor, it must be shown that
  \begin{enumerate}
  \item $m$ satisfies the functor laws,
  \item most restriction maps of every functor $mS$ are injective, even if $S$ is just a regular sheaf,
  \item when we send a regular sheaf $S$ through $m$, then $mS$ must still be a sheaf.
  \end{enumerate}
  It is easy to show (a) and (b), but when trying to prove (c), one should run into an issue: here we must construct a unique element $x \in mS(a)$ subject to further conditions by making use of the original sheaf condition in $S$. But the way $mS$ is defined only lets us apply the sheaf condition of $S$ to obtain such an element $x \in S(b)$ at a lower level $b \le a$. It then seems impossible to lift this element up to the desired level $mS(a)$. This is, of course, no counterexample. However, it is a gap in the argument, and Spivak does not offer a proof in \cite{Spivak2009METRICRO}.
\end{enumerate}


\clearpage
\tableofcontents

\clearpage
\pagenumbering{arabic}
\clearpage

\clearpage
\section{Categorical Background} \label{Sec:Preliminaries}

In this section, we will summarize established results and definitions from the literature or introduce slightly modified results that are tailored to our applications.

We presuppose that the reader understands category theory up to colimits and Kan extensions. An introduction to these concepts can, for instance, be found in Riehl's textbook \citetitle{riehl2017category} \cite{riehl2017category}.

We also expect the reader to be familiar with simplicial objects, but we will briefly introduce them in \cref{Sec:Simplex}. We recommend Richter's textbook \citetitle{Richter_2020} \cite[Chapter~10]{Richter_2020} as a reference for simplicial objects.

We will briefly use coends as an intermediate step for the proof of \cref{Lem:CoequalizerFormula}, but it is not required to be familiar with coends to understand everything else in this thesis. For a textbook reference for coends, see Theorem 1.2.1 (1.2.3) in Riehl's other textbook \citetitle{Riehl_2014} \cite{Riehl_2014}.

The metric realization mentioned in the introduction has a well-known topological counterpart, called the geometric realization. It is easier to understand than the metric realization, so we provide the construction of the geometric realization as a left Kan extension in \cref{Subsection:GeometricRealization}. The variants of the metric realization introduced in \cref{Sec:EPMetAndMetRe} will be constructed in a similar fashion.

Some textbooks define the geometric realization directly using quotients, like \cite[10.6.1]{Richter_2020}. We recommend that readers who are familiar with the geometric realization but not as a Kan extension read \cref{Subsection:GeometricRealization} before \cref{Sec:EPMetAndMetRe}. For a textbook reference that also defines the geometric realization as a left Kan extension and then derives the description as a coequalizer, see the special case \cite[Example 1.5.3]{Riehl_2014} of \cite[Construction 1.5.1]{Riehl_2014}.

We will also need some basic sheaf theory to express classical fuzzy sets as sheaves, but we do not expect the reader to be familiar with these concepts. We provide all the necessary definitions and results in \cref{Sec:Locales} and \cref{Sec:SheavesOnLocales}.

\subsection{Currying of Functors} \label{Prelim:Currying}

As mentioned in the introduction, we will require an equivalence referred to as {\it currying} by computer scientists. In computer science, this usually refers to the isomorphism
$$C^{A\times B} \cong {C^B}^A,$$
where $A,B,C$ are sets and $Y^X$ denotes the set of maps from $X$ to $Y$.

This isomorphism has a categorical counterpart for categories $\mathcal{A}, \mathcal{B}, \mathcal{C}$ where $\mathcal{Y}^\mathcal{X}$ denotes the category of functors from $\mathcal{X}$ to $\mathcal{Y}$. We can also restrict $\mathcal{C}^\mathcal{B}$ to a subcategory if we impose an additional condition.
This version is specifically tailored to be used later in \cref{Thm:EquivCValSVal}.

\begin{Lemma}\label{Lem:Curry} Let $\mathcal{A}, \mathcal{B}, \mathcal{C}$ be categories and let $\mathcal{S}$ be a full subcategory of $\mathcal{C}^\mathcal{B}$.\smallbreak
  Then there is an isomorphism of categories
  $$\mathcal{T} \cong \mathcal{S}^\mathcal{A}, $$
  where $\mathcal{T}$ is the full subcategory of $\mathcal{C}^{\mathcal{A} \times \mathcal{B}}$ that has as objects all functors $F : \mathcal{A} \times \mathcal{B} \rightarrow \mathcal{C}$ such that $F(A,-) : \mathcal{B} \rightarrow \mathcal{C}$ is an object in $\mathcal{S}$ for any object $A$ in $\mathcal{A}$.
  \begin{Proof}
    We define the functors $\curry : \mathcal{T} \rightarrow \mathcal{S}^\mathcal{A}$
    \begin{align*}
       &\curry(F : \mathcal{A} \times \mathcal{B} \rightarrow \mathcal{C})(A) = F(A,-)\\
       &\curry(\alpha : F \Rightarrow G)_A = \alpha_{A,-},
    \end{align*}
    and $\uncurry : \mathcal{S}^\mathcal{A} \rightarrow \mathcal{T}$
    \begin{align*}
      &\uncurry(F : \mathcal{A} \rightarrow \mathcal{C}^\mathcal{B})(A,B) = F(A)(B)\\
      &\uncurry(F : \mathcal{A} \rightarrow \mathcal{C}^\mathcal{B})(a : A \rightarrow A',b : B \rightarrow B') = F(a)_{B'} \circ F(A)(b)\\
      &\uncurry(\alpha : F \Rightarrow G)_{A,B} = (\alpha_A)_B.
    \end{align*}
    We verify that $\uncurry \circ \curry = \Id{\mathcal{T}}$:
    \begin{align*}
      \uncurry(\curry(F))(A,B) &= \curry(F)(A)(B) = F(A,-)(B) = F(A,B)\\
      \uncurry(\curry(\alpha))_{A,B} &= (\curry(\alpha)_A)_B = \alpha_{A,B}.
    \end{align*}
    And that $\curry \circ \uncurry = \Id{\mathcal{S}^\mathcal{A}}$:
    \begin{align*}
       \curry(\uncurry(F))(A)(B) &= \uncurry(F)(A,-)(B) = \uncurry(F)(A,B) = F(A)(B)\\
       \curry(\uncurry(F))(A)(b) &= \uncurry(F)(A,-)(b)\\
                                 &= \uncurry(F)(\id{A},b) = F(\id{A})_{B'} \circ F(A)(b) = F(A)(b)\\
      (\curry(\uncurry(\alpha))_A)_B &= \uncurry(\alpha)_{A,B}= (\alpha_A)_B.
    \end{align*}
  \end{Proof}
\end{Lemma}

\subsection{Colimits and Full Subcategories}

We will need a concrete definition of cocones for the proof of \cref{Prop:KanExtYonedaSubcat}, so we introduce our preferred definition of colimits here. This specific variant of colimits is not required to understand the rest of the thesis.

We will also provide a lemma to compute colimits in a full subcategory within the larger category. This lemma will later be used to establish the existence of McInnes et al.'s finite metric realization in \cref{Def:FinMetRe}.

\begin{Definition} \cite[Definitions~3.1.1--3.1.3]{riehl2017category} Let $\mathcal{J}$ be small and let $\mathcal{C}$ be locally small.
  \begin{enumerate}
  \item For an object $C$ in $\mathcal{C}$, the \defind{constant functor} $\Delta(C) : \mathcal{J} \rightarrow \mathcal{C}$ sends
    \begin{itemize}
    \item all objects $J$ to $C$,
    \item all morphisms $j : J \rightarrow J'$ to $\id{C}$.
    \end{itemize}
  \item A \defind{cocone} under the diagram $F : \mathcal{J} \rightarrow \mathcal{C}$ with nadir $C$ is a natural transformation $\lambda : F \Rightarrow \Delta(C)$.
  \item Cocones under a diagram $F : \mathcal{J} \rightarrow \mathcal{C}$ define a functor $\cocone{F}{-} : \mathcal{C} \rightarrow \SET$ that sends
    \begin{itemize}
    \item an object $C$ in $\mathcal{C}$ to the set $\cocone{F}{C}$ of cocones under $F$ with nadir $C$,
    \item a morphism $f : C \rightarrow C'$ to the map $\cocone{F}{f} : \cocone{F}{C} \rightarrow \cocone{F}{C'}$ that sends a cocone $\lambda : F \Rightarrow \Delta(C)$ to the cocone $\Delta(f) \circ \lambda : F \Rightarrow \Delta(C')$.
    \end{itemize}
    A \defind{colimit} is a representing object of this functor: $$\cocone{F}{-} \cong \Hom{\mathcal{C}}{\colim{F}}{-}.$$
  \end{enumerate}
\end{Definition}

\begin{Lemma} \label{Lem:ColimitSubcat} Let $\mathcal{J}$ be small and let $\mathcal{S}$ be a full subcategory of $\mathcal{C}$.

  Let $F : \mathcal{J} \rightarrow \mathcal{C}$ be a diagram where $F(J)$ is in $\mathcal{S}$ for every $J$ in $\mathcal{J}$. Then $F$ can also be regarded as a functor $F : \mathcal{J} \rightarrow \mathcal{S}$.

  If a colimit object $\colim{F : \mathcal{J} \rightarrow \mathcal{C}}$ of $F : \mathcal{J} \rightarrow \mathcal{C}$ formed in $\mathcal{C}$ is in $\mathcal{S}$, then $\colim{F : \mathcal{J} \rightarrow \mathcal{C}}$ is also a colimit of $F : \mathcal{J} \rightarrow \mathcal{S}$ formed in the full subcategory $\mathcal{S}$.
  \begin{Proof}
    Let $\colim{F : \mathcal{J} \rightarrow \mathcal{C}}$ be a colimit object of $F : \mathcal{J} \rightarrow \mathcal{C}$ formed in $\mathcal{C}$.

    Then $\colim{F}$ is a representing object for $\cocone{F}{-}$ and we have the natural isomorphism
    $$\cocone{F}{-} \cong \Hom{\mathcal{C}}{\colim{F}}{-} : \mathcal{C} \rightarrow \SET.$$
    We restrict this natural isomorphism indexed by $C$ to $\mathcal{S}$ and obtain the natural isomorphism
    $$\cocone{F}{-} \cong \Hom{\mathcal{S}}{\colim{F}}{-} : \mathcal{S} \rightarrow \SET.$$
    Thus, $\colim{F}$ is also a representing object of the cocone functor $\cocone{F}{-} : \mathcal{S} \rightarrow \SET$ and thus a colimit of $F : \mathcal{J} \rightarrow \mathcal{S}$.
  \end{Proof}
\end{Lemma}

\subsection{Left Kan Extensions Along the (Restricted) Yoneda Embedding} \label{Prelim:LeftKanExt}

It is well-known that the existence of Kan extensions can be shown via the colimit formula. In this formula, a specific variant of the comma category appears, so we introduce this variant here.

\begin{Definition} \label{Def:Comma} Let $K : \mathcal{C} \rightarrow \mathcal{D}$ be a functor and let $D$ be an object in $\mathcal{D}$.
  \begin{enumerate}
    \item The \defind{comma category} $K \downarrow D$ has
    \begin{itemize}
    \item pairs $(C,d)$ as objects where $C$ is an object in $\mathcal{C}$ and $d : K(C) \rightarrow D$,
    \item morphisms $c : C \rightarrow C'$ with $d'\circ K(c)=d$ as morphisms from $(C,d)$ to $(C',d')$.
      \begin{center}
        \begin{tikzcd}
          K(C) \arrow[rr, "K(c)"] \arrow[rd, "d"'] &                     & K(C') \arrow[ld, "d'"] \\
          &  D  &      
        \end{tikzcd}
      \end{center}
    \end{itemize}
  \item The \defind{projection functor} $P^D : K \downarrow D \rightarrow \mathcal{C}$ is given by
    \begin{align*}
      &P^D(C,d) = C\\
      &P^D(c)   = c.
    \end{align*}
  \end{enumerate}
\end{Definition}

A more general variant appears in \cite[Exercise~1.3.vi]{riehl2017category}. The variant in \cref{Def:Comma} is introduced in the paragraphs before the colimit formula \cite[Theorem~6.2.1]{riehl2017category}, which we will summarize next.

\begin{Theorem} \cite[Theorem~6.2.1]{riehl2017category} \label{Thm:ColimitFormula} Let $F : \mathcal{C} \rightarrow \mathcal{E}$ and $K : \mathcal{C} \rightarrow \mathcal{D}$ be functors. If the functor $FP^D : K \downarrow D \rightarrow \mathcal{E}$ has a colimit for every object $D$ in $\mathcal{D}$, then $F$ has a left Kan extension given by $$\Lan{K}{F}(D) = \colim{FP^D}.$$ 
\end{Theorem}

This formula is useful for multiple purposes. It also allows us to show that Kan extensions exist in cases where $\mathcal{E}$ is not necessarily cocomplete, if we can show that at least the colimits $\colimNotScaling{FP^D}$ exist. And of course, if $\mathcal{E}$ is cocomplete, Kan extensions always exist.

\begin{Corollary} \label{Cor:ColimitFormula} Let $K : \mathcal{C} \rightarrow \mathcal{D}$ and let $\mathcal{C}$ be small and $\mathcal{D}$ locally small. When $\mathcal{E}$ is cocomplete, left Kan extensions along $K$ exist for all functors $F : \mathcal{C} \rightarrow \mathcal{E}$.
\end{Corollary}

If we already know that a Kan extension $\Lan{K}{F}$ exists, we can also use \cref{Thm:ColimitFormula} to compute the action $\Lan{K}{F}(D)$ by computing the colimit $\colimNotScaling{FP^D}$. Important special cases of this are left Kan extensions along the Yoneda embedding.

\begin{Definition} \label{Def:Yoneda} \cite[Corollary~2.2.6]{riehl2017category} The covariant \defind{Yoneda embedding} $y : \mathcal{C} \rightarrow \SET^{\mathcal{C}^\op}$ is the functor defined by
  \begin{align*}
    &y(C) = \Hom{\mathcal{C}}{-}{C}\\
    &y(f) = (-) \circ f.
  \end{align*}
\end{Definition}

In the case of a Kan extension $\Lan{y}{F}$ of $F : \mathcal{C} \rightarrow \mathcal{E}$ along $y : \mathcal{C} \rightarrow \SET^{\mathcal{C}^\op}$, we can derive a formula to compute the colimits in \cref{Thm:ColimitFormula} in terms of coequalizers.

\begin{Lemma} \label{Lem:CoequalizerFormula} Let $\mathcal{C}$ be small, let $\mathcal{E}$ be cocomplete, and let $F : \mathcal{C} \rightarrow \mathcal{E}$. \smallbreak
  Then the colimits in \cref{Thm:ColimitFormula} for left Kan extensions $\Lan{y}{F}$ of $F$ along the Yoneda embedding $y : \mathcal{C} \rightarrow \SET^{\mathcal{C}^\op}$ are given on an object $D : \mathcal{C}^\op \rightarrow \SET$ by the coequalizer
  $$\Lan{y}{F} = \colim{FP^D} = \coeq{\;\coprod_{\substack{f : C \rightarrow C'\\s \in D(C)}} F(C') \quad \overset{\phi}{\underset{\psi}{\rightrightarrows}} \quad \coprod_{\substack{C \in \mathcal{C}\\s \in D(C)}} F(C)}.$$
  where $\phi$ is the unique morphism with $\phi \circ \iota_{f,s} = \iota_{C',D(f)(s)} \circ F(f)$ and $\psi$ is the unique morphism with $\psi \circ \iota_{f,s} = \iota_{C,s}$.
  \smallbreak
  Furthermore, the action of $\Lan{y}{F}$ on a morphism $\alpha : D \Rightarrow D'$ is given by
  \[ e : \Lan{y}{F}(D) \rightarrow \Lan{y}{F}(D') \quad \text{ with } \quad  e \circ \pi \circ \iota_{C,s} = \pi' \circ \iota_{C,\alpha_C(s)}. \]
  Here $\pi : \coprod_{\substack{C \in \mathcal{C}\\s \in D(C)}} F(C) \rightarrow \Lan{y}{F}(D)$ is the epimorphism into the coequalizer and $\iota_{C,s} : F(C) \rightarrow \coprod_{\substack{C \in \mathcal{C}\\s \in D(C)}} F(C)$ are the inclusion morphisms to the coproduct.
  \begin{Proof}
    By \cite[Theorem~1.2.6]{Riehl_2014} the colimit $\colim{FP^D}$ is the coend of the functor
    $$H = \Hom{\SET^{\mathcal{C}^\op}}{y(-)}{D} \cdot F(-) : \mathcal{C}^\op \times \mathcal{C} \rightarrow \mathcal{E}.$$
    By \cite[Theorem~1.2.1, (1.2.4)]{Riehl_2014}, we can express this coend as the coequalizer
    \[ \colim{FP^D}=\int^{C\in\mathcal{C}}H(C,C) = \coeq{\;\coprod_{f : C \rightarrow C'} H(C',C) \quad \overset{\phi_1}{\underset{\psi_1}{\rightrightarrows}} \quad \coprod_{C \in \mathcal{C}} H(C,C)}, \]
    
    where $\phi_1$ is the unique morphism with $\phi_1 \circ \iota_{f} \circ \iota_s = \iota_{C'} \circ \iota_s \circ F(f)$ and $\psi_1$ is the unique morphism with $\psi_1 \circ \iota_{f} \circ \iota_s = \iota_{C} \circ \iota_{F(f)(s)}$.
    
    By the Yoneda lemma, we have $$\Hom{\SET^{\mathcal{C}^\op}}{y(C)}{D} = \Hom{\SET^{\mathcal{C}^\op}}{\Hom{\mathcal{C}}{-}{C}}{D} \cong D(C)$$ for every object $C$ in $\mathcal{C}$ and hence
    $$H(C,C') = D(C) \cdot F(C') = \coprod_{D(C)}F(C').$$
    Thus we can rewrite the coequalizer as
    $$\coeq{\;\coprod_{\substack{f : C \rightarrow C'\\s \in D(C)}} F(C') \quad \overset{\phi}{\underset{\psi}{\rightrightarrows}} \quad \coprod_{\substack{C \in \mathcal{C}\\s \in D(C)}} F(C)}.$$
    For the action on a morphism $\alpha : D \Rightarrow D'$, we can use the universal property of the coequalizer $\Lan{y}{F}(D)$. For this, consider the diagram
    \begin{center}
      \begin{tikzcd}[column sep=large, row sep=large]
        \coprod_{\substack{f : C \rightarrow C'\\s \in D(C)}} F(C') \arrow[r, "\phi", shift left=1.5] \arrow[d, "k"] \arrow[r, "\psi"', shift right=1.5] & \coprod_{\substack{C \in \mathcal{C}\\s \in D(C)}} F(C) \arrow[r, "\pi"] \arrow[d, "h"] & \Lan{y}{F}(D) \arrow[d, "e", dashed] \\
        \coprod_{\substack{f : C \rightarrow C'\\s \in D'(C)}} F(C') \arrow[r, "\phi'", shift left=1.5] \arrow[r, "\psi'"', shift right=1.5]                   & \coprod_{\substack{C \in \mathcal{C}\\s \in D'(C)}} F(C) \arrow[r, "\pi'"]                        & \Lan{y}{F}(D')                       
      \end{tikzcd}
    \end{center}
    where $h$ is the unique morphism with $h \circ \iota_{C,s} = \iota_{C,\alpha_C(s)}$ and  $k$ is the unique morphism with $k \circ \iota_{f,s} = \iota_{f,\alpha_C(s)}$.
    Using the naturality of $\alpha$ and the universal property of the top left coproduct, it can be shown that the left square commutes for $\phi,\phi'$ and $\psi,\psi'$. Then we also have $\pi' \circ h \circ \psi = \pi' \circ h \circ \phi$, because
    $$\pi' \circ h \circ \psi = \pi' \circ \psi' \circ k = \pi' \circ \phi' \circ k = \pi' \circ h \circ \phi.$$
    With the universal property of the top right coequalizer we get the unique morphism $e$ with the desired properties.
  \end{Proof}
\end{Lemma}

\begin{Remark} \label{Rem:CoequalizerFormula} In \cref{Lem:CoequalizerFormula}, we can replace the target category $\SET^{\mathcal{C}^\op}$ of $y : \mathcal{C} \rightarrow \SET^{\mathcal{C}^\op}$ with an arbitrary {\bf full} subcategory $\mathcal{F}$ that at least includes every functor in the image of the Yoneda embedding. The lemma still holds, because we only form colimits in $\mathcal{E}$, while the functors in $\SET^{\mathcal{C}^\op}$ or $\mathcal{F}$ serve as indexing sets for the colimit in $\mathcal{E}$. We are allowed to replace those indexing sets with isomorphic ones. However, it is essential that $\mathcal{F}$ is a full subcategory. Otherwise the Yoneda lemma does not necessarily apply in the proof of \cref{Lem:CoequalizerFormula}.
\end{Remark}

We will later use \cref{Lem:CoequalizerFormula} with \cref{Rem:CoequalizerFormula} to derive \cref{Cor:SimplicialKanExt}, which guarantees the existence of Spivak's metric realization \cref{Def:MetReSpivak} and our version in \cref{Def:MetRe} and their right adjoints.

One reason to consider left Kan extensions along the Yoneda embedding is because we always obtain an adjoint.

\begin{Remark} \label{Prop:KanExtYoneda} \cite[Remark~6.5.9]{riehl2017category} The left Kan extension $\Lan{y}{F}$ of $F : \mathcal{C} \rightarrow \mathcal{E}$ along $y$ given by \cref{Thm:ColimitFormula}
  \begin{center}
    \begin{tikzcd}[column sep=large, row sep=large]
      \mathcal{C} \arrow[rr, "F"] \arrow[rd, "y"'] & {}                           & \mathcal{E} \arrow[ld, "R", bend left=30]      \\
      & \SET^{\mathcal{C}^\op} \arrow[ru, "\Lan{y}{F}", pos=0.45, bend left=0] & {} 
      \arrow[from=2-2, to=1-3, "\dashv", sloped, rotate=90, phantom, bend right=13]
    \end{tikzcd}
  \end{center}
  is left adjoint to the functor $R = \Hom{\mathcal{E}}{F(-)}{-} : \mathcal{E} \rightarrow \SET^{\mathcal{C}^\op}$.
\end{Remark}

When we replace $\SET^{\mathcal{C}^\op}$ with a full subcategory $\mathcal{F}$, we cannot rely on this result as presented in \cite[Remark~6.5.9]{riehl2017category}. We thus introduce a slightly more general result that allows $\SET^{\mathcal{C}^\op}$ to be replaced if additional conditions hold. This version is specifically tailored to be used later in \cref{Cor:SimplicialKanExt}. The following proof is a direct generalization of a proof for \cref{Prop:KanExtYoneda}.

\begin{Proposition}\label{Prop:KanExtYonedaSubcat} Let $\mathcal{C}$ be small, let $F : \mathcal{C} \rightarrow \mathcal{E}$ be a functor, and let $\mathcal{S}$ be a full subcategory of $\SET^{\mathcal{C}^{\op}}$ such that
  \begin{itemize}
  \item the image of the Yoneda embedding lies in $\mathcal{S}$,
  \item the image of the functor $R = \Hom{\mathcal{E}}{F(-)}{-} : \mathcal{E} \rightarrow \mathcal{S}$ also lies in $\mathcal{S}$.
  \end{itemize}
  Then the left Kan extension $\Lan{y}{F}$ of $F$ along $y$ given by \cref{Thm:ColimitFormula}
  \begin{center}
    \begin{tikzcd}[column sep=large, row sep=large]
      \mathcal{C} \arrow[rr, "F"] \arrow[rd, "y"'] & {}                           & \mathcal{E} \arrow[ld, "R", bend left=30]      \\
      & \mathcal{S} \arrow[ru, "\Lan{y}{F}", pos=0.45, bend left=0] & {} 
      \arrow[from=2-2, to=1-3, "\dashv", sloped, rotate=90, phantom, bend right=13]
    \end{tikzcd}
  \end{center}
  is left adjoint to the functor $R = \Hom{\mathcal{E}}{F(-)}{-} : \mathcal{E} \rightarrow \mathcal{S}$ that assigns to
  \begin{itemize}
  \item an object $E \in \Ob{\mathcal{E}}$ the object $R(E) = \Hom{\mathcal{E}}{F(-)}{E} \in \mathcal{S}$,
  \item a morphism $e \in \Hom{\mathcal{E}}{E}{E'}$ the natural transformation $R(e) : R(E) \Rightarrow R(E')$ with components
    $$R(e)_C = \Hom{\mathcal{E}}{F(C)}{e} : \Hom{\mathcal{E}}{F(C)}{E} \rightarrow \Hom{\mathcal{E}}{F(C)}{E'}, f \mapsto e \circ f.$$
  \end{itemize}
  \begin{Proof} \quad \\
    We show that $\Lan{y}{F} : \mathcal{S} \rightarrow \mathcal{E}$ is left adjoint to $R : \mathcal{E} \rightarrow \SET^{\mathcal{C}^{\op}}$ by constructing bijections
    \begin{equation} \label{eq:homBijection}
      \Hom{\mathcal{S}}{S}{R(E)} \cong \cocone{FP^{S}}{E} \cong \Hom{\mathcal{E}}{\Lan{y}{F}(S)}{E}
    \end{equation}
    that are natural in $S$ and $E$ for all objects $S \in \mathcal{S}$ and objects $E$ in $\mathcal{E}$. Here,
    $P^S : y \downarrow S \rightarrow \mathcal{C}$ is the projection functor for the comma category $y \downarrow S$.
    \bigbreak
    1. The second bijection in (\ref{eq:homBijection}) follows from the colimit formula. The universal property of the colimit states that there is a bijection
    \[ \cocone{FP^S}{E} \cong \Hom{\mathcal{E}}{\colim{FP^{S}}}{E} \cong \Hom{\mathcal{E}}{\Lan{y}{F}(S)}{E} \]
    that is natural in $E$. Its naturality in $S$ follows from the fact that $\Lan{y}{F} : \mathcal{S} \rightarrow \mathcal{E}$ is a functor and the functoriality of the colimit.
    \bigbreak
    2. To construct the first bijection in (\ref{eq:homBijection}), we consider the comma category $y \downarrow S$.
    \begin{itemize}
    \item Its objects are pairs $(C,\nu)$ where $C$ is an object in $\mathcal{C}$ and $\nu : \Hom{\mathcal{C}}{-}{C} \Rightarrow S$ is a natural transformation.
      By the Yoneda lemma, the latter are in bijection with elements of $S(C)$. For each $s \in S(C)$ the unique natural transformation $\nu^s : \Hom{\mathcal{C}}{-}{C} \Rightarrow S$ with $\nu^s_C(\id{C}) = s$ has component morphisms $$\nu^s_{C'} : \Hom{\mathcal{C}}{C'}{C} \rightarrow S(C'), \; f \mapsto S(f)(s).$$
    \item A morphism $c : (C, \nu) \rightarrow (C',\nu')$ in $y \downarrow S$ is a morphism $c : C \rightarrow C'$ in $\mathcal{C}$ that satisfies $\nu' \circ \Hom{\mathcal{C}}{-}{c} = \nu$, or equivalently,
      \begin{equation} \label{eq:yoneda}
        \nu'_{C'}(c) = S(c)(\nu'_{C'}(\id{C'})) = \nu_C(\id{C}).
      \end{equation}
    \end{itemize}
    The projection $P^S : y \downarrow S \rightarrow \mathcal{C}$ sends each object $(C,\nu)$ to $C$ and each morphism to itself.
    \bigbreak
    3. We show that there are bijections $\Hom{\mathcal{S}}{S}{R(E)} \cong \cocone{FP^{S}}{E}$ that are natural in $E$ and $S$. For this, we consider the maps
    \begin{align*}
      \phi_{S,E} : \quad &\Hom{\mathcal{S}}{S}{R(E)} &\rightarrow &\qquad \cocone{FP^S}{E}\\
                  &\alpha : S \Rightarrow R(E)         &\mapsto &\qquad \lambda^\alpha : FP^S \Rightarrow \Delta(E)\\
                  &                     &   &\qquad \lambda^\alpha_{(C, \nu)} = \alpha_C(\nu_C(\id{C})) : F(C) \rightarrow E\\
      \psi_{S,E} : \quad &\cocone{FP^S}{E} &\rightarrow &\qquad \Hom{\mathcal{S}}{S}{R(E)}\\
                  &\lambda : FP^S \Rightarrow \Delta(E)       &\mapsto &\qquad \beta^\lambda : S \Rightarrow R(E)\\
                  &                      &   &\qquad \beta^\lambda_C : S(C) \rightarrow \Hom{\mathcal{E}}{F(C)}{E}, s \mapsto \lambda_{(C,\nu^s)}.
    \end{align*}

    To show that $\lambda^\alpha$ is indeed a cocone for each natural transformation $\alpha : S \Rightarrow R(E)$, we compute for a morphism $c : (C,\nu) \to (C',\nu')$ in $y \downarrow S$
    \begin{align*}
      \lambda^\alpha_{(C',\nu')} \circ F(c) \quad \overset{(\text{def }\lambda^\alpha)}&{=} \; \; \alpha_{C'} (\nu'_{C'}(\id{C'})) \circ F(c)\\
                          \overset{(\text{def }R(E))}&{=} \, (R(E)(c) \circ \alpha_C)(\nu'_{C'}(\id{C'}))\\
                          \overset{(\text{nat }\alpha)}&{=} \; \; \, (\alpha_C \circ S(c))(\nu'_{C'}(\id{C'})) \\
                          \overset{(\ref{eq:yoneda})}&{=} \quad \; \alpha_C(\nu_C(\id{C})) = \lambda^\alpha_{(C,\nu)}.
    \end{align*}

    To show that $\beta^\lambda : S \Rightarrow R(E)$ is a natural transformation for each cocone $\lambda : FP^S \Rightarrow \Delta(E)$, note that by (\ref{eq:yoneda}), each morphism $c : C \to C'$ in $\mathcal{C}$ defines a morphism $c : (C, \nu^{S(c)(s')}) \to (C', \nu^{s'})$ in $y \downarrow S$ for all $s' \in S(C')$, because one has $$\nu_{C'}^{s'}(c) = S(c)(\nu^{s'}_{C'}(\id{C'})) = S(c)(s') = \nu_{C}^{S(c)(s')}(\id{C}).$$ This implies for all $s' \in S(C')$:
    \begin{align*}
      \beta_{C}^{\lambda} \circ S(c)(s') =& \beta_{C}^{\lambda}(S(c)(s'))\\
      =& \lambda_{(C,\nu^{S(c)(s')})}\\
      \stackrel{\lambda \text{ cocone}}{=}& \lambda_{(C',\nu^{s'})} \circ F(c) = \beta_{C'}^{\lambda}(s') \circ F(c) = (R(E)(c) \circ \beta_{C'}^{\lambda})(s').
    \end{align*}
    A direct computation then shows that $\psi_{S,E}$ is the inverse of $\phi_{S,E}$.
    \begin{align*}
      (\psi_{S,E} \circ \phi_{S,E})(\alpha)_C(s) &= \psi_{S,E}(\lambda^{\alpha})_C(s) = \lambda^{\alpha}_{(C,\nu^s)} = \alpha_C(\nu^s_C(\id{C})) = \alpha_C(s)\\
      (\phi_{S,E} \circ \psi_{S,E})(\lambda)_{(C,\nu)} &= \phi_{S,E}(\beta^\lambda)_{(C,\nu)} = \beta^\lambda_C(\nu_C(\id{C})) = \lambda_{(C,\nu)}.
    \end{align*}
    The naturality of $\phi_{S,E}$ follows, because we have for each natural transformation $\beta : S' \Rightarrow S$,
    morphism $e : E \to E'$ and each object $(C,\nu)$ in $y \downarrow S$
    \begin{align*}
      \phi_{S',E'}(\Hom{\mathcal{S}}{\beta}{R(e)}(\alpha))_{(C,\nu)} \quad \overset{(\text{def }\phi)}&{=} \quad (\Hom{\mathcal{S}}{\beta}{R(e)}(\alpha)_C)(\nu_C(\id{C}))\\
                                           \overset{(\text{hom func.})}&{=} \,  (e \circ \alpha_C \circ \beta_C)(\nu_C(\id{C}))\\
                                           \overset{(\text{def } \Delta)}&{=} \quad \Delta(e) \circ \alpha_C((\beta \circ \nu)_C(\id{C}))\\
                                           \overset{(\text{def } \phi)}&{=} \quad (\Delta(e) \circ \phi_{S,E}(\alpha))_{(C,\beta\circ\nu)}.
    \end{align*}
  \end{Proof}
\end{Proposition}

\subsection{The Simplex Category and (Co)Simplicial Objects} \label{Sec:Simplex}
We will use the same definition and mostly the same indexing convention for the simplex category as in Richter's textbook \citetitle{Richter_2020} \cite[Chapter~10]{Richter_2020}. However, we will explicitly add the index $n$ to the face $\delta^n_i$ and degeneracy morphisms $\sigma^n_j$ while Richter simply writes $\delta_i$ and $\sigma_i$. Just like Richter, we use the topologist's convention, where objects of the simplex category are denoted $[n] = \{ 0,1,\ldots,n \}$ instead of the algebraist's convention where $[n] = \{ 0,1,\ldots,n - 1 \}$.
\begin{Definition} \cite[Lemma~10.1.1]{Richter_2020} The \defind{simplex category} \defind{$\Delta$} has
  \begin{itemize}
  \item as objects the finite non-empty \defind{ordinal numbers} $[n] = \{ 0,1,\ldots,n \}$ for $n \in \mathbb{N}$,
  \item as morphisms $f : [m] \rightarrow [n]$ weakly monotonic maps $f : \{ 0,1,\ldots,m \} \rightarrow \{ 0,1,\ldots,n \}$.
  \end{itemize}
\end{Definition}

All morphisms of this category can be factored into two types of elementary morphisms.

\begin{Definition} \cite[Lemma~10.1.2]{Richter_2020}  The $i$th \defind{face morphism} $\delta^n_i : [n-1] \rightarrow [n]$ for $0 \le i \le n$ and the $j$th \defind{degeneracy morphism} $\sigma^n_j : [n+1] \rightarrow [n]$ for $j \in [n]$ are the morphisms
    \[ \delta^n_i(k) =
      \begin{cases*}
        k   & $0 \le k < i$\\
        k+1 & $i \le k < n$
      \end{cases*} \qquad
      \sigma^n_j(k) =
      \begin{cases*}
        k   & $0 \le k \le j$\\
        k-1 & $j < k \le n+1.$
      \end{cases*}
    \]
  
\end{Definition}

\begin{Lemma} \cite[Lemma~10.1.4]{Richter_2020} \label{prop:factorisation-simplex}
  Every morphism $f : [n] \to [m]$ in $\Delta$ can be expressed uniquely as a composite
  \[
    f = \delta^{m}_{i_1} \circ \cdots \circ \delta^{n+s}_{i_r} \circ \sigma^{n+s}_{j_1} \circ \cdots \circ \sigma^{n+1}_{j_s}
  \]
  with $0 \le i_r < \cdots < i_1 \le m$ and $0 \le j_1 < \cdots < j_s < n$, where $m = n - s + r$.
\end{Lemma}

Having defined the simplex category, we can define simplicial objects over arbitrary categories $\mathcal{C}$ as functors from the opposite of the simplex category into $\mathcal{C}$.
\begin{Definition} \cite[Lemma~10.2.1]{Richter_2020} Let $\mathcal{C}$ be a category.
  \begin{enumerate}
  \item A functor $S : \Delta^{\op} \rightarrow \mathcal{C}$ is called a \defind{simplicial object} in $\mathcal{C}$. Natural transformations between such functors are called \defind{simplicial morphisms}. The functor category $\mathcal{C}^{\Delta^\op}$ is called the \defind{category of simplicial objects} in $\mathcal{C}$.
  \item A functor $S : \Delta \rightarrow \mathcal{C}$ is called a \defind{cosimplicial object} in $\mathcal{C}$. Natural transformations between such functors are called \defind{cosimplicial morphisms}. The functor category $\mathcal{C}^{\Delta}$ is called the  \defind{category of cosimplicial objects} in $\mathcal{C}$.
  \end{enumerate}
\end{Definition}

A (co)simplicial object in $\SET$ is also called a \textbf{(co)}\defind{simplicial set} and a (co)simplicial morphism in $\SET$ is called a \textbf{(co)}\defind{simplicial map}. The category $\SET^{\Delta^\op}$ of simplicial objects in $\SET$ is also called the \defind{category of (co)simplicial sets}.

Just like morphisms in the simplex category can be factored into the face and degeneracy maps, we can also decompose simplicial objects into elementary components that satisfy certain rules. 
\begin{Remark} \cite[(10.2.1)]{Richter_2020}  \label{Rem:SimpCoSimpRelations} Let $\mathcal{C}$ be a category.
  \begin{enumerate}
  \item A simplicial object $C : \Delta^{op} \to \mathcal{C}$ is given by
    \begin{itemize}
    \item a family $(C_n)_{n \in \mathbb{N}_0}$ of objects $C_n \in \text{Ob}\mathcal{C}$,
    \item families of morphisms $d^i_n : C_n \to C_{n-1}$ for $n \in \mathbb{N}$, $0 \leq i \leq n$, the \defind{face maps},
    \item families of morphisms $s^j_n : C_n \to C_{n+1}$ for $n \in \mathbb{N}_0$ and $0 \leq j \leq n$, the \defind{degeneracies}
    \end{itemize}
    with $C_n = C([n])$, $d^i_n = C(\delta^{n+1}_i)$ and $s^j_n = C(\sigma^n_j)$, satisfying the \defind{simplicial relations}
    \begin{align*}
      d^i_{n-1} \circ d^j_n &= d^{j-1}_{n-1} \circ d^i_n \quad \text{for } i < j \\
      s^i_{n+1} \circ s^j_n &= s^{j+1}_{n+1} \circ s^i_n \quad \text{for } i \leq j, \text{ and} \\
      d^i_n \circ s^j_{n-1} &= \begin{cases}
        s^{j-1}_n \circ d^i_{n-1}  & i < j \\
        \text{id}_{C_n}          & i \in \{j, j+1\} \\
        s^j_n \circ d^{i-1}_{n-1} & i > j+1.
      \end{cases}
    \end{align*}
  \item A cosimplicial object $C : \Delta \to \mathcal{C}$ is given by
    \begin{itemize}
    \item a family $(C^n)_{n \in \mathbb{N}_0}$ of objects $C^n \in \text{Ob}\mathcal{C}$,
    \item families of morphisms $d_i^n : C^{n-1} \to C^n$ for $n \in \mathbb{N}$, $0 \leq i \leq n$, the \defind{coface maps},
    \item families of morphisms $s_j^n : C^{n+1} \to C^n$ for $n \in \mathbb{N}_0$ and $0 \leq j \leq n$, the \defind{codegeneracies}
    \end{itemize}
    with $C^n = C([n])$, $d_i^n = C(\delta^{n+1}_i)$ and $s_j^n = C(\sigma^n_j)$, satisfying the \defind{cosimplicial relations}
    \begin{align*}
      d_j^{n+1} \circ d_i^n &= d_i^{n+1} \circ d_{j-1}^n \quad \text{for } i < j \\
      s_j^n \circ s_i^{n+1} &= s_i^n \circ s_{j+1}^{n+1} \quad \text{for } i \leq j, \text{ and} \\
      s_j^{n-1} \circ d_i^n &= \begin{cases}
        d_i^{n-1} \circ s_{j-1}^n  & i < j \\
        \text{id}_{C^n}          & i \in \{j, j+1\} \\
        d_{i-1}^{n-1} \circ s_j^n & i > j+1.
      \end{cases}
    \end{align*}
  \end{enumerate}
\end{Remark}

\subsection{The Geometric Realization} \label{Subsection:GeometricRealization}
The geometric realization realizes a simplicial set $S$ as a topological space by mapping elements of $S$ to basic building blocks called the topological $n$-simplices. Elements of $S_0$ map to points, elements of $S_1$ to line segments, elements of $S_2$ to triangles, and so on. We then glue them according to the face and degeneracy maps of $S$. The geometric realization is then defined as a left Kan extension along the Yoneda embedding. We begin by defining the topological $n$-simplices.

\begin{Definition} \cite[Definition 10.6.1]{Richter_2020} \label{Def:TopSimplex}\\ Let $\mathbb{R}_{\geq 0}^n = \{ (x_0,\ldots , x_{n-1}) \in \mathbb{R}^n \, | \, x_i \geq 0 \; \forall i \}$.
  \begin{itemize}
  \item For $n \in \mathbb{N}_0$, the \defind{topological $n$-simplex} is the subspace
    \[ \Delta^{n} = \{ (x_0,\ldots,x_n) \in \mathbb{R}_{\geq 0}^{n+1} \, | \, x_0 + \cdots + x_n = 1 \, \} \subseteq \mathbb{R}_{\geq 0}^{n+1} \]
    equipped with the standard topology.
  \item For $n \in \mathbb{N}$ and $i \in \{0, \ldots, n\}$ the $i$th \defind{face map} $f^{n}_i : \Delta^{n-1} \rightarrow \Delta^{n}$ inserts 0 at the $i$th position in the coordinate vector and shifts the remaining entries to the right by one:
    \[ f^{n}_i(x_0,\ldots,x_{n-1}) = (x_0, \ldots, x_{i-1},0,x_i, \ldots, x_{n-1}). \]
  \item For $n \in \mathbb{N}_0$ and $i \in \{0, \ldots, n\}$ the $i$th \defind{degeneracy map} $s^{n}_i : \Delta^{n+1} \rightarrow \Delta^{n}$ adds the $i$th and $(i+1)$th coordinates:
    \[s^{n}_i(x_0,\ldots,x_{n+1}) = (x_0, \ldots, x_{i-1}, x_{i} + x_{i+1}, x_{i+2}, \ldots, x_{n+1}). \]
  \end{itemize}
\end{Definition}

\begin{herefigure}
  \centering
  \begin{tikzpicture}[scale=1.5]
    \begin{scope}[shift={(0,0)}]
      \draw[thick,->] (0,0) -- (2,0) node[anchor=west]{$x_0$};
      \fill[black] (0,0) circle (1.5pt) node[below, black] {$0$}; 
      
      \coordinate (A) at (1.5,0,0);

      \fill[red] (A) circle (1.5pt) node[below, black] {$1$}; 
      \draw (1,-1.5) node {$\Delta^{0}$};
    \end{scope}
    \begin{scope}[shift={(3.5,0)}]
      \draw[thick,->] (0,0) -- (2,0) node[anchor=west]{$x_0$};
      \draw[thick,->] (0,0) -- (0,2) node[anchor=south]{$x_1$};
      \fill[black] (0,0) circle (1.5pt) node[below left, black] {$0$}; 

      \coordinate (A) at (1.5,0,0);
      \coordinate (B) at (0,1.5,0);

      \fill[red] (A) circle (1.5pt) node[below, black] {$1$}; 
      \fill[red] (B) circle (1.5pt) node[left, black] {$1$}; 

      \draw[very thick, red, postaction={decorate, decoration={markings, mark=at position 0.6 with {\arrow[scale=1.25]{stealth}}}}] (A) -- (B);
      \draw (1,-1.5) node {$\Delta^{1}$};
    \end{scope}
    \begin{scope}[shift={(7,0)}]

      \draw[thick,->] (0,0,0) -- (2,0,0) node[anchor=west]{$x_0$};
      \draw[thick,->] (0,0,0) -- (0,2,0) node[anchor=south]{$x_1$};
      \draw[thick,->] (0,0,0) -- (2,0,2) node[anchor=north west]{$x_2$};
      \fill[black] (0,0) circle (1.5pt) node[below left, black] {$0$}; 

      \coordinate (A) at (1.5,0,0);
      \coordinate (B) at (0,1.5,0);
      \coordinate (C) at (1.5,0,1.5);

      \fill[red] (A) circle (1.5pt) node[shift={(0.2, 0.2)}, black] {$1$}; 
      \fill[red] (B) circle (1.5pt) node[left, black] {$1$};                 
      \fill[red] (C) circle (1.5pt) node[shift={(-0.2, -0.2)}, black] {$1$}; 

      \draw[very thick, red, postaction={decorate, decoration={markings, mark=at position 0.6 with {\arrow[scale=1.25]{stealth}}}}] (A) -- (B);
      \draw[very thick, red, postaction={decorate, decoration={markings, mark=at position 0.6 with {\arrow[scale=1.25]{stealth}}}}] (B) -- (C);
      \draw[very thick, red, postaction={decorate, decoration={markings, mark=at position 0.6 with {\arrow[scale=1.25]{stealth}}}}] (A) -- (C);

      \fill[red, opacity=0.25] (A) -- (B) -- (C) -- cycle;
      \draw (1,-1.5) node {$\Delta^{2}$};
    \end{scope}
  \end{tikzpicture}
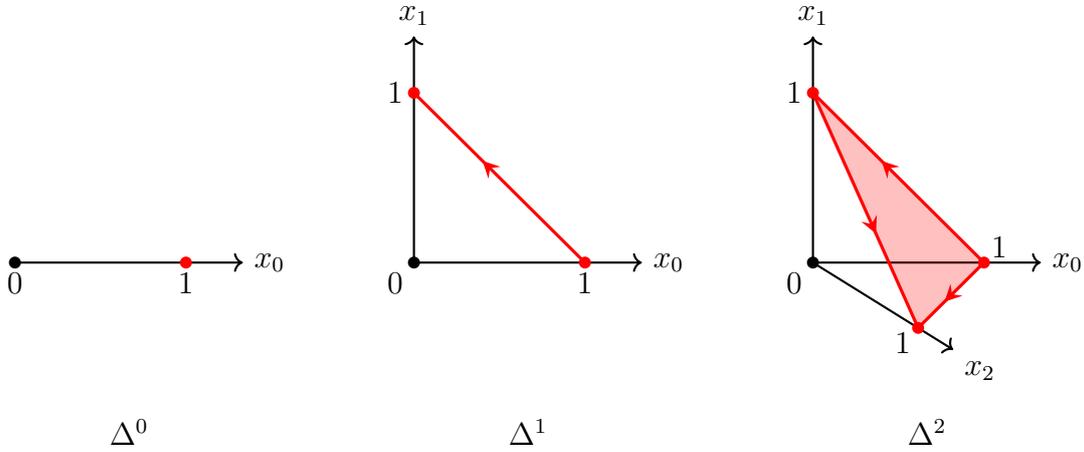
  \caption{\label{metricsimplex} The topological $n$-simplex for $n = 0, 1, 2$.}
\end{herefigure}

It can be shown that the face and degeneracy maps satisfy the relations from \cref{Rem:SimpCoSimpRelations}. This allows us to organize the topological $n$-simplices and the face and degeneracy maps into a single functor out of $\Delta$.
\begin{Definition} \label{Def:G} The topological $n$-simplices define a functor $G : \Delta \rightarrow \Top$ with
  \[ G([n]) = \Delta^n, \qquad G(\delta^{n+1}_i) = f^{n+1}_i, \qquad G(\sigma^n_j) = s^n_j. \]
\end{Definition}

With this functor, we can derive the geometric realization.
\begin{Definition} \cite[Example 1.5.3]{Riehl_2014} \label{Def:TopRe} The \defind{geometric realization} $\TopRe := \Lan{y}{G}$ is the left Kan extension of $G$ along the Yoneda embedding $y$. \;\\
  \begin{center}
    \begin{tikzcd}
      \Delta \arrow[rr, "G"] \arrow[rd, "y"'] & {} \arrow[d, "\eta", Rightarrow, shift right] & \Top \\
      & \SET^{\Delta^\op} \arrow[ru, "\TopRe = \Lan{y}{G}"', dashed]  &
    \end{tikzcd}
  \end{center}
  Its left adjoint is called the \defind{singular nerve} $\Sing = \Hom{\Top}{G(-)}{-}$.
\end{Definition}

The existence of the Kan extension in \cref{Def:TopRe} is guaranteed to exist by \cref{Cor:ColimitFormula}. \Cref{Prop:KanExtYonedaSubcat} guarantees the adjoint.

With the formula in \cref{Lem:CoequalizerFormula}, we can describe the geometric realization more concretely in terms of topological simplices.

\begin{Lemma} \label{Lem:GeomCoeq} The geometric realization $\TopRe : \SET^{\Delta^\op} \rightarrow \Top$ assigns to
  \begin{itemize}
  \item a simplicial set $S : \Delta^{\op} \rightarrow \SET$, the topological space
    \[ \TopRe(S) = \coprod_{\substack{n \in \mathbb{N}_0\\ s \in S_n}} \Delta^n / \sim_G \]
    with the equivalence $\sim_G$ generated by
    \[ \iota_{m,S(f)(s)}(x) \sim_G \iota_{n,s}(G(f)(x))  \] \bigbreak
    for all $f : [m] \rightarrow [n], \; s \in S_n, \; x \in \Delta^m$.
  \item a simplicial map $\alpha : S \Rightarrow S'$ the continuous map
    \begin{align*}
      \TopRe&(\alpha) : \TopRe(S) \rightarrow \TopRe(S'), \quad [\iota_{n,s}(x)]) \mapsto [\iota_{n,\alpha_n(s)}(x)].
    \end{align*}
  \end{itemize}
  \begin{Proof}
    By \cref{Lem:CoequalizerFormula} the left Kan extension $\Lan{y}{G}(S)$ can be expressed as a coequalizer
    $$\coeq{ \coprod_{\substack{f : [m] \rightarrow [n]\\ s \in S_n}} \Delta^{m}  \overset{\iota_{n,s} \circ G(f)}{\underset{\iota_{m,S(f)(s)}}{\rightrightarrows}} \coprod_{\substack{m \in \mathbb{N}_0\\ s \in S_m}} \Delta^{m} },$$
    which can be expressed as the quotient in the statement above.
    
    \Cref{Lem:CoequalizerFormula} also provides the desired expression for the action on morphisms, if one replaces the coequalizer surjection $\pi$ with the equivalence class notation $[-]$.
  \end{Proof}
\end{Lemma}

The formula in \cref{Lem:GeomCoeq} appears in \cite[Definition~10.6.1]{Richter_2020} as an alternative definition of \cref{Def:TopRe}.

\subsection{Locales} \label{Sec:Locales}
In this thesis, we will need locales, a form of point-free topology. An ordinary topological space $(X,\mathcal{O}_X)$ is defined on a carrier set $X$ that is equipped with a topology $\mathcal{O}_X$. Instead of demanding the open sets to be subsets of some other set $X$, we could simply have a set $\mathcal{L}$ that comes with enough operations and rules such that the elements of $\mathcal{L}$ behave like open sets but are not necessarily made up of points.

Open sets can be subsets of each other. To mirror this, we will demand that $\mathcal{L}$ comes equipped with a partial order $\le$. In the resulting poset category $(\mathcal{L}, \le)$, the coproducts and products, called join $(\vee)$ and meet $(\wedge)$, are natural candidates to mirror the union $(\cup)$ and intersection $(\cap)$ of open sets, if they exist. So we shall demand that arbitrary coproducts and finite products exist in $(\mathcal{L},\le)$. The intersection and union of sets also satisfy the infinite distributivity law, so we will demand it as well. For a textbook reference, see \citeauthor{maclane1992sheaves}, \citetitle{maclane1992sheaves} \cite[Chapter~IX]{maclane1992sheaves}.

\begin{Remark} A \defind{poset} $(P,\le)$ can be viewed as a category with the elements of $P$ as objects. Between objects $p,p'$ there is a single morphism $* : p \rightarrow p'$ if $p \le p'$, otherwise the set $\Hom{(P,\le)}{p}{p'}$ is empty.
\end{Remark}

\begin{Definition} \cite[Chapter~IX,1]{maclane1992sheaves} A poset $(\mathcal{L},\le)$ is called a \defind{locale} if:
  \begin{itemize}
  \item suprema $\bigvee_{i\in I}a_i$ exist for any family $(a_i)_{i\in I},$
  \item infima $\bigwedge_{i\in I}b_i$ exist for any {\bf finite} family $(b_i)_{i\in I},$
  \end{itemize}
  such that the infinite distributivity law is satisfied:
  \[ b \wedge (\bigvee_{i\in I}a_i) = \bigvee_{i\in I}(b \wedge a_i) \qquad \text{for all families }(a_i \in \mathcal{L})_{i\in I} \text{ and } b \in \mathcal{L}. \]
  The suprema and infima are also called \defind{join} ($\bigvee$) and \defind{meet} ($\bigwedge$).
\end{Definition}

All ordinary topologies $\mathcal{O}_X$ must include the empty set and the entire space as open sets. These sets can be viewed as least and greatest elements of the poset $(\mathcal{O}_X, \subseteq)$. Locales again mirror this property.
\begin{Notation} A poset $(\mathcal{L},\le)$ has a greatest element and a least element, given by the supremum of the entire set and the supremum of the empty set, respectively. We will denote them by
  \[ \top := \bigvee_{a \in \mathcal{L}} a \qquad \text{ and } \qquad \bot := \bigvee_{a \in \emptyset} a. \]
  The element $\top$ is usually called \defind{top} and $\bot$ is usually called \defind{bottom}.
\end{Notation}

The obvious example of a locale is the poset of open subsets of a topological space. The examples we will need in this thesis are subsets of the real numbers $\mathbb{R}$ or extended real numbers $\mathbb{R} \cup \{ \infty \}$ that have all supremums and infimums.
\begin{Example} \label{Example:Locales} \qquad
  \begin{itemize}
  \item For any topological space $(X,\mathcal{O}_X)$, the open sets form the locale $(\mathcal{O}_X, \subseteq)$.
  \item Any closed interval $[a,b] \subseteq \mathbb{R}$ yields the locale $([a,b], \le)$ where the join and meet are given by the supremum and the infimum in $[a,b]$.
  \item When we reverse the ordering from the previous example, we get the locales $([a,b], \ge)$ where the join and meet are given by the infimum and supremum, respectively. 
  \item If we include infinity with the convention $x \le \infty$ for all $x \in [0,\infty]$, the nonnegative reals with infinity $[0,\infty]$ form the locale $([0,\infty], \le)$.
  \item We can again reverse the order from the previous example and obtain the locale $([0,\infty], \ge)$. In this case, $\infty$ is the bottom element. 
  \end{itemize}
\end{Example}

We have defined locales to resemble topological spaces. Similarly, morphisms of locales should resemble continuous maps.

A continuous map $f : X \rightarrow Y$ maps points from $X$ to points of $Y$, but the condition of continuity is defined backwards. Every open set from the target space $O \subseteq Y$ must also be sent to an open set via the inverse image $f^{-1} : \mathcal{P}(Y) \rightarrow \mathcal{P}(X)$. Here $\mathcal{P}(A)$ denotes the power set of a set $A$.

Because locales do not contain points and the elements $a \in \mathcal{L}$ mirror open sets, we define a morphism from $(\mathcal{L},\le)$ to $(\mathcal{L}',\le')$ as a map $g : \mathcal{L}' \rightarrow \mathcal{L}$ in the other direction, to mirror the inverse image.

The inverse image maps subsets $O \subseteq O'$ to subsets $f^{-1}(O) \subseteq f^{-1}(O')$, so we will demand a morphism of locales to be monotone with respect to the underlying poset. To mirror the other usual properties of the inverse image with respect to the intersection and union of sets, we demand that morphisms of locales preserve finite meets and arbitrary joins.

\begin{Definition} \cite[Chapter~IX,1]{maclane1992sheaves} Let $(\mathcal{L},\le)$ and $(\mathcal{L}',\le')$ be locales. A \defind{morphism of locales} from $(\mathcal{L},\le)$ to $(\mathcal{L}',\le')$ is a monotone map $f : \mathcal{L}' \rightarrow \mathcal{L}$ that preserves finite meets and arbitrary joins.

Locales and their morphisms form the \defind{category of locales} $\Loc$.
\end{Definition}

\begin{Remark} \label{Rem:MorphismOfLocales} Equivalently, a morphism of locales is a functor that preserves finite coproducts and arbitrary products. In that case, the functor axioms correspond to monotonicity.
  
\end{Remark}

\begin{Example} \label{Example:MorphismsOfLocales} \;
  \begin{itemize}
  \item The constant map $$f : [c,d] \rightarrow [a,b], \qquad x \mapsto a$$ defines the morphism of locales $f : ([a,b], \le) \rightarrow ([c,d], \le)$.
  \item The map $$i : [0,1] \rightarrow [0,\infty], \qquad x \mapsto \begin{cases}
        -\log(x) & x \neq 0\\
        \infty     & x = 0
  \end{cases}$$
    defines the isomorphism of locales $i : ([0,\infty],\le) \rightarrow ([0,1], \ge)$.
  \end{itemize}
\end{Example}

\begin{Remark} \label{Rem:IsoBottom} An isomorphism of locales $\phi : \mathcal{L} \rightarrow \mathcal{L}'$  maps non-bottom elements $a \neq \bot_{\mathcal{L}'}$ to non-bottom elements $\phi(a) \neq \bot_{\mathcal{L}}$.
  \begin{Proof}
    Let $a \neq \bot_{\mathcal{L}'}$. As $\phi$ is an isomorphism, the underlying map $\phi : \mathcal{L}' \rightarrow \mathcal{L}$ has an inverse $\phi^{-1} : \mathcal{L} \rightarrow \mathcal{L}'$ that also represents a morphism of locales.\smallbreak
    Assume $\phi(a) = \bot_{\mathcal{L}}$. Then we have $$a = \phi^{-1}(\phi(a)) = \phi^{-1}(\bot_{\mathcal{L}}) = \bot_{\mathcal{L}'},$$ where the last equality is due to the fact that $\phi^{-1}$ preserves empty joins. \smallbreak
    This contradicts $a \neq \bot_{\mathcal{L}'}$, thus $\phi(a) \neq \bot_{\mathcal{L}}$.
  \end{Proof}
\end{Remark}

We will need one more definition from \citeauthor{BarrFuzzy} \cite{BarrFuzzy} to later derive his central result about valued sets (\cref{Thm:EquivCM}) and an additional one for simplicial objects over this category (\cref{Thm:EquivCValSVal}). This property does not directly correspond to a topological property like connectedness, so we cannot make a topological analogy here.

\begin{Definition} \cite[Section~3]{BarrFuzzy} A locale $\mathcal{L}$ is called \defind{totally connected} if $a \wedge b \neq \bot$ for all elements $a,b \neq \bot$ in $\mathcal{L}$.
\end{Definition}

All the locales we will use later in this thesis are totally connected. 
\begin{Example} The locales $([a,b], \le)$ and $([0,\infty], \le)$ from \cref{Example:Locales} are totally connected as we have $x \wedge y = \min(x,y)$.
\end{Example}

Locales without this property exist, as shown in the following example, but we will not use such locales later. 
\begin{Example} The locale $([0,1] \times [0,1], \le)$ where $(a,b) \le (c,d) :\Leftrightarrow a \le c \text{ and } b \le d$ is not totally connected.
  \begin{Proof}
    The suprema and infima are given by
    \[ \bigvee_{i\in I}(a_i,b_i) = (\sup_{i\in I}(a_i),\sup_{i\in I}(b_i)) \qquad \bigwedge_{i\in I}(a_i,b_i) = (\inf_{i\in I}(a_i),\inf_{i\in I}(b_i)) \]
    We can visualize this nicely for two elements $x \in [0,1] \times [0,1]$ and $y \in [0,1] \times [0,1]$: 
      
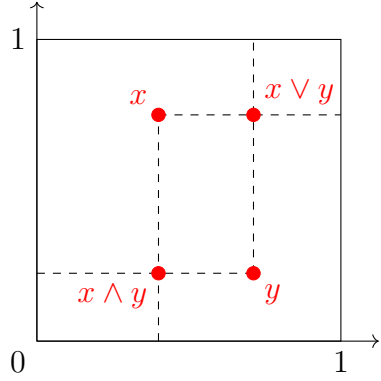
\begin{center}
        \begin{tikzpicture}
          \def\xleft{1.6}    
          \def\xright{2.85}   
          \def\ytop{3}     
          \def\ybottom{0.9}  
          
          \draw[->] (0,0) -- (4.5,0) node[right] {};
          \draw[->] (0,0) -- (0,4.5) node[above] {};
          
          \node[below left] at (0,0) {$0$};
          \node[below] at (4,0) {$1$};
          \node[left] at (0,4) {$1$};
          
          \draw (0,0) -- (0,4) -- (4,4) -- (4,0) -- cycle;
          
          \coordinate (x) at (\xleft,\ytop);        
          \coordinate (xvory) at (\xright,\ytop); 
          \coordinate (y) at (\xright,\ybottom);    
          \coordinate (xwedgey) at (\xleft,\ybottom); 
          
          \draw[dashed] (\xleft,\ytop) -- (4,\ytop);  
          \draw[dashed] (0,\ybottom) -- (\xright,\ybottom); 
          
          \draw[dashed] (\xleft,0) -- (\xleft,\ytop);     
          \draw[dashed] (\xright,\ybottom) -- (\xright,4);   
          
          \filldraw[red] (x) circle (2.5pt) node[above left] {$x$};
          \filldraw[red] (xvory) circle (2.5pt) node[above right] {$x \vee y$};
          \filldraw[red] (y) circle (2.5pt) node[below right] {$y$};
          \filldraw[red] (xwedgey) circle (2.5pt) node[below left] {$x \wedge y$};
        \end{tikzpicture}
      \end{center}
      We can see that the meet of the elements $(0,1),(1,0) \neq \bot$ is $(0,1) \wedge (1,0) = (0,0) = \bot$.
    \end{Proof}
  \end{Example}

\subsection{Sheaves on Locales} \label{Sec:SheavesOnLocales}
  One of the goals of this thesis is to derive the metric realization of fuzzy simplicial sets. In this process, we will encounter sheaves on locales. In this subsection, we present the conventional sheaf condition (\cref{Def:SheavesOnLocales}), as well as an equivalent alternative definition (\cref{Def:SheavesOnLocalesNoBot}) tailored to the specific applications in this thesis. An abstract treatment of sheaves on locales can be found in \cite[Chapter~IX]{maclane1992sheaves}. However, because we only need sheaves on locales to summarize results from Barr \cite[Section~2]{BarrFuzzy}, we will instead use his less abstract definition provided in \cite[Section~2]{BarrFuzzy}.
  \bigbreak

We begin with the definition of presheaves. These are simply contravariant functors into the category $\SET$.
\begin{Definition} Let $\mathcal{C}$ be a category. A functor $F : \mathcal{C}^{\op} \rightarrow \SET$ is called a \defind{presheaf} on $\mathcal{C}$. The \defind{category of presheaves on $\mathcal{C}$} is the functor category $\SET^{\mathcal{C}^{\op}}$.
\end{Definition}

\begin{Notation} For presheaves $F : (P,\leq)^{\op} \rightarrow \SET$ on a poset category, we will write $F_{a \geq b} : F(a) \rightarrow F(b)$ for the restriction maps $F(f : a \rightarrow b) : F(a) \rightarrow F(b)$ as there is only one morphism $f : a \rightarrow b$ that witnesses $a \geq b$. Note that we also flip the poset relation in the index to reduce contravariance confusion: $F_{a \geq b}$ maps from $F(a)$ to $F(b)$.
\end{Notation}

Sheaves are presheaves that satisfy the sheaf condition. We require this condition only for locales, where the sheaf condition takes a much simpler form.

\begin{Definition} \cite[Section~2]{BarrFuzzy} \label{Def:SheavesOnLocales} Let $\mathcal{L}$ be a locale and $F : \mathcal{L}^{\op} \rightarrow \SET$ be a presheaf.
  \begin{itemize}
  \item For $a \in \mathcal{L}$, a family $(a_i \in \mathcal{L})_{i \in I}$ with $a = \bigvee_{i \in I}a_i$ is called a \defind{cover} of $a$.
  \item Given a cover $(a_i)_{i \in I}$ of $a$, a family $(x_i \in F(a_i))_{i \in I}$ is called \defind{compatible} if
    \[ F_{a_i\geq (a_i \, \wedge \, a_j)}(x_i) = F_{a_j\geq (a_i \, \wedge \, a_j)}(x_j) \qquad \text{for all } i,j \in I. \]
  \item $F$ is called a sheaf on $\mathcal{L}$ if it satisfies the \defind{sheaf condition}: \smallbreak
    For every $a$ with cover $(a_i)_{i \in I}$ and every compatible family $(x_i \in F(a_i))_{i \in I}$, there exists a unique $x \in F(a)$ such that
    $F_{a\geq a_i}(x) = x_i$ for all $i \in I$.
    
    Sheaves on $\mathcal{L}$ with natural transformations between them form the category $\Sheaf{\mathcal{L}}$. It is a full subcategory of $\SET^{\mathcal{L}^{\op}}$.
  \end{itemize}
\end{Definition}

\begin{Remark} \cite[Section~2]{BarrFuzzy} \label{Rem:SheafEmptyCover} Let $F : \mathcal{L}^{\op} \rightarrow \SET$ be a sheaf. Then $F(\bot) \cong \{ * \}$.
  \begin{Proof}
    As $\bot = \bigvee \emptyset$, the empty family $\emptyset$ is a cover of $\bot$.
    Due to the sheaf condition, there must exist a unique element $x \in F(\bot)$ such that the restriction conditions $F_{\bot \geq a_i}(x) = x_i$ are satisfied for all $i$ in the empty index set. Since there are no elements in the empty family, there are no restriction conditions to check, so any element $x \in F(\bot)$ satisfies them vacuously. The sheaf condition states that there is a unique such element in $F(\bot)$, thus $F(\bot)$ must be a singleton set.
  \end{Proof}
\end{Remark}

When defining certain sheaves, it is often necessary to define them by case distinction where $\bot$ maps to $\{ * \}$ and everything else maps to something more interesting. To avoid writing down this case distinction, we can write a single definition to lift a functor $F : \mathcal{L}_{\neq\bot} \rightarrow \SET$ that is not defined at $\bot$ to one that includes $\bot$. If the result is supposed to be a sheaf, this lifting is unique up to unique isomorphism. Note that for a locale $\mathcal{L}$, $\mathcal{L}_{\neq\bot} = \mathcal{L} \setminus \{ \bot \}$ is merely a poset.
\begin{Definition} \label{Def:LiftBot} Let $F : \mathcal{L}_{\neq\bot} \rightarrow \SET$ be a functor.
  Then we define the functor $F^\bot : \mathcal{L} \rightarrow \SET$ by
    \begin{align*}
    F^{\bot}(a) &=
           \begin{cases}
             F(a) & a \neq \bot\\
             \{ * \}  & a = \bot
           \end{cases}\\
    F^{\bot}(* : a \le b)(x) &= 
           \begin{cases}
             F(* : a \le b)(x) & a \neq \bot \text{ and } b \neq \bot\\
             *         & a = \bot \text{ or } b = \bot.
           \end{cases}
  \end{align*}
\end{Definition}

\begin{Lemma} \label{Lem:ExtendToSheaf} Let $F : \mathcal{L}_{\neq\bot} \rightarrow \SET$ be a functor. Then $F^\bot$ is a sheaf if and only if $F$ satisfies the \defind{bottomless sheaf condition}: \smallbreak
  For all covers $(a_i \in \mathcal{L}_{\neq\bot})_{i \in I}$ of an element $a \in \mathcal{L}_{\neq\bot}$ and compatible families $(x_i \in F(a_i))_{i \in I}$, there is a unique $x \in F(a)$ such that $F(* : a_i \le a)(x) = x_i$.
  \begin{Proof}
    \begin{enumerate}
      \item Let $F^{\bot}$ satisfy the ordinary sheaf condition. We prove that $F$ satisfies the bottomless sheaf condition: \smallbreak
        Let $(a_i \in \mathcal{L}_{\neq\bot})_{i \in I}$ be a cover of $a \in \mathcal{L}_{\neq\bot}$.
        As $\mathcal{L}_{\neq\bot} \subseteq \mathcal{L}$ and because $F^{\bot}$ and $F$ act identically on non-bottom elements,
        we can then use the ordinary sheaf condition of $F^{\bot}$ to derive the unique element for every compatible family.
      \item Let $F$ satisfy the bottomless sheaf condition. We prove that $F^{\bot}$ satisfies the ordinary sheaf condition: \smallbreak
        Let $(a_i \in \mathcal{L})_{i \in I}$ be a cover of $a \in \mathcal{L}$. \smallbreak
        If $a = \bot$, we must assign $*$ as the unique element for any compatible family (\cref{Rem:SheafEmptyCover}). \smallbreak
        If $a \in \mathcal{L}_{\neq\bot}$, we can remove every $i$ from $I$ where $a_i = \bot$ and obtain a new family
        $(a_i \in \mathcal{L}_{\neq\bot})_{i \in J}$ which covers $a \in \mathcal{L}_{\neq\bot}$. Then we can use the bottomless sheaf condition of $F$ to obtain the unique element for any compatible family of elements.
    \end{enumerate}
  \end{Proof}
\end{Lemma}

We can also formulate the procedure of adjoining a bottom element as an equivalence of categories.

\begin{Definition} \label{Def:SheavesOnLocalesNoBot} The category $\SheafSt{\mathcal{L}}$ is the full subcategory of $\SET^{\mathcal{L}_{\neq\bot}}$ that contains only functors $F : \mathcal{L}_{\neq\bot} \rightarrow \SET$ which satisfy the condition from \cref{Lem:ExtendToSheaf}.
  
\end{Definition}

\begin{Corollary} \label{Cor:EquivSheafStSheaf} There is an equivalence of categories between $\Sheaf{\mathcal{L}}$ and $\SheafSt{\mathcal{L}}$.
  \begin{Proof}
    We can easily extend \cref{Def:LiftBot} to a functor $(-)^\bot : \SheafSt{\mathcal{L}} \rightarrow \Sheaf{\mathcal{L}}$ by extending the case distinction to morphisms
    \begin{align*}
      (\alpha^\bot)_a= 
           \begin{cases}
             \alpha_a    & a \neq \bot\\
             * \mapsto * & a = \bot.
           \end{cases}
    \end{align*}
    It is then just as easy to see that $(-)^\bot : \SheafSt{\mathcal{L}} \rightarrow \Sheaf{\mathcal{L}}$ and the restriction functor $(-)|_{\mathcal{L}_{\neq\bot}} : \Sheaf{\mathcal{L}} \rightarrow \SheafSt{\mathcal{L}}$ form an equivalence of categories.
  \end{Proof}
\end{Corollary}
The technical reason we do not get a convenient isomorphism of categories in this case is that the singleton set $\{ * \}$ is not unique, but only unique up to unique isomorphism. So a sheaf $F : \mathcal{L} \rightarrow \SET$ from $\Sheaf{\mathcal{L}}$ might map $\bot$ to a different singleton set $\{ *' \}$ but $(F|_{\mathcal{L}_{\neq\bot}})^{\bot}$ will map $\bot$ to the specific singleton set $\{ * \}$ that was chosen in \cref{Def:LiftBot}.

We conclude this section by investigating how categories of sheaves on different locales are related. More general versions of the following results can be found in \cite[Chapter~IX,1]{maclane1992sheaves}. We will present more concrete proofs in terms of Barr's sheaf condition \cite[Section~2]{BarrFuzzy} from \cref{Def:SheavesOnLocales}.

The relationship between different sheaf categories can be explored naturally in the language of category theory: A morphism between locales $\phi : \mathcal{L} \rightarrow \mathcal{L}'$ induces a functor between the categories $\Sheaf{\mathcal{L}}$ and $\Sheaf{\mathcal{L}'}$.

\begin{Lemma} \label{Def:GeometricMorphism} A morphism of locales $\phi : \mathcal{L} \rightarrow \mathcal{L}'$ induces a functor $\Sheaf{\phi} : \Sheaf{\mathcal{L}} \rightarrow \Sheaf{\mathcal{L}'}$ defined by
  \begin{align*}
    &\Sheaf{\phi}(F) = F \circ \phi\\
    &\Sheaf{\phi}(\alpha : F \Rightarrow F')_a = \alpha_{\phi(a)}.
  \end{align*}
  \begin{Proof}
    Note that $\phi$ is a functor $\phi : \mathcal{L}' \rightarrow \mathcal{L}$ by \cref{Rem:MorphismOfLocales}, and thus the restriction maps of $F \circ \phi$ are given by $$(F \circ \phi)_{a \ge b} = (F \circ \phi)(* : a \ge b) = F(* : \phi(a) \ge \phi(b)) = F_{\phi(a) \ge \phi(b)}$$ for all $a,b \in \mathcal{L}'$.
    We verify that $F \circ \phi : \mathcal{L}' \rightarrow \SET$ is a sheaf for any sheaf $F : \mathcal{L} \rightarrow \SET$. \smallbreak
    Let $(a_i \in \mathcal{L}')_{i \in I}$ be a cover of $a \in \mathcal{L}'$. As $\phi$ preserves joins, $(\phi(a_i) \in \mathcal{L})_{i \in I}$ is a cover of $\phi(a) \in \mathcal{L}$. Any compatible family $(x_i \in F \circ \phi(a_i))_{i \in I}$ in the context of $F \circ \phi$ is also compatible in the context of $F$, because $\phi$ preserves binary meets. \smallbreak
    The sheaf condition of $F$ grants us a unique $x \in F(\phi(a))$ with $F_{\phi(a) \ge \phi(a_i)}(x) = x_i$. This is also the unique element needed for the sheaf condition of $F \circ \phi$, because $F_{\phi(a) \ge \phi(a_i)}(x) = (F \circ \phi)_{a \ge a_i}(x)$.
  \end{Proof}
\end{Lemma}

This construction provides the action on morphisms to make the process of assigning the category $\Sheaf{\mathcal{L}}$ to the locale $\mathcal{L}$ functorial as well.

\begin{Lemma} \label{Def:SheafFunctor} There is a functor $\SheafFunctor : \Loc \rightarrow \Cat$ that assigns
  \begin{itemize}
  \item a locale $\mathcal{L}$ the category $\Sheaf{\mathcal{L}}$,
  \item a morphism of locales $\phi : \mathcal{L} \rightarrow \mathcal{L}'$ the morphism from \cref{Def:GeometricMorphism}.
  \end{itemize}
  \begin{Proof} We verify the functor laws.
    $$(\Sheaf{\psi} \circ \Sheaf{\phi})(F) = \Sheaf{\psi}(F \circ \phi) = F \circ \phi \circ \psi = \Sheaf{\psi \circ \phi}(F)$$
    \begin{align*}
      (\Sheaf{\psi} \circ \Sheaf{\phi})(\alpha : F \Rightarrow F')_a &= \Sheaf{\psi}(\Sheaf{\phi}(\alpha : F \Rightarrow F'))_a\\
                                            &= \Sheaf{\phi}(\alpha : F \Rightarrow F')_{\phi(\psi(a))}\\
                                            &= \alpha_{\psi \circ \phi(a)} = (\Sheaf{\psi \circ \phi})(\alpha : F \Rightarrow F')_a
    \end{align*}
  \end{Proof}
\end{Lemma}

Functors preserve isomorphisms, so we directly obtain an intuitive result: Isomorphic locales have isomorphic sheaf categories.

\begin{Corollary} \label{Cor:SheafIso} An isomorphism of locales $i : \mathcal{L} \rightarrow \mathcal{L}'$ induces an isomorphism of categories $\Sheaf{i} : \Sheaf{\mathcal{L}} \rightarrow \Sheaf{\mathcal{L}'}$.
  
\end{Corollary}


\clearpage
\section{Valued Sets} \label{Sec:ValuedSets}
As mentioned in the introduction, McInnes et al.\,\cite{UMAP} use the metric realization functor from Spivak's draft \cite[Section~3]{Spivak2009METRICRO}. The domain of this functor is a variant of simplicial fuzzy sets. Spivak \cite[Section~1]{Spivak2009METRICRO} defines different variants of fuzzy sets and cites Barr \cite{BarrFuzzy} for them. However, Barr \cite{BarrFuzzy} introduces multiple variants of these categories and Spivak \cite{Spivak2009METRICRO} uses two equivalent special cases of general categories that Barr only mentions at the end in the paragraph after \cite[Theorem~6]{BarrFuzzy}. Many minor errors and inconsistencies in Spivak's draft \cite{Spivak2009METRICRO} make this difficult to follow.

Spivak attempts to identify the simplicial objects over one of the two chosen categories of fuzzy sets with an equivalent category. He only sketches this out in a paragraph after \cite[Definition~1.4]{Spivak2009METRICRO} and gaps remain in the argument, see \cref{Sec:Issues}~(\ref{Issue2}).

McInnes et al. \cite{UMAP} cite Spivak \cite{Spivak2009METRICRO} and Barr \cite{BarrFuzzy}, but reproduce almost every issue in \cite{Spivak2009METRICRO} and also fail to specify which special cases from Barr \cite{BarrFuzzy} are used.

We will motivate and summarize the first general variant of these categories from the paragraph after \cite[Theorem~6]{BarrFuzzy} in \cref{SubSec:ClassicalValuedSets}. Unlike Barr, we will refer to the general case as classical valued sets (\cref{Def:ClassicalValSets}) to differentiate it from the special case that Spivak uses, which we will also call classical fuzzy sets in \cref{Def:ClassicalValSetsSpecialCases}~(1), like Spivak \cite[Definition~1.1]{Spivak2009METRICRO}.

We also motivate and summarize the second general, sheaf-theoretic variant of this category from the paragraph after \cite[Theorem~6]{BarrFuzzy} in \cref{SubSec:SheafValuedSets}. We will again refer to this category as valued sets (\cref{Def:SheafValuedSets}), whereas Barr calls them fuzzy sets. We refer to the special case that Spivak uses as fuzzy sets, see \cref{Def:SheafValSetsSpecialCases}~(1). We also provide an alternative proof for the equivalence of both categories from Barr \cite[Theorem~3]{BarrFuzzy}. Barr only provides one of the two functors (\cref{Def:M}) of the equivalence and proves that it is fully faithful and essentially surjective. We contribute the other functor in \cref{Def:C} and establish the equivalence with both functors and natural isomorphisms in \cref{Thm:EquivCM}. Sketches of the object map of the functor in \cref{Def:C} can already be found in \cite[Section~1]{Spivak2009METRICRO}. We need both functors later to compute the actions of the metric realization and related functors in \cref{Subsection:ClassicalMetricRealization}.

We also define the different variants of simplicial objects of valued sets and explicitly construct the equivalence of categories between them that Spivak only sketches after \cite[Definition~1.4]{Spivak2009METRICRO}. We also fill the remaining gaps discussed in \cref{Sec:Issues}, issue \ref{Issue2}.

Spivak constructs his metric realization \cite[Section~3]{Spivak2009METRICRO} in terms of the special case of fuzzy sets. We will introduce classical variants in \cref{Def:ClassicalValSetsSpecialCases}~(2) and equivalent sheaf-theoretic variants in \cref{Def:SheafValSetsSpecialCases}~(2) as another special case of Barr's categories \cite{BarrFuzzy}, which we will call (classical) normed sets. These classical and sheaf-theoretic normed sets are equivalent to Spivak's special case of classical and sheaf-theoretic fuzzy sets as shown in \cref{Cor:ClassicalFuzzyNormIso} and \cref{Cor:FuzzyNormIso}. Constructing the metric realization in terms of normed sets simplifies computations in \cref{Subsection:ClassicalMetricRealization}.

As there is no established naming convention for all of these categories in the references \cite{Spivak2009METRICRO, BarrFuzzy, UMAP}, we provide a translation table in the index of this thesis.

\subsection{Classical Valued Sets} \label{SubSec:ClassicalValuedSets}
An ordinary set $X$ is fully defined by the membership predicate $\in$. Membership is binary: we either have $x \in X$ or $x \notin X$. A fuzzy set allows for different strengths of membership between these two extremes. A mathematical object $x$ could be a full member of strength $1$, a non-member of strength $0$, or anything in between.
We can encode this in regular set theory by equipping a set $X$ with a map $\mu_X : X \rightarrow (0,1]$ that assigns a membership strength to every element, where all elements with strength $0$ are simply not included in $X$.

\begin{Definition}
  A \defind{classical fuzzy set} $(X,\mu_X)$ is a set $X$ equipped with a map $\mu_X : X \rightarrow (0,1]$, called the \defind{membership map}. For an element $x \in X$, the value $\mu_X(x) \in (0,1]$ is called the \defind{membership strength} of $x$ in $(X,\mu_X)$.
\end{Definition}
To organize fuzzy sets into a category, we must define morphisms between fuzzy sets.
For a map $f : X \rightarrow Y$ between ordinary sets, whenever there is an element $x \in X$, there is also an element $f(x) \in Y$. An intuitive way to lift this idea to fuzzy sets $(X,\mu_X), (Y,\mu_Y)$ is to demand that elements $x \in X$ of a membership strength $\mu_X(x)$ are sent to elements with equal or greater membership strength.
\begin{Definition} Let $(X,\mu_X)$ and $(Y,\mu_Y)$ be classical fuzzy sets. A morphism of fuzzy sets $f : (X,\mu_X) \rightarrow (Y,\mu_Y)$ is a map $f : X \rightarrow Y$ such that $$\mu_Y(f(x)) \geq \mu_X(x) \quad \forall x \in X.$$
\end{Definition}

This is a specific type of classical valued set category. Other examples of valued sets are normed sets; these are what we get when we forget the structure of a normed vector space but retain the norm:
\begin{Definition} A \defind{classical normed set} $(X, \norm{-}_X)$ is a set $X$ equipped with a map $\norm{-}_X : X \rightarrow [0,\infty)$, called the \defind{norm}.
  
\end{Definition}

Morphisms between normed structures are usually defined to be the morphisms that are non-expansive with respect to the norms, so we also demand this.
\begin{Definition} A \defind{morphism of classical normed sets} from $(X, \norm{-}_X)$ to $(Y, \norm{-}_Y)$ is a map $f : X \rightarrow Y$ with $\norm{f(x)}_Y \leq \norm{x}_X$ for all $x \in X$.
\end{Definition}

The reason we define morphisms of normed sets to be non-expansive and morphisms of fuzzy sets to be expansive is only to fit the convention in the literature. For fuzzy sets, the convention is that elements must be sent to elements of greater than or equal strength as explained above. For normed sets we demand that they are compatible with established ``normed'' categories. For example, the category of normed vector spaces and non-expansive maps can be sent to the category of normed sets with a forgetful functor that only erases the vector space structure but keeps the norm.
\bigbreak

In general, we define classical valued sets as sets equipped with a map that takes values in a locale, with morphisms defined as maps compatible with the ordering of the locale.
\begin{Definition} \label{Def:ClassicalValSets} Let $(\mathcal{L},\leq)$ be a locale.
  The category \defind{$\CVal{\mathcal{L},\leq}$} of classical $(\mathcal{L},\leq)$-valued sets consists of
  \begin{itemize}
  \item \defind{classical $(\mathcal{L},\leq)$-valued sets} $(X, \mu_X)$ as objects, where $X$ is a set and $\mu_X : X \rightarrow \mathcal{L}_{\neq\bot}$,
  \item \defind{morphisms of $(\mathcal{L},\leq)$-valued sets} from $(X,\mu_X)$ to $(Y,\mu_Y)$: maps $f : X \rightarrow Y$ with $\mu_X(x) \leq \mu_Y(f(x))$ for all $x \in X$.
  \end{itemize}
\end{Definition}

This definition is equivalent to $\text{Fuz}_0(L)$ from the paragraph after \cite[Theorem~6]{BarrFuzzy}. We exclude the bottom element $\bot$ directly by specifying $\mathcal{L}_{\neq\bot}$ as the set of possible values, while Barr demands that the inverse image of the bottom element, which he notes as 0, is empty.

We can then recover the specific example categories described above from this definition.
\begin{Definition} \label{Def:ClassicalValSetsSpecialCases} \quad
  \begin{enumerate}
  \item \label{Def:ClassicalValSetsSpecialCases:1} For the locale $([0,1],\leq)$, we obtain the \defind{category of classical fuzzy sets}
    \begin{center}
      \defind{$\CFuz$} $:= \CVal{[0,1],\leq}$.
    \end{center}
  \item For the locale $([0,\infty],\geq)$, we obtain the \defind{category of classical normed sets}
    \begin{center}
      \defind{$\CNSet$} $:= \CVal{[0,\infty],\geq}$.
    \end{center}
  \end{enumerate}
\end{Definition}

The category of classical fuzzy sets in \cref{Def:ClassicalValSetsSpecialCases} (1) coincides with the one in Spivak \cite[Definition 1.1]{Spivak2009METRICRO}.

We conclude this section by investigating how classical valued sets with values in different locales are related. We would expect that isomorphic locales induce isomorphic categories of classical valued sets. This is the case.
However, not every morphism of locales grants us a functor between the respective classical valued sets, because an arbitrary morphism of locales might map every element to $\bot$, which we have excluded in the definition of the value map. Other morphisms that only send $\bot$ to $\bot$ can also induce functors between categories that are not an equivalence, but we will only need the claim for isomorphisms. 

\begin{Lemma} \label{Def:IsoCVal} An isomorphism of locales $\phi : \mathcal{L} \rightarrow \mathcal{L}'$ induces a functor that is an isomorphism of categories
  $$\CVal{\phi} : \CVal{\mathcal{L}'} \rightarrow \CVal{\mathcal{L}}$$
  defined by
  \begin{align*}
    &\CVal{\phi}(X,\mu_X) = (X,\phi \circ \mu_X)\\
    &\CVal{\phi}(f : (X,\mu_X) \rightarrow (Y,\mu_Y)) = f : (X,\phi \circ\mu_X) \rightarrow (Y,\phi \circ\mu_Y).
  \end{align*}
  \begin{Proof}
    Note that the composite $\phi \circ \mu_X : X \rightarrow \mathcal{L}_{\neq \bot}$ is well defined, because a morphism of locales $\phi : \mathcal{L} \rightarrow \mathcal{L}'$ is a map $\phi : \mathcal{L}' \rightarrow \mathcal{L}$ in the opposite direction and due to \cref{Rem:IsoBottom}.

    The fact that $\phi$ is monotone ensures that $\CVal{\phi}(f)$ is a morphism of $\mathcal{L}$-valued sets for every morphism of $\mathcal{L}'$-valued sets $f$.

    It is easy to see that the induced functor is an isomorphism.
  \end{Proof}
\end{Lemma}

We do not want to claim novelty for this isomorphism of categories. We will see in \cref{SubSec:SheafValuedSets} that if $\mathcal{L}$ is totally connected, the category $\CVal{\mathcal{L}}$ is equivalent to a subcategory of $\Sheaf{\mathcal{L}}$. Sheaf theory tells us that the categories $\Sheaf{\mathcal{L}}$ and $\Sheaf{\mathcal{L}'}$ are isomorphic for isomorphic locales $\mathcal{L}$ and $\mathcal{L}'$ \cite[Chapter~IX, Section~5, Proposition~5]{maclane1992sheaves}.

\begin{Corollary} \label{Cor:ClassicalFuzzyNormIso} The category of classical fuzzy sets and the category of classical normed sets from \cref{Def:ClassicalValSetsSpecialCases} are isomorphic.
  \begin{Proof}
    The isomorphism $i : ([0,\infty],\le) \rightarrow ([0,1], \ge)$ from \cref{Example:MorphismsOfLocales} induces the isomorphism of categories $\CVal{i} : \CVal{([0,1], \ge)} \rightarrow \CVal{([0,\infty],\le)}$.
  \end{Proof}
\end{Corollary}

\subsection{Valued Sets as Sheaves} \label{SubSec:SheafValuedSets}
For a totally connected locale $\mathcal{L}$, classical $\mathcal{L}$-valued sets can be viewed as sheaves. Given a classical $\mathcal{L}$-valued set $(X,\mu_X)$, we can describe it as the family of sets $(X^{\ge a})_{a \in \mathcal{L}_{\neq\bot}}$ where $$X^{\ge a} = \{\,x \in X \, \mid \, \mu_X(x) \ge a \,\}$$ only contains the elements with a value of at least $a$. We just have to add the inclusion maps as restrictions and we have a sheaf.
\begin{Theorem} \label{Def:M} \cite[Theorem 3]{BarrFuzzy} Let $\mathcal{L}$ be a totally connected locale. There is a functor $\M : \CVal{\mathcal{L}} \rightarrow \Sheaf{\mathcal{L}}$ defined by
  \begin{align*}
    \M&(X,\mu_X)(a) =
       \begin{cases}
         \{ * \}  & a = \bot\\
          X^{\ge a} & a \neq \bot
       \end{cases} \\
    \M&(X,\mu_X)_{a \ge b}(x) =
       \begin{cases}
         * & b = \bot\\
         x & b \neq \bot
       \end{cases}\\
    \M&(f : (X,\mu_X) \rightarrow (Y,\mu_Y))_a =
       \begin{cases}
         * \mapsto * & a = \bot\\
         f|_{\M(X,\mu_X)(a)}^{\M(Y,\mu_Y)(a)} & a \neq \bot
       \end{cases}\\
  \end{align*}
  \begin{Proof}
    \begin{itemize}
    \item $\M(X,\mu_X)_{a \ge b} : \M(X,\mu_X)(a) \rightarrow \M(X,\mu_X)(b)$ is well defined:
      \smallbreak
      Let $a \ge b \neq \bot$ and $x \in \M(X,\mu_X)(a)$. Then $\mu_X(x) \ge a \ge b$, thus $x \in \M(X,\mu_X)(b)$.\\
      It is easy to see that these maps satisfy the functor laws.
    \item $\M(X,\mu_X)$ satisfies the sheaf condition:
      \smallbreak
      Let $(a_i \in \mathcal{L})_{i\in I}$ be a cover of $a \in \mathcal{L}$.
      \begin{itemize}
        \item For $a = \bot$, the sheaf condition is satisfied by \cref{Rem:SheafEmptyCover}.
        \item For $a \neq \bot$, let $(x_i)_{i\in I}$ be compatible. The compatibility condition
          \[ \M(X,\mu_X)_{a_i\ge a_i \, \wedge \, a_j}(x_i) = \M(X,\mu_X)_{a_j\ge a_i \, \wedge \, a_j}(x_j)  \quad \forall i,j \in I \]
          reduces to $x_i = x_j \; \forall i,j \in I$ with $a_i,a_j \neq \bot$, because we have $a_i \, \wedge \, a_j \neq \bot$ due to the total connectedness of $\mathcal{L}$.
          So all $x_i$ where $a_i \neq \bot$ coincide as one $x$.
          This $x$ is the unique element with $\M(X,\mu_X)_{a\ge a_i}(x) = x_i$ for all $i \in I$ with $a_i \neq \bot$. For all other $x_i = * \in \M(X,\mu_X)(\bot)$ we have $\M(X,\mu_X)_{a\ge \bot}(x) = * = x_i$ as well.
      \end{itemize}
    \item $\M(f)$ is a natural transformation:
      \smallbreak
      Let $f : (X,\mu_X) \rightarrow (Y,\mu_Y)$ and $a \ge \bot$. The valued set morphism condition of $f$ ensures that $f|_{\M(X,\mu_X)(a)}^{\M(Y,\mu_Y)(a)}$ is well defined.
      Naturality trivially follows as the restrictions for $a \ge \bot$  are just inclusions. For $a = \bot$ this is trivially true as well, as there is just one morphism that maps from the one element set to itself.
    \end{itemize}
  \end{Proof}
\end{Theorem}

By \cref{Lem:ExtendToSheaf} or \cref{Cor:EquivSheafStSheaf}, we could have defined $\M$ without the case distinctions. We include them in this case, but we will usually omit them later.

When we view a classical fuzzy set $(X,\mu_X)$ as a sheaf, in general we can only ever view a part of it by evaluating $\M(X,\mu_X)(a)$ which lets us see all elements of value greater than or equal than $a$. It might be tempting to evaluate $\M(X,\mu_X)(\bot)$ and expect to view all elements, but this simply gives us the singleton set $\{ * \}$ (\cref{Rem:SheafEmptyCover}).
\begin{Example} \label{Example:MActsOnFuzzy} \;
  \begin{itemize}
    \item For a finite classical fuzzy set $X = \{ x,y,z,w \}$ with
    \begin{align*}
      \mu_X& : X \rightarrow (0,1]\\
      \mu_X&(x) = \mu_X(y) = 1/3\\
      \mu_X&(z) = 1/2\\
      \mu_X&(w) = 1,
    \end{align*}
    we obtain the fuzzy set
    \[
      \M(X,\mu_X)(a) =
      \begin{cases}
        \{ * \}&a = 0\\
        \{ w,z,x,y \}&a \in (0,1/3]\\
        \{ w,z \}&a \in (1/3,1/2]\\
        \{ w \}&a \in (1/2,1].\\
      \end{cases}
    \]
    Here we have $\M(X,\mu_X)(a) = X$ for any $a \in (0,1/3]$.
  \item For an infinite classical fuzzy set $Y = \mathbb{N}$
    with
    \begin{align*}
      \mu_Y& : Y \rightarrow (0,1]\\
      \mu_Y&(n) = 1/n,
    \end{align*}
    we obtain the fuzzy set
    \[
      \M(Y,\mu_Y)(a) =
      \begin{cases}
        \{ * \}&a = 0\\
        \cdots &\\
        \{ 1,2,3,4 \}&a \in (1/5,1/4]\\
        \{ 1,2,3 \}&a \in (1/4,1/3]\\
        \{ 1, 2 \}&a \in (1/3,1/2]\\
        \{ 1 \}& a \in (1/2,1].
      \end{cases}
    \]
    In this case we have $\M(Y,\mu_Y)(a) \subsetneq Y$ for every $a$.
  \end{itemize}
\end{Example}

Not every sheaf can be viewed as a classical valued set because arbitrary sheaves allow for restriction maps to be non-injective: If $S \in \Sheaf{\mathcal{L}}$, there may be distinct elements $x \neq y$ in $S(a)$ which restrict to the same element $S_{a\ge b}(x) = S_{a\ge b}(y)$ for some $b \in \mathcal{L}$. So $x,y$ are ``not equal at membership strength $a$'' but are ``equal at strength $b$''. We can avoid this by only considering sheaves with injective restrictions.
\begin{Definition} \label{Def:SheafValuedSets} Let $\mathcal{L}$ be a totally connected locale.
  \begin{itemize}
  \item A sheaf $S : \mathcal{L}^\op \rightarrow \SET$ is called an \defind{$\mathcal{L}$-valued set} if the restriction map $S_{a \ge b} : S(a) \rightarrow S(b)$ is injective for all $a \ge b \neq \bot$.
  \item The \defind{category of $\mathcal{L}$-valued sets} \defind{$\Val{\mathcal{L}}$} is the full subcategory of $\Sheaf{\mathcal{L}}$ with $\mathcal{L}$-valued sets as objects.
  \end{itemize}
  From now on, for an $\mathcal{L}$-valued set $S$, we denote $S^{\ge a} := S(a)$ for $a \neq \bot$.
\end{Definition}

\begin{Corollary} \label{Cor:M} The functor $\M : \CVal{\mathcal{L}} \rightarrow \Sheaf{\mathcal{L}}$ defines a functor $\M : \CVal{\mathcal{L}} \rightarrow \Val{\mathcal{L}}$ into valued sets.
  \begin{Proof}
    The restriction maps $\M(X,\mu_X)_{a \ge b}$ are simply inclusions for $\bot \neq b \le a$ and thus trivially injective.
  \end{Proof}  
\end{Corollary}

\begin{Remark} \label{Rem:SheafValuedSets} Equivalently, we can view $\Val{\mathcal{L}}$ as the subcategory of $\SheafSt{\mathcal{L}}$ of sheaves where {\bf all} restrictions are injective, see \cref{Cor:EquivSheafStSheaf}.
  
\end{Remark}

The category $\Val{\mathcal{L}}$ as we define it in \cref{Def:SheafValuedSets} is identical to Mon($\mathcal{L}$) from \cite[Section~2]{BarrFuzzy}.

Barr showed that the categories $\CVal{\mathcal{L}}$ and $\Val{\mathcal{L}}$ are equivalent if $\mathcal{L}$ is totally connected. The specific variant of the equivalence we use in this thesis is mentioned in the paragraph just after \cite[Theorem~6]{BarrFuzzy}. While the category $\Val{\mathcal{L}}$ might not be a topos, $\Sheaf{\mathcal{L}}$ always is. It is the effective completion of $\Val{\mathcal{L}}$. So if one needs all operations available in a topos, $\Sheaf{\mathcal{L}}$ is the correct category to use according to Barr \cite[Section~1]{BarrFuzzy}. In this thesis we will not use topos-theoretic constructions.

We will now construct the equivalence between the categories $\CVal{\mathcal{L}}$ and $\Val{\mathcal{L}}$. Barr only constructs the functor $\M : \CVal{\mathcal{L}} \rightarrow \Val{\mathcal{L}}$ from \cref{Cor:M} and proves that it is fully faithful and essentially surjective \cite[Theorem~3]{BarrFuzzy}. We will also construct the functor in the other direction and all natural isomorphisms involved in the equivalence explicitly, as we need them later to translate back and forth between these categories.

Before we construct the functor from $\Val{\mathcal{L}}$ to $\CVal{\mathcal{L}}$, we will introduce a few definitions that help to characterize $\mathcal{L}$-valued sets.
For an $\mathcal{L}$-valued set $S : \mathcal{L}^\op \rightarrow \SET$, recall that $S^{\ge a} := S(a)$ is the set of all elements that have a value greater than or equal to $a$. If $b \ge a$, then the elements $S^{\ge b}$ can be embedded into $S^{\ge a}$ via the restriction maps. If we have an element $x \in S^{\ge a}$ and some other element $y \in S^{\ge b}$ with $b > a$ that restricts to $x$, then $x$ must have some larger value than $a$. But if there is no such element, then $x$ must have value $a$. We will denote the set of such elements by $S^{=a}$.

\begin{Definition} \label{Def:^=} Let $\mathcal{L}$ be a totally connected locale, let $S : \mathcal{L}^\op \rightarrow \SET$ be a valued set and let $a \ne \bot$. We define
  \[ S^{=a} = \{ x \in S^{\ge a} \; | \; \text{there are no } b > a, y \in S^{\ge b} \text{ with } S_{b \ge a}(y) = x \}. \]
\end{Definition}

\begin{Example} Consider again the fuzzy sets $X$ and $Y$ from \cref{Example:MActsOnFuzzy}.
  Then we obtain
  \[
    \M(X,\mu_X)^{= a} =
    \begin{cases}
      \{ x,y \}&a = 1/3\\
      \{ z \}&a = 1/2\\
      \{ w \}& a = 1\\
      \emptyset     & \text{else},
    \end{cases}
  \]
  and
  \[
    \M(Y,\mu_Y)^{= a} =
    \begin{cases}
      \{ n \}&a = 1/n\\
      \emptyset     & \text{else}.
    \end{cases}
  \]
\end{Example}

Given an element $x \in S^{\ge a}$ not necessarily in $S^{=a}$, we can find the true level of $x$, which is the highest level containing an element that restricts to $x$.
\begin{Definition} \label{Def:mu} Let $\mathcal{L}$ be a totally connected locale and let $S : \mathcal{L}^\op \rightarrow \SET$ be an $\mathcal{L}$-valued set. The map $\mu^{a}_{S} : S^{\ge a} \rightarrow \mathcal{L}_{\neq \bot}$ is defined by
  $$\mu^{a}_{S}(x) = \bigvee \{ \, b \in \mathcal{L}_{\neq \bot} \; | \; \text{$b \ge a$ and there is $y \in S^{\ge b}$ with $S_{b\ge a}(y) = x$ } \, \}.$$
\end{Definition}
We can define a map that sends elements $x \in S^{\ge a}$ at any level $a$ to their true level $S^{=\mu^{a}_S(x)}$. Since we do not know this level a priori, we must map into the coproduct of all possible $S^{=b}$, except for $\bot$, as $S^{=\bot} = \{ * \}$.
\begin{Definition} Let $\mathcal{L}$ be a totally connected locale and let $S : \mathcal{L}^\op \rightarrow \SET$ be a valued set. The map
  $$\uparrow^a_S : S^{\ge a} \rightarrow \{\, \iota_b(y) \, \mid \, y \in S^{=b} \text{ and } b \ge a \,\} \subseteq \coprod_{b \in \mathcal{L}_{\neq \bot}}S^{=b}$$
  is defined by
  $$\uparrow^a_S(x) = \iota_{\mu^{a}_S(x)}(y), \qquad \text{where } y \in S^{=\mu^{a}_S(x)} \text{ with } S_{\mu^{a}_S(x) \ge a}(y) = x.$$
\end{Definition}

Now we can define the other functor in the equivalence from $\Val{\mathcal{L}}$ to $\CVal{\mathcal{L}}$. We assign to an $\mathcal{L}$-valued set the coproduct of all its levels. A morphism of valued sets $\alpha : S \Rightarrow T$ is assigned a map that applies $\alpha_a$ at each level $a$. As the result might end up at a different level, we send it to its correct level through $\uparrow^a_T$.

The general idea for the object map of the following functor $\C$ is sketched out in \cite[Section~1]{Spivak2009METRICRO}.

\begin{Lemma} \label{Def:C} Let $\mathcal{L}$ be a totally connected locale. There is a functor $\C : \Val{\mathcal{L}} \rightarrow \CVal{\mathcal{L}}$ defined by
  \begin{align*}
    \C&(S) = \left(X,\mu_S\right)\text{ where}\\
     &X= \coprod_{a \in \mathcal{L}_{\neq \bot}}S^{=a}\\
     &\mu_S : X \rightarrow \mathcal{L}_{\neq \bot}\\
     &\mu_S(\iota_a(x)) = a\\
    \C&(\alpha : S \Rightarrow T) \circ \iota_a(x) = \uparrow^a_T  \circ \alpha_{a}(x).
  \end{align*}
  \begin{Proof} We verify the functor laws. For the identity $\id{S}$, we have
    \begin{align*}
      \C(\id{S}) \circ \iota_a(x)
      &= \uparrow^a_S  \circ \id{S^{\ge a}}(x)\\
      &= \uparrow^a_S(x)\\
      &= \iota_{\mu^{a}_S(x)}(y) \qquad &\text{where } \mu^{a}_S(x) = a\\
      &                      &\text{ with } S_{\mu^{a}_S(x) \ge a}(y) = S_{a \ge a}(y) = y = x\\
      &= \iota_a(x).
    \end{align*}
    Here we have $\mu^{a}_S(x) = a$ as $x \in S^{=a}$ already is at its true level, no other element restricts to $x$.
    For two morphisms $\alpha : S \Rightarrow T, \beta : T \Rightarrow U$ of valued sets, we must perform a computation with tedious bookkeeping:
    \begin{align*}
       & \C(\beta) \circ \C(\alpha) \circ \iota_a(x)\\
      =& \C(\beta) \circ \uparrow^a_T (\alpha_{a}(x))\\
      =& \C(\beta)(\iota_b(y))                  &\text{where } b = \mu^{a}_T(\alpha_a (x)) \ge a \text{ and } y \in T^{=b} \\
       &                               &\text{unique with } T_{b \ge a}(y) = \alpha_a (x)  \\
      =& \uparrow^b_U  \circ \beta_b(y)\\
      =& \iota_c(z)                        &\text{where } c = \mu^{b}_U(\beta_b (y)) \ge b \text{ and } z \in U^{=c} \\
       &                               &\text{unique with } U_{c \ge b}(z) = \beta_b(y)  \\
      \stackrel{(*)}{=}& \iota_d(w)                        &\text{where } d = \mu^{a}_U(\beta_a \circ \alpha_a (x)) \ge a \text{ and } w \in U^{=d} \\
       &                               &\text{unique with } U_{d \ge a}(w) = \beta_a \circ \alpha_{a}(x)  \\
      =& \uparrow^a_U (\beta_a \circ \alpha_{a}(x))\\
      =& \C(\beta \circ \alpha) \circ \iota_a(x),\\
    \end{align*}
    where we used the sheaf condition in $(*)$. For this, we compute
    \begin{align*}
         U_{c \ge a}(z)
      &= U_{b\ge a} \circ U_{c \ge b}(z)\\
      &= U_{b\ge a} \circ \beta_b(y)\\
      &= \beta_a \circ T_{b\ge a}(y)\\
      &= \beta_a \circ \alpha_a(x) = U_{d \ge a}(w).
    \end{align*}
    Then by injectivity of the restrictions, we obtain
    \[ U_{c \ge c \land d}(z) = U_{d \ge c\land d}(w), \]
    so there is a unique element in $c \lor d$ that restricts to $z$ and $w$, but the memberships $z \in U^{=c}$, $w \in U^{=d}$ imply that there are no larger levels with elements that restrict to $z$ and $w$. Thus $c$ and $d$ must be the same and $z$ and $w$ must be equal elements at that level. 
  \end{Proof}
\end{Lemma}

Having defined the functors in both directions in \cref{Cor:M} and \cref{Def:C}, we can prove that they indeed define an equivalence.

\begin{Theorem} \label{Thm:EquivCM} Let $\mathcal{L}$ be a totally connected locale. The functors $\M : \CVal{\mathcal{L}} \rightarrow \Val{\mathcal{L}}$ and $\C : \Val{\mathcal{L}} \rightarrow \CVal{\mathcal{L}}$ define an equivalence of categories between $\CVal{\mathcal{L}}$ and $\Val{\mathcal{L}}$.
  \begin{Proof}
    We construct natural isomorphisms $\eta : \Id{\CVal{\mathcal{L}}} \Rightarrow \C\M$ and $\epsilon : \M\C \Rightarrow \Id{\Val{\mathcal{L}}}$.
    For the components
    \begin{align*}
      \eta&_{(X,\mu_X)} : {(X,\mu_X)} \rightarrow \left(\coprod_{a \in \mathcal{L}_{\neq\bot}} \{ x \in X \; | \; \mu_X(x) = a \}, \; \mu_{\M(X,\mu_X)}\right)\\
       &\text{where } \mu_{\M(X,\mu_X)} \circ \iota_a(x) = a = \mu_X(x)\\
    \end{align*}
    it is obvious that we must set
    \begin{align*}
      \eta&_{(X,\mu_X)}(x) = \iota_{\mu_X(x)}(x)\\
      \eta&_{(X,\mu_X)}^{-1}(\iota_{a}(x)) = x.
    \end{align*}
    To verify that this is indeed an isomorphism, we compute:
    \begin{align*}
      \eta^{-1}_{(X,\mu_X)} \circ \eta_{(X,\mu_X)}(x) &= \eta^{-1}_{(X,\mu_X)}(\iota_{\mu_X(x)}(x)) = x\\
      \eta_{(X,\mu_X)} \circ \eta^{-1}_{(X,\mu_X)}(\iota_{a}(x)) &= \eta_{(X,\mu_X)}(x) = \iota_{\mu_X(x)}(x) = \iota_{a}(x).
    \end{align*}
    To verify the naturality of $\eta$ and $\eta^{-1}$, it suffices to check one of the directions of the isomorphism, so we check $\eta$. The naturality square simplifies to
    \begin{center}
      \begin{tikzcd}
        (X,\mu_X) \arrow[r, "\eta_{(X,\mu_X)}"] \arrow[d, "f"'] & \left(\coprod_{a \in \mathcal{L}_{\neq\bot}} \{ x \in X \; | \; \mu_X(x) = a \}, \; \mu_{\M(X,\mu_X)}\right) \arrow[d, "\C\M(f)"] \\
        (Y,\mu_Y) \arrow[r, "\eta_{(Y,\mu_Y)}"]                 & \left(\coprod_{a \in \mathcal{L}_{\neq\bot}} \{ y \in Y \; | \; \mu_Y(y) = a \}, \; \mu_{\M(Y,\mu_Y)}\right),               
      \end{tikzcd}
    \end{center}
    so we compute
    \begin{align*}
      &\qquad \;\C\M(f) \circ \eta_{(X,\mu_X)}(x)\\
      \overset{(\text{def } \eta)}&{=} \; \C\M(f) \circ \iota_{\mu_X(x)}(x)\\
      \overset{(\text{def } \C)}&{=} \; \uparrow_{\M(Y,\mu_Y)}^{\mu_X(x)} \M(f)_{\mu_X(x)}(x)\\
      \overset{(\text{def } \uparrow)}&{=} \; \iota_{\mu^{\mu_X(x)}_{\M(Y,\mu_Y)}(f(x))} (y) \qquad &&\text{where } \M(Y,\mu_Y)_{(\mu^{\mu_X(x)}_{\M(Y,\mu_Y)} \circ f(x)) \ge \mu_X(x)}(y) = f(x) \numberthis{eq:whereMf}\\
      \overset{(\text{def } \M)}&{=} \; \iota_{\mu_Y(f(x))} (y)\\
      \overset{(*)}&{=} \; \iota_{\mu_Y(f(x))} (f(x))\\
      \overset{(\text{def } \eta)}&{=} \; \eta_{(Y,\mu_Y)}(f(x)),
    \end{align*}
    where $y = f(x)$ in $(*)$ is due to
    $$y \overset{(\text{def } \M)} = \M(Y,\mu_Y)_{(\mu^{\mu_X(x)}_{\M(Y,\mu_Y)} \circ f(x)) \ge \mu_X(x)}(y) \overset{\eqref{eq:whereMf}}{=} y.$$
    The components of $\epsilon : \M\C \Rightarrow \Id{\Val{\mathcal{L}}}$ are natural isomorphisms
    $$\epsilon_S : \M(\coprod_{a \in \mathcal{L}_{\neq \bot}}S^{=a},\mu_S) \Rightarrow S$$
    with components
    \begin{align*}
                    (\epsilon&_S)_b : \M(\coprod_{a \in \mathcal{L}_{\neq \bot}}S^{=a},\mu_S)^{\ge b} \Rightarrow S^{\ge b}\\
      \text{ where }(\epsilon&_S)_b : \{\, \iota_a(x) \, \mid \, x \in S^{=a} \text{ and } a \ge b \,\} \rightarrow S^{\ge b} && b \neq \bot\\
                    (\epsilon&_S)_b : \{\, * \,\} \rightarrow \{\, * \,\}.                                       && b = \bot\\
    \end{align*}
    We set
    \begin{align*}
      (\epsilon&_S)_b(\iota_a(x)) = S_{a \ge b}(x)\\
      (\epsilon&_S^{-1})_b(x) = \uparrow^b_S(x)\\
    \end{align*}
    and again verify that this is an isomorphism
    \begin{align*}
      (\epsilon_S)_b \circ (\epsilon_S^{-1})_b(x) \quad
      \overset{(\text{def } \epsilon^{-1})}{=}& \quad (\epsilon_S)_b(\uparrow^b_S(x))\\
      \overset{(\text{def } \uparrow)}{=}& \quad (\epsilon_S)_b(\iota_{\mu^b_S(x)}(y)) && \text{where } S_{\mu^b_S(x) \ge b}(y) = x \numberthis{eq:whereS1}\\
      \overset{(\text{def } \epsilon)}{=}& \quad S_{\mu^b_S(x) \ge b}(y) \quad \stackrel{\eqref{eq:whereS1}}{=} \quad x
    \end{align*}
    \begin{align*}
      (\epsilon_S^{-1})_b \circ (\epsilon_S)_b (\iota_a(x))
      \overset{(\text{def } \epsilon)}&{=} \quad (\epsilon_S^{-1})_b(S_{a \ge b}(x))\\
      \overset{(\text{def } \epsilon^{-1})}&{=}  \quad \uparrow^b_S \circ S_{a \ge b}(x)\\
      \overset{(\text{def } \uparrow)}&{=} \quad \iota_{\mu^b_S(S_{a \ge b}(x))}(y) \quad\text{where } S_{\mu^b_S(S_{a \ge b}(x)) \ge b}(y) = S_{a \ge b}(x) \numberthis{eq:whereS2}\\
      \overset{(*)}&{=} \quad \iota_{a}(y) \quad \stackrel{(**)}{=} \quad \iota_{a}(x),
    \end{align*}
    where $\mu^b_S(S_{a \ge b}(x)) = a$ in $(*)$ by \cref{Def:mu} and $x \in S^{=a}$ and where $x = y$ in $(**)$ is due to the injectivity of $S_{a \ge b}$ because we have
\[ S_{a \ge b}(y) = S_{\mu^b_S(S_{a \ge b}(x)) \ge b}(y) \overset{\eqref{eq:whereS2}}{=} S_{a \ge b}(x). \]

    For the naturality of $\epsilon_S$, we check the square
    \begin{center}
      \begin{tikzcd}
        \M\C(S) \arrow[d, "\M\C(S)_{b \ge c}"'] \arrow[r, "(\epsilon_S)_b"] & S^{\ge b} \arrow[d, "S_{b\ge c}"] \\
        \M\C(S) \arrow[r, "(\epsilon_S)_c"]                 & S^{\ge c}              
      \end{tikzcd}
    \end{center}
    by the computation
    \begin{align*}
      (\epsilon_S)_c \circ \M\C(S)_{b \ge c}(\iota_a(x))
      \overset{(\text{def } \M, \C)}&{=} (\epsilon_S)_c(\iota_a(x))\\
      \overset{(\text{def } \epsilon)}&{=} \; S_{a\ge c}(x) \\
      \overset{(\text{func } S)}&{=} S_{b\ge c} \circ S_{a\ge b}(x) \overset{(\text{def } \epsilon)}&{=} S_{b\ge c} \circ (\epsilon_S)_b (\iota_a(x)).
    \end{align*}
    Finally, for the naturality of $\epsilon$, we must check the square
    \begin{center}
      \begin{tikzcd}
        \M\C(S) \arrow[d, "\M\C(\alpha)"'] \arrow[r, "\epsilon_S"] & S \arrow[d, "\alpha"] \\
        \M\C(T) \arrow[r, "\epsilon_T"]                 & T              
      \end{tikzcd}.
    \end{center}
    Again, by a concrete computation, we have
    \begin{align*}
      & \qquad \; \; (\epsilon_T)_b \circ \M\C(\alpha)_b(\iota_a(x))\\
      \overset{(\text{def } \M,\C)}&{=} (\epsilon_T)_b \circ \uparrow^a_{T} \circ \alpha_a(x)\\
      \overset{(\text{def } \uparrow)}&{=} \; \; (\epsilon_T)_b \circ \iota_{\mu^a_T(\alpha_a(x))}(y) &&\text{where } T_{\mu^a_T(\alpha_a(x)) \ge a}(y) = \alpha_a(x) \numberthis{eq:whereT}\\
      \overset{(\text{def } \epsilon)}&{=} \;\; T_{\mu^a_T(\alpha_a(x)) \ge b}(y)\\
      \overset{(\text{func } T)}&{=} \; T_{a\ge b} \circ T_{\mu^a_T(\alpha_a(x)) \ge a}(y)\\
      \overset{\eqref{eq:whereT}}&{=} \;  \quad T_{a\ge b} \circ \alpha_a(x)\\
      \overset{(\text{nat } \alpha)}&{=} \;\; \alpha_b \circ S_{a \ge b}(x)\\
      \overset{(\text{def } \epsilon)}&{=} \;\;\alpha_b \circ (\epsilon_S)_b(\iota_a(x)).
    \end{align*}
  \end{Proof}
\end{Theorem}

Recall the special cases of valued sets from \cref{Def:ClassicalValSetsSpecialCases}. For each of them, we can  recover their equivalent sheaf-theoretic variants as a special case of the equivalence from \cref{Thm:EquivCM}.
\begin{Definition} \label{Def:SheafValSetsSpecialCases} \quad
  \begin{itemize}
  \item For the locale $([0,1],\leq)$, we define the \defind{category of fuzzy sets} as
    \begin{center}
      \defind{$\Fuz$} $:= \Val{[0,1],\leq}.$
    \end{center}
    This category is equivalent to $\CFuz$ by \cref{Thm:EquivCM}.
  \item For the locale $([0,\infty],\geq)$, we obtain the \defind{category of normed sets} as
    \begin{center}
      \defind{$\NSet$} $:= \Val{[0,\infty],\geq}.$
    \end{center}
    This category is equivalent to $\CNSet$ by \cref{Thm:EquivCM}.
  \end{itemize}
\end{Definition}

The category $\Fuz$ from \cref{Def:SheafValSetsSpecialCases} is identical to the sheaf-theoretic variant of fuzzy sets that Spivak uses in \cite{Spivak2009METRICRO}.

By \cref{Thm:EquivCM}, we know that $\Val{\mathcal{L}}$ is equivalent to $\CVal{\mathcal{L}}$. In \cref{Def:IsoCVal}, we proved that $\CVal{\mathcal{L}}$ is isomorphic to $\CVal{\mathcal{L}'}$ if the locales $\mathcal{L}$ and $\mathcal{L}'$ are isomorphic. We should thus expect an isomorphism of locales to induce a similar equivalence between the sheaf-theoretic variants $\Val{\mathcal{L}}$ and $\Val{\mathcal{L}'}$.

\begin{Lemma} An isomorphism of locales $\phi : \mathcal{L} \rightarrow \mathcal{L}'$ induces the isomorphism of categories $\Sheaf{\phi}|^{\Val{\mathcal{L}'}}_{\Val{\mathcal{L}}} : \Val{\mathcal{L}} \rightarrow \Val{\mathcal{L}'}$.
  \begin{Proof}
     We must verify that for every $\mathcal{L}$-valued set $F$, the sheaf $\Val{\phi}(F) = \Sheaf{\phi}(F) = F \circ \phi$ is an $\mathcal{L}'$-valued set.

     Let $F$ be an $\mathcal{L}$-valued set. Then every restriction map $F_{a \ge b}$ is injective for $a \ge b \neq \bot$.
     
     Let $a' \ge b' \neq \bot$. Then $(F \circ \phi)_{a' \ge b'} = F_{\phi(a') \ge \phi(b')}$ is injective, because $F$ is an $\mathcal{L}$-valued set, and because $\phi(b') \neq \bot$ by \cref{Rem:IsoBottom}.
  \end{Proof}
\end{Lemma}

\begin{Corollary} \label{Cor:FuzzyNormIso} The category of fuzzy sets and the category of normed sets from \cref{Def:SheafValSetsSpecialCases} are isomorphic.
  \begin{Proof}
    The isomorphism $i : ([0,\infty],\le) \rightarrow ([0,1], \ge)$ of locales from \cref{Example:MorphismsOfLocales} induces the isomorphism of categories $\Sheaf{i} : \Fuz \rightarrow \NSet$.
  \end{Proof}
\end{Corollary}

\subsection{Simplicial Valued Sets} \label{SubSec:SimplicialValuedSets}

Recall that the category of simplicial objects in $\mathcal{C}$ is the functor category $\mathcal{C}^{\Delta^\op}$ consisting of functors $S : \Delta^\op \rightarrow \mathcal{C}$ as objects and natural transformations between them as morphisms.

The Yoneda embedding $y : \Delta \rightarrow \SET^{\Delta^\op}$ from \cref{Def:Yoneda} is a functor into the category of simplicial sets. This allows us to construct functors that map simplicial sets into another category $\mathcal{E}$ as left Kan extensions of functors $F : \Delta \rightarrow \mathcal{E}$
\begin{Remark} \label{Rem:KanAlongYoneda} Every functor $F : \Delta \rightarrow \mathcal{E}$ into a cocomplete category guarantees us a left Kan extension by \cref{Cor:ColimitFormula}.
  \begin{center}
    \begin{tikzcd}
      \Delta \arrow[rr, "F"] \arrow[rd, "y"'] & {} \arrow[d, "\eta", Rightarrow, shift right] & \mathcal{E} \\
      & \SET^{\Delta^\op} \arrow[ru, "\Lan{y}{F}"', dashed]  &
    \end{tikzcd}
  \end{center}
\end{Remark}
Furthermore, these left Kan extensions along the Yoneda embedding are left adjoint to $R = \Hom{\mathcal{E}}{F(-)}{-} : \mathcal{E} \rightarrow \SET^{\Delta^\op}$ by \cref{Prop:KanExtYoneda}.

We cannot directly mirror this construction for simplicial objects in the category of classical $\mathcal{L}$-valued sets. One reason for this is that the Yoneda embedding $y : \Delta \rightarrow \SET^{\Delta^\op}$ does not map into a category of simplicial objects over $\CVal{\mathcal{L}}$, because $y$ does not equip the sets $y([n])_m$ with a value map. The functor $R$ also does not equip the hom sets $\Hom{\mathcal{E}}{F([n])}{-}$ with a value map.
These issues are not resolved when switching to simplicial objects in the equivalent category $\Val{\mathcal{L}}$.

To resolve this, we must mold the category of simplicial objects in $\Val{\mathcal{L}}^{\Delta^\op}$ into a shape that fits the Yoneda embedding. This amounts to the construction of yet another equivalence of categories: An object $S$ in $\Val{\mathcal{L}}^{\Delta^\op}$ is a functor $S : \Delta^\op \rightarrow \SET^{\mathcal{L}_{\neq\bot}^\op}$ satisfying additional conditions from \cref{Def:SheafValuedSets}. By uncurrying (\cref{Lem:Curry}), we can also view it as a functor $S : (\Delta\times\mathcal{L}_{\neq\bot})^\op \rightarrow \SET$ that satisfies the properly translated additional conditions.
This category has the Yoneda embedding $y : (\Delta\times\mathcal{L}_{\neq\bot}) \rightarrow \SET^{{(\Delta\times\mathcal{L}_{\neq\bot})}^\op}$ and admits left Kan extensions of functors $F : (\Delta\times\mathcal{L}_{\neq\bot}) \rightarrow \SET$ along it.

\begin{Definition} \label{Def:SimpValSet} \quad Let $\mathcal{L}$ be a totally connected locale.
  \begin{enumerate}
  \item A \defind{simplicial classical $\mathcal{L}$-valued set} is a simplicial object in $\CVal{\mathcal{L}}$.

    The \defind{category of simplicial classical $\mathcal{L}$-valued sets} $\CVal{(\mathcal{L},\le)}^{\Delta^\op}$ is the full subcategory of $\SET^{(\Delta\times\mathcal{L})^\op}$ with uncurried simplicial $\mathcal{L}$-valued sets as objects.

  \item An \defind{uncurried simplicial $\mathcal{L}$-valued set} is a functor $S : (\Delta\times\mathcal{L})^\op \rightarrow \SET$ such that $S([n],-) : \mathcal{L}^\op \rightarrow \SET$ is an $\mathcal{L}$-valued set for every object $[n]$ in $\Delta$.

    The \defind{category of uncurried simplicial $\mathcal{L}$-valued sets} \defind{$\SVal{\mathcal{L}}$} is the full subcategory of $\SET^{(\Delta\times\mathcal{L})^\op}$ with uncurried simplicial $\mathcal{L}$-valued sets as objects.
  \end{enumerate}
\end{Definition}

\begin{Remark} \label{Rem:SimpValSet} Similar to \cref{Rem:SheafValuedSets}, we can view $\SVal{\mathcal{L}}$ as the subcategory of functors $S : (\Delta\times\mathcal{L}_{\neq\bot})^\op \rightarrow \SET$ such that $S([n],-)$ is an $\mathcal{L}$-valued set in the sense of \cref{Rem:SheafValuedSets}.
  
\end{Remark}

This definition of the category of uncurried simplicial $\mathcal{L}$-valued sets is inspired by the second paragraph after \cite[Definition~1.4]{Spivak2009METRICRO}. As mentioned at the beginning of \cref{Sec:ValuedSets} and in \cref{Sec:Issues}, \cref{Issue2}, Spivak's category is more cumbersome to work with than ours because it requires a product Grothendieck topology on $\Delta \times \mathcal{L}$. It also lacks the injectivity condition from \cref{Def:SheafValuedSets}. We instead demand that $S([n],-) : \mathcal{L}^\op \rightarrow \SET$ is a sheaf-theoretic fuzzy set as in \cref{Def:SheafValuedSets}, which both restores the injectivity condition and eliminates the need for Grothendieck topologies entirely.

We now show that our definition indeed admits the desired equivalence of categories. This amounts to composing the equivalence from \cref{Thm:EquivCM} with the category-theoretic currying isomorphism from \cref{Lem:Curry}.
\begin{Lemma} \label{Thm:EquivCValSVal} Let $\mathcal{L}$ be totally connected. Then there is an equivalence between the categories $\CVal{\mathcal{L}}^{\Delta^\op}$ and $\SVal{\mathcal{L}}$. \smallbreak
    We define the functors $\SM$ and $\SC$ as composites of the previously constructed equivalences from \cref{Thm:EquivCM} and \cref{Lem:Curry}:
    \begin{center}
      \begin{tikzcd}[sep=small]
        \CVal{\mathcal{L}}^{\Delta^\op} \arrow[rr, "\M \circ -", bend left = 20] \arrow[rrrr, "\SM", bend left, shift left=3] & \text{\cref{Thm:EquivCM}}  &
        \Val{\mathcal{L}}^{\Delta^\op} \arrow[ll, "\C \circ -", bend left = 20] \arrow[rr, "\uncurry", bend left = 20] & \text{\cref{Lem:Curry}} & \SVal{\mathcal{L}} \arrow[ll, "\curry", bend left = 20]
        \arrow[llll, "\SC", bend left, shift left=3]
      \end{tikzcd}
    \end{center}
    where
    \begin{itemize}
    \item $\M : CVal{\mathcal{L}} \rightarrow \Val{\mathcal{L}}$ is the functor from \cref{Def:M},
    \item $\C : \Val{\mathcal{L}} \rightarrow \CVal{\mathcal{L}}$  is the functor from \cref{Def:C},
    \item and $\curry$ and $\uncurry$ are the functors defined in \cref{Lem:Curry}.
    \end{itemize}
    
    When we simplify the composites $\SM := \uncurry \circ (\M \circ -)$ and $\SC := (\C \circ -) \circ \curry$, we get
    \begin{align*}
      \SM&(S)([n],a)=\M(S_n)^{\ge a}                       &    \SC&(S)_n = \C(S([n],-))\\
      \SM&(S)(f,* : a \ge b) = \M(S(f))_b \circ \M(S)_{a \ge b}   &    \SC&(S)(g : [n] \rightarrow [m]) = \C(S(g,-))\\
      \SM&(\alpha : S \rightarrow T)_{[n],a} = \M(\alpha_n)_a                 &    \SC&(\alpha : S \Rightarrow T)_n = \C(\alpha_{n,-})
    \end{align*}
\end{Lemma}

Recall the classical special cases from \cref{Def:ClassicalValSetsSpecialCases} and equivalent sheaf-theoretic cases from \cref{Def:SheafValSetsSpecialCases}. For each of them, we obtain two equivalent categories of simplicial objects as a special case of \cref{Thm:EquivCValSVal}.
\begin{Definition} \label{Def:SimpValSetsSpecialCases} \quad
  \begin{enumerate}
  \item For the equivalent categories $\CFuz$ and $\Fuz$, the associated categories of simplicial objects are the \defind{category of simplicial classical fuzzy sets} $\CFuz^{\Delta^\op}$ and the \defind{category of uncurried simplicial fuzzy sets}
    \begin{center}
      \defind{$\SFuz$} $:= \SVal{[0,1],\leq}.$
    \end{center}
  \item For the equivalent categories $\CNSet$ and $\NSet$, the associated categories of simplicial objects are the \defind{category of simplicial classical normed sets} $\CNSet^{\Delta^\op}$ and the \defind{category of uncurried simplicial normed sets}
    \begin{center}
      \defind{$\SNSet$} $:= \SVal{[0,\infty],\geq}.$
    \end{center}
  \end{enumerate}
\end{Definition}

We finally derive an analogous result to \cref{Rem:KanAlongYoneda}. For this, we must first verify that the image of the Yoneda embedding lies in $\SVal{\mathcal{L}}$.
\begin{Lemma} \label{Lem:ValYonedaEmb} The Yoneda embedding $y : \Delta \times \mathcal{L}_{\neq\bot} \rightarrow \SET^{(\Delta \times \mathcal{L}_{\neq\bot})^\op}$ takes values in $\SVal{\mathcal{L}}$ and defines a functor $y : \Delta \times \mathcal{L}_{\neq\bot} \rightarrow \SVal{\mathcal{L}}$.
  \begin{Proof}
    We verify that $S = y([m],b)([n],-) : \mathcal{L}_{\neq\bot} \rightarrow \SET$ is a valued set for all $n,m \in \mathbb{N}_0$ and $b \in \mathcal{L}_{\neq\bot}$.
    We first compute the action of the restriction maps for $f : [n] \rightarrow [m]$ and $* : c \le a$.
    \begin{align*}
        S_{c \ge d}(f,* : c \le d) &= y([m],b)([n],-)(f,*)\\
                               &= \Hom{\Delta \times \mathcal{L}_{\neq\bot}}{(\id{[n]}, * : b \le c)}{([m],b)}(f,* : c \le d)\\
                               &= (f, * : b \le d)
    \end{align*}
    It is obvious that these maps are injective.
    For the sheaf condition, we use the criterion from \cref{Lem:ExtendToSheaf}.
    Let $(a_i)_{i \in I}$ be a cover of $a \in \mathcal{L}_{\neq\bot}$. Then $(a_i)_{i \in I}$ cannot be the empty cover. Now let $(f_i,*)_{i \in I}$ be a compatible family. Due to the compatibility condition and the restriction maps we have $\{f_i\}_{i \in I} = \{ f \}$, where $f$ is our unique element.
  \end{Proof}
\end{Lemma}

We can then apply \cref{Prop:KanExtYonedaSubcat}, which is specifically adapted to this setting.
\begin{Lemma} \label{Cor:SimplicialKanExt} Let $\mathcal{L}$ be a totally connected locale and let $F : \Delta \times \mathcal{L}_{\neq\bot} \rightarrow \mathcal{E}$ be a functor into a cocomplete category $\mathcal{E}$ such that
  \begin{enumerate}
  \item $F([n],* : a \le b)$ is an epimorphism for all $\bot \neq a \le b$,
  \item \label{item:uniqueMorph} for every cover $(a_i \in \mathcal{L}_{\neq\bot})_{i \in I}$ of $a \in \mathcal{L}_{\neq\bot}$ and every family of morphisms $(f_i : F([n],a_i) \rightarrow E)$ with
    $$f_i \circ F([n],* : a_i \land a_j \le a_i) = f_j \circ F([n],* : a_i \land a_j \le a_j),$$
    there is a unique $f : F([n], a) \rightarrow E$ with $f_i = f \circ F([n],* : a_i \le a)$.
  \end{enumerate}
  Then there is a left Kan extension along the Yoneda embedding
  \begin{center}
    \begin{tikzcd}
      \Delta \times \mathcal{L}_{\neq\bot} \arrow[rr, "F"] \arrow[rd, "y"'] & {} \arrow[d, "\eta", Rightarrow, shift right] & \mathcal{E} \\
      & \SVal{\mathcal{L}} \arrow[ru, "\Lan{y}{F}"', dashed]  &
    \end{tikzcd}
  \end{center}
  that is left adjoint to $\Hom{\mathcal{E}}{F(-,-)}{-} : \mathcal{E} \rightarrow \SVal{\mathcal{L}}$.
  \begin{Proof}
    The existence of this Kan extension is guaranteed by \cref{Cor:ColimitFormula}. For the adjoint, we use \cref{Prop:KanExtYonedaSubcat}. For this, we must verify the required conditions on $\SVal{\mathcal{L}}$ listed in \cref{Prop:KanExtYonedaSubcat}.
    \begin{enumerate}
    \item Has been verified in \cref{Lem:ValYonedaEmb}.
    \item To see that $\Hom{\mathcal{E}}{F(-,-)}{E}$ is an object in $\SVal{\mathcal{L}}$, we must verify that $\Hom{\mathcal{E}}{F([n],-)}{E} : \mathcal{L}^\op \rightarrow \SET$ is an $\mathcal{L}$-valued set:
      \begin{itemize}
      \item $\Hom{\mathcal{E}}{F([n],* : a \le b)}{E} : \Hom{\mathcal{E}}{F([n],b)}{E} \rightarrow \Hom{\mathcal{E}}{F([n], a)}{E}$ is injective for any $\bot \neq a \le b$:
        Let $f,g \in \Hom{\mathcal{E}}{F([n],b)}{E}$. We have:
        \begin{align*}
          \Hom{\mathcal{E}}{F([n],* : a \le b)}{E}(f) &= \Hom{\mathcal{E}}{F([n],* : a \le b)}{E}(g)\\
          f \circ F([n],* : a \le b) &= g \circ F([n],* : a \le b)
        \end{align*}
        Since we have demanded that $F([n],* : a \le b)$ are epimorphisms, we can cancel on the right and obtain $f = g$.
      \item For the sheaf condition, we can directly apply \cref{Lem:ExtendToSheaf}, because the functor $\Hom{\mathcal{E}}{F([n],-)}{E}$ satisfies the bottomless sheaf condition by \ref{item:uniqueMorph}.
      \end{itemize}
    \end{enumerate}
  \end{Proof}
\end{Lemma}

\clearpage
\section{Extended Pseudo-Metric Spaces and (Finite) Metric Realizations} \label{Sec:EPMetAndMetRe}
It is well known that simplicial sets can be used to describe many topological spaces via the geometric realization $\TopRe : \SET^{\Delta^\op} \rightarrow \Top$. To construct the geometric realization, one defines a functor $T : \Delta \rightarrow \Top$ that interprets objects $[0],[1],[2],[3],\ldots$ as points, line segments, triangles, tetrahedrons, \ldots, known as the topological $n$-simplices. Then we form a left Kan extension along the Yoneda embedding $y$.
\begin{center}
  \begin{tikzcd}
    \Delta \arrow[rr, "T"] \arrow[rd, "y"'] & {} \arrow[d, "\eta", Rightarrow, shift right] & \Top \\
    & \SET^{\Delta^\op} \arrow[ru, "\TopRe = \Lan{y}{T}"', dashed]  &
  \end{tikzcd}
\end{center}
The existence of this Kan extension is guaranteed because $\Top$ is cocomplete, which allows us to use the colimit formula for Kan extensions. We can think of this as defining what $\TopRe$ should do on basic building blocks for topological spaces, then extending to combinations of those building blocks. For a more explicit derivation, see \cref{Subsection:GeometricRealization}.

We want to define a similar concept for simplicial normed sets. We can use the same basic building blocks, but now we must include the notion of membership strength, which we could interpret as some kind of distance, surface area, or volume. However, topological spaces only have an abstract notion of neighborhood, but there is no notion of distance. Intuitively, we could replace topological spaces with metric spaces and then use points, line segments, triangles, tetrahedrons, \ldots of different scales as basic building blocks.

However, the category of metric spaces is not cocomplete, so the existence of the relevant Kan extension is not guaranteed. This motivates the category of extended pseudo-metric spaces, which includes all ordinary metric spaces but is cocomplete. The associated metric realization was first defined by Spivak \cite[Section~3]{Spivak2009METRICRO}.
\bigbreak
We will begin with the definition of the category of extended pseudo-metric spaces in \cref{Subsection:EPMetSpaces}. Spivak proved that this category is cocomplete \cite[Section~2]{Spivak2009METRICRO}. We provide a more explicit proof of this result.

We then construct the metric realization and the respective adjoint in \cref{Subsection:MetricRealization}. Our construction is similar to \cite[Section~3]{Spivak2009METRICRO}, although we use normed sets instead of fuzzy sets to simplify computations in \cref{Subsection:ClassicalMetricRealization}. We discuss how Spivak's construction relates to ours in \cref{Sec:SpivakMetRe} using the results from \cref{Subsection:ClassicalMetricRealization}.

Both Spivak's metric realization \cite[Section~3]{Spivak2009METRICRO} and ours from \cref{Subsection:MetricRealization} are constructed in terms of the sheaf-theoretic variant of fuzzy sets and normed sets. The sheaf-theoretic variants make the construction easier, but have the drawback that the constructed functors are harder to understand. To understand them better, we compute their actions in terms of the equivalent category of classical normed sets $\CNSet^{\Delta^\op}$ in \cref{Subsection:ClassicalMetricRealization}. This is our main contribution.

Finally, we verify the construction of McInnes et al.'s finite variant of Spivak's metric realization in \cref{Sec:FinMetRe}.

\subsection{Extended Pseudo-Metric Spaces} \label{Subsection:EPMetSpaces}
We begin with the definition of extended pseudo-metric spaces, which is almost identical to the definition of metric spaces. {\it Extended} refers to the fact that the metric may take values in the extended positive real number line $[0,\infty]$. {\it Pseudo} refers to the fact that the metric may assign a distance of 0 to different points.
\begin{Definition} \cite[Definition~2.1]{Spivak2009METRICRO} An \defind{extended pseudo-metric space} $(M,d)$ consists of a set $M$ and a map $d : M \times M \rightarrow [0,\infty]$ such that for all $x,y,z \in M$:
  \begin{enumerate}
  \item[(M1)] $d (x,x) = 0$
  \item[(M2)] $d (x,y) = d (y,x)$
  \item[(M3)] $d (x,z) \leq d (x,y) + d (y,z)$,
  \end{enumerate}
  where $r < \infty$ for all $r \in [0,\infty)$, $\infty \le \infty$ and $\infty + \infty = \infty$.
  
  A map $f : X \rightarrow Y$ between extended pseudo-metric spaces $(X,d_X)$ and $(Y,d_Y)$ is called \defind{non-expansive} if
  \[ d_Y(f(x),f(y)) \leq d_X(x,y) \qquad \forall x,y \in X. \]
  Extended pseudo-metric spaces and non-expansive maps form the category \defind{$\EPMet$}.
\end{Definition}
\begin{Remark} The category of metric spaces \defind{$\Met$} is a full subcategory of $\EPMet$.\\
  Ordinary metric spaces require the stricter condition that $d (x,y) = 0$ if and only if $x = y$. Additionally, the distance between two points is always finite.
\end{Remark}

Spivak refers to these spaces as uber-metric spaces in \cite[Definition~2.1]{Spivak2009METRICRO}. We use the same name as McInnes et al. \cite[Definition~6]{UMAP}.

As mentioned at the beginning of this section, we are interested in $\EPMet$ because it is cocomplete \cite[Lemma~2.2]{Spivak2009METRICRO}. We can show this by proving that arbitrary coproducts and coequalizers exist.
Coproducts in the category $\EPMet$ are relatively easy to define; the key ingredient here is to set the distance between different components of a coproduct to $\infty$. This is only possible in $\EPMet$ and is the reason why $\Met$ is not cocomplete.
\begin{Example} The category $\Met$ has almost no coproducts: \\
  Let $(M_1,d_{M_1})$ and $(M_2,d_{M_2})$ be nonempty metric spaces. Their binary coproduct cannot exist.\\
  \begin{Proof}
    Let $(P,d_P)$ be the coproduct of $(M_1,d_{M_1})$ and $(M_2,d_{M_2})$ with the inclusions $\iota_1,\iota_2$. For any $c > 0$, consider the two-point metric space $(\{p_1,p_2\},d_c)$ where the distance between $p_1$ and $p_2$ is $c$:
    \[ d_c(p_i,p_j) =
      \begin{cases}
        c & \text{for } i \neq j\\
        0 & \text{for } i = j
      \end{cases} \]
    We can use the universal property of the coproduct to construct a morphism that separates the components of $P$. Consider the diagram with $f_1(x_1) = p_1$ and $f_2(x_2) = p_2$ for all $x_1 \in X_1$ and $x_2 \in X_2$:
    \begin{center}
      \begin{tikzcd}
        (M_1,d_{M_1}) \arrow[r, "\iota_1"] \arrow[rd, "f_1"'] & (P,d_P) \arrow[d, "h", dashed] & \arrow[l, swap, "\iota_2"] (M_2,d_{M_2}) \arrow[ld, "f_2"] \\
        & (\{p_1,p_2\},d_c)  &
      \end{tikzcd}
    \end{center}
    By the universal property of the coproduct, there would have to be a non-expansive map $h : (P,d_P) \rightarrow (\{ p_1,p_2 \},d_c)$ with $h \circ \iota_1 = f_1$ and $h \circ \iota_2 = f_2$.
    Then the non-expansiveness of $h$ would imply that the distance between elements of different components in $P$ must be at least $c$:  \[ d_c(h \circ \iota_1(x_1),h \circ \iota_2(x_2)) = c \leq d_P(\iota_1(x_1), \iota_2(x_2)) \qquad \text{for any } x_1 \in M_1,x_2 \in M_2  \]
    The definition of metric spaces forces us to set the distance between elements of different components to some concrete number $d_P(\iota_1(x_1),\iota_2(x_2)) \neq \infty$. But we can always construct $h$ again for any two-point space with an even larger distance $c \geq d_P(\iota_1(x_1),\iota_2(x_2))$, so this binary coproduct of nonempty components cannot exist. In fact, any coproduct where more than one component is nonempty cannot exist in the category $\Met$.
  \end{Proof}
\end{Example}

\begin{Lemma} \cite[Lemma~2.2]{Spivak2009METRICRO} The category $\EPMet$ has arbitrary coproducts.\\
  Let $(M_i,d_i)_{i\in I}$ be a family of objects in $\EPMet$.\\
  Their coproduct is given by $\left(\amalg_{i\in I}M_i, d\right)$, where $\amalg_{i\in I}M_i$ is the coproduct in $\SET$ and $d$ is the unique map induced by the universal property of the coproduct in set with:
  \begin{align*}
    d& : \amalg_{i\in I}M_i \times \amalg_{i\in I}M_i \rightarrow [0,\infty]\\
    d& \circ (\iota_i \times \iota_j)(x,y) =
       \begin{cases}
         d_i(x,y) &\text{if } i = j\\
         \infty        &\text{if } i \neq j\\
       \end{cases}\qquad \forall i,j \in I
  \end{align*}
  \begin{Proof}
    
    It is easy to verify by case distinction that (M1), (M2), and (M3) are satisfied for $d$, thus $\left(\amalg_{i\in I}M_i, d\right)$ is an extended pseudo-metric space. We then verify that all morphisms in the underlying coproduct define morphisms in $\EPMet$:
    \begin{itemize}
    \item The canonical injections $\iota_i : M_i \rightarrow \amalg_{i\in I}M_i$ obtained from the coproduct in $\SET$ are non-expansive:
      \smallbreak
      For $x,y \in M_i$, we have $d(\iota_i(x),\iota_i(y)) = d_i(x,y)$
    \item Let $(N,d_N)$ be another extended pseudo-metric space. Given a family $(f_i : M_i \rightarrow N)_{i \in I}$ of non-expansive maps, let $f : \amalg_{i\in I}M_i \rightarrow N$ be the unique map with $f \circ \iota_i = f_i$. Then $f$ is non-expansive:
      \smallbreak
      For $\iota_i(x),\iota_i(y) \in \amalg_{i\in I}M_i$, we have:
      \[ d_N(f(\iota_i(x)),f(\iota_i(y))) = d_N(f_i(x),f_i(y)) \leq d_i(x,y) = d(\iota_i(x),\iota_i(y)) \]
      For $\iota_i(x),\iota_j(y) \in \amalg_{i\in I}M_i$ with $i \neq j$, the right-hand side reduces to $\infty$, so the inequality is always satisfied.
    \end{itemize}
  \end{Proof}
\end{Lemma}

Coequalizers in $\EPMet$ are more difficult to define. We can again draw inspiration from $\SET$, where a coequalizer object is a quotient set $M/\mathord{\sim}$. But we cannot simply define a metric on $M/\mathord{\sim}$ by setting $d_{M/\mathord{\sim}}([x],[y]) = d_M(x,y)$, since this is not guaranteed to be well-defined on the equivalence classes. If we naively define $$d_{M/\mathord{\sim}}([x],[y]) = \inf_{\substack{p \sim x\\ q \sim y}} d_M(p,q),$$ we have a well-defined map that satisfies (M1) and (M2) but the triangle inequality (M3) could still be violated if there are ``shortcut equivalence classes'': For example, consider $M \subseteq \mathbb{R}^2$ with the Euclidean metric $d_M$ and the equivalence classes $[x],[y]$ and $[z]$ as depicted.
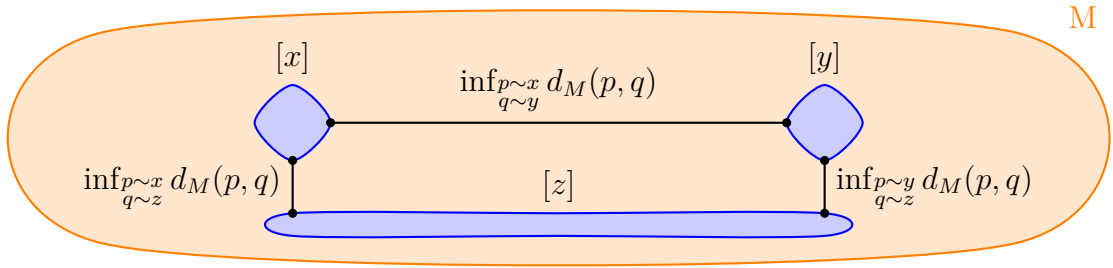
\begin{center}
  \begin{tikzpicture}
    \newcommand{\setPoint}[3]{
      \begin{scope}
        \filldraw[black] (#1,#2) circle (1.5pt); 
        \coordinate (#3) at (#1,#2);
      \end{scope}
    }
    
    \filldraw[fill=orange!20, draw=orange, thick]
    plot[smooth cycle, tension=0.5] coordinates {(-2,-1.6) (10,-1.6) (10,1.2) (-2,1.2)};
    \filldraw[fill=blue!20, draw=blue, thick]
    plot[smooth cycle, tension=0.5] coordinates {(4,-1.2) (0.5,-1.2) (0.5,-1.5) (4,-1.5) (7.5,-1.5) (7.5,-1.2)};
    \node[anchor=south] at (4, -1.2) {$[z]$};
    \setPoint{0.5}{-1.2}{b}
    \setPoint{7.5}{-1.2}{c}
    
    \filldraw[fill=blue!20, draw=blue, thick]
    plot[smooth cycle, tension=0.5] coordinates {(0.5,0.5) (1,0) (0.5,-0.5) (0,0)};
    \node[anchor=south] at (0.5, 0.5) {$[x]$};
    \setPoint{0.5}{-0.5}{a}
    \setPoint{1}{0}{e}
    
    \filldraw[fill=blue!20, draw=blue, thick]
    plot[smooth cycle, tension=0.5] coordinates {(7.5,0.5) (8,0) (7.5,-0.5) (7,0)};
    \node[anchor=south] at (7.5, 0.5) {$[y]$};
    \setPoint{7.5}{-0.5}{d}
    \setPoint{7}{0}{f}
    
    \draw[thick] (a) -- (b) node[midway, left] {$\inf_{\substack{p \sim x\\ q \sim z}} d_M(p,q)$};
    \draw[thick] (c) -- (d) node[midway, right] {$\inf_{\substack{p \sim y\\ q \sim z}} d_M(p,q)$};
    
    \draw[thick] (e) -- (f) node[midway, above] {$\inf_{\substack{p \sim x\\ q \sim y}} d_M(p,q)$};

    \node[color=orange] at (10.9,1.4) {M};
  \end{tikzpicture}
\end{center}
Here, $[y]$ acts as the shortcut. If we move from $[x]$ to $[z]$ and then to $[y]$ the combined distance can be shorter than the direct distance from $[x]$ to $[y]$.\\
To satisfy the triangle inequality, we must define $d_{M/\mathord{\sim}}([x],[y])$ to be the infimum of ``direct and indirect paths'' from $x$ to $y$ where we may switch to a different element of the equivalence class at each intermediate step:

\begin{Definition} \cite[Lemma~2.2]{Spivak2009METRICRO} Let $M$ be an extended pseudo-metric space, let $\sim$ be an equivalence relation on $M$, and let $x,y \in M$. A sequences of pairs $(p_i \in M,q_i \in M)_{1 \leq i \leq n}$ is called an $\sim$-path from $x \in M$ to $y \in M$, if
  \begin{enumerate}
  \item $q_i \sim p_{i+1}$ for all $1 \leq i < n$,
  \item $x \sim p_1$ and $q_n \sim y$.
  \end{enumerate}
  The length $\pathlen{(p_i,q_i)_{1 \leq i \leq n}}$ of a $\sim$-path $(p_i,q_i)_{1 \leq i \leq n}$ is defined as the sum of all distances along the path:
  \[ \pathlen{(p_i,q_i)_{1 \leq i \leq n}} = d_M(p_1,q_1)+\cdots+d_M(p_n,q_n) \]
    \begin{center}
    \begin{tikzpicture}
      \filldraw[fill=orange!20, draw=orange, thick]
      plot[smooth cycle, tension=0.5] coordinates {(-1,0.5) (3,3.5) (7, 3) (10, 4) (12, 4) (13.5,0) (11,-2) (7,-3) (2, -4)};
      \newcommand{\equivalenceClass}[7]{
        \begin{scope}[shift={(#1,#2)}]
          \filldraw[fill=blue!20, draw=blue, thick]
          plot[smooth cycle, tension=0.5] coordinates #7;
          \node[anchor=south] at (0, 0.8) {$[#3]$};
          \node at (0, 0) {$\sim$};
          \filldraw[black] (-0.35, 0) circle (1.5pt) node[anchor=#5] {$#3$}; 
          \filldraw[black] (0.35, 0) circle (1.5pt) node[anchor=#6] {$#4$};  
          \coordinate (#3-point) at (-0.35, 0);
          \coordinate (#4-point) at (0.35, 0);
        \end{scope}
      }
      \equivalenceClass{0.5}{-0}{x}{p_1}{south}{south}{{(-0.8,0.3)(0,0.7)(0.7,0.7) (1,0) (0.2,-0.7) (-0.4,-0.6)}}
      \equivalenceClass{2.5}{-3}{q_1}{p_2}{north}{north}{{(-0.8,0.7)(0,0.7)(0.7,0.8) (1,-0.1) (0.2,-0.7) (-0.6,-0.6)}}
      \equivalenceClass{5}{-2}{q_2}{p_3}{south}{north}{{(-0.9,0.7)(0,0.7)(0.7,0.6) (1,0.1) (0.3,-0.7) (-0.5,-0.6)}}
      \equivalenceClass{10}{2}{q_{n-1}}{p_n}{south}{south}{{(-1,0.6)(0,0.6)(0.7,0.7) (1,0) (0.5,-0.7) (0,-0.5) (-0.6,-0.5)}}
      \equivalenceClass{12}{0}{q_n}{y}{north}{south}{{(-0.7,0.4)(0,0.7)(0.6,0.7) (0.8,0) (0.2,-0.6) (-0.5,-0.6)}}
      \filldraw[black] (6.5, 0) circle (1.5pt) node[anchor=south] {$q_4$};  
      \coordinate (q_4-point) at (6.5, 0);
      \filldraw[black] (6.75, 0) circle (0.5pt);
      \filldraw[black] (7, 0) circle (0.5pt);
      \filldraw[black] (7.25, 0) circle (0.5pt);
      \filldraw[black] (7.5, 0) circle (0.5pt);
      \filldraw[black] (7.75, 0) circle (0.5pt);
      
      \filldraw[black] (8, 0) circle (1.5pt) node[anchor=north] {$p_{n-1}$};  
      \coordinate (p_{n-1}-point) at (8, 0);
      
      \draw[thick] (p_1-point) -- (q_1-point);
      \draw[thick] (p_2-point) -- (q_2-point);
      \draw[thick] (p_3-point) -- (q_4-point);
      \draw[thick] (p_{n-1}-point) -- (q_{n-1}-point);
      \draw[thick] (p_n-point) -- (q_n-point);
    \end{tikzpicture}
  \end{center}
\end{Definition}

\begin{Lemma} \cite[Lemma~2.2]{Spivak2009METRICRO} \label{Lem:EPMetQuot} The category $\EPMet$ has arbitrary quotients.
  Let $(M,d_M)$ be an extended pseudo-metric space and $\mathord{\sim} \subseteq M \times M$ be an arbitrary equivalence relation. The quotient $M/\mathord{\sim}$ equipped with the extended pseudo-metric
  \begin{align*}
    d_{M/\mathord{\sim}}& : M/\mathord{\sim} \; \times \; M/\mathord{\sim} \; \rightarrow \; [0,\infty]\\
    d_{M/\mathord{\sim}}&([x],[y]) = \inf \{\; \pathlen{(p_i,q_i)_{1 \leq i \leq n}} \; | \; (p_i,q_i)_{1 \leq i \leq n} \text{ is a $\sim$-path from $x$ to $y$} \}
  \end{align*}
  forms an extended pseudo-metric space with the following universal property:
  \smallbreak
  The canonical surjection $\pi : M \rightarrow M/\mathord{\sim}, \, x \mapsto [x]$ is non-expansive. For every non-expansive map $ f : M \rightarrow N$ that is constant on the equivalence classes, there is a unique non-expansive $\widetilde{f} : M/\mathord{\sim} \rightarrow N$.

  \begin{center}
    \begin{tikzcd}
      N & M/\mathord{\sim} \arrow[l, "\exists! \widetilde{f}"', dashed]    \\
      & M \arrow[lu, "{f|_{[x]} \text{is constant}}", "f"'] \arrow[u, "\pi"']
    \end{tikzcd}
  \end{center}

  \begin{Proof}
    \begin{itemize}
      \item $(M/\mathord{\sim}, d_{M/\mathord{\sim}})$ is an extended pseudo-metric space: Let $x,y,z \in M$.
      \begin{enumerate}
      \item[(M1)] $d_{M/\mathord{\sim}}([x],[x]) = 0$ holds, because the infimum ranges over the direct path $(p_i,q_i)_{1 \leq i \leq 1}$ with $p_1 = q_1 = x$ that obviously has length 0.
      \item[(M2)] symmetry holds:
        \smallbreak
        Let $(p_i,q_i)_{1 \leq i \leq n}$ be a $\sim$-path that participates in $d_{M/\mathord{\sim}}([x],[y])$.
        \smallbreak
        Then the reversed $\sim$-path $(q_{n+1-i},p_{n+1-i})_{1 \leq i \leq n}$ participates in $d_{M/\mathord{\sim}}([y],[x])$.\\
        As these $\sim$-paths obviously have the same length, the infima in $d_{M/\mathord{\sim}}([x],[y])$ and $d_{M/\mathord{\sim}}([y],[x])$ range over the same values and are therefore equal.
      \item[(M3)] the triangle inequality holds:
        \smallbreak
        Let $(p_i,q_i)_{1 \leq i \leq n}$ and $(p'_i,q'_i)_{1 \leq i \leq n'}$ be $\sim$-paths that participate in\\$d_{M/\mathord{\sim}}([x],[y])$ and $d_{M/\mathord{\sim}}([y],[z])$, respectively.
        
        Then we can form a new $\sim$-path $(p''_i,q''_i)_{1 \leq i \leq n+n'}$ by concatenation
        \[ p''_i :=
          \begin{cases}
            p_i      & i \leq n\\
            p'_{i-n} & i > n
          \end{cases}
          \qquad q''_i :=
          \begin{cases}
            q_i      & i \leq n\\
            q'_{i-n} & i > n
          \end{cases}, \]
        that participates in $d_{M/\mathord{\sim}}([x],[z])$. It is again obvious that
        \[ \pathlen{(p''_i,q''_i)_{1 \leq i \leq n+n'}} = \pathlen{(p_i,q_i)_{1 \leq i \leq n}} + \pathlen{(p'_i,q'_i)_{1 \leq i \leq n'}} \]
        so we have
        $d_{M/\mathord{\sim}}([x],[z]) \leq d_{M/\mathord{\sim}}([x],[y])$ + $d_{M/\mathord{\sim}}([y],[z])$.
      \end{enumerate}

    \item $\pi$ is non-expansive:
      \[ d_{M/\mathord{\sim}}([x],[y]) \leq d_M(p_1, q_1) = d_M(x,y) \qquad \text{with } n=1 \text{ and } p_1 = x, q_1 = y \]
    \item Given another extended pseudo-metric space $(N,d_N)$ and a non-expansive map $f : M \rightarrow N$ such that $f_{[x]}$ is constant for all $x \in M$, we obtain a unique map $\widetilde{f} : M/\mathord{\sim} \rightarrow N$ with $\widetilde{f} \circ \pi = f$ from the universal property of the quotient in Set. We verify that $\widetilde{f}$ is non-expansive:
      \smallbreak
      Any $\sim$-path that participates in the infimum in $d_{M/\mathord{\sim}}([x],[y])$ can be transported through $f$ to an $=$-path in $N$, where we view $=$ as a relation on $N$:
      \smallbreak
      Let $(p_i,q_i)_{1 \leq i \leq n}$ be a $\sim$-path from $x$ to $y$. As $f$ is constant on $\sim$-equivalence classes, we have:
      \[ f(x) = f(p_1), \quad f(q_n) = f(y), \quad f(p_{i+1}) = f(q_i) \; \; \forall 1 \leq i < n \]
      Thus $(f(p_i),f(q_i))_{1 \leq i \leq n}$ is an $=$-path between $f(x)$ and $f(y)$ in $N$.
      Due to the triangle inequality in $N$, the direct distance between $f(x)$ and $f(y)$ cannot be larger than the length of any $=$-path between these points. Additionally, as $f$ is non-expansive, the length of the path cannot increase when transported through $f$, so for any $\sim$-path from $x$ to $y$, we have:
      \begin{align*}
        d_N(f(x),f(y)) \leq& \; \pathlen{(f(p_i),f(q_i))_{1 \leq i \leq n}} & \text{(M3)} \\
        \leq& \; \pathlen{(p_i,q_i)_{1 \leq i \leq n}}     & (\text{non-expansiveness of } f)
      \end{align*}
      With this upper bound for all participating paths, we conclude
      \[ d_N(\widetilde{f}([x]), \widetilde{f}([y])) = d_N(f(x),f(y)) \leq d_{M/\mathord{\sim}}([x],[y]). \]
    \end{itemize}
  \end{Proof}
\end{Lemma}

Intuitively, the quotient metric can only decrease the distance between points by introducing shortcuts, but can never increase distances. This also implies that open balls around points may become larger when forming the quotient. We can formalize this intuition as follows.
\begin{Remark} \label{Rem:QuotientMetric} Let $M$ be an extended pseudo-metric space and let $\sim$ be an equivalence relation on $M$. Then we have:
  \begin{enumerate}
  \item $d_{M/\mathord{\sim}}([x],[y]) \leq d_M(x,y)$
  \item $\pi(B^{M}_{<\epsilon}(x)) \subseteq B^{M/\mathord{\sim}}_{<\epsilon}([x])$ 
  \end{enumerate}
  where $B_{<\epsilon}^N(x) = \{ y \in N \, | \, d_N(x,y) < \epsilon \}$ denotes the open ball of radius $\epsilon$ around $x$ with the metric in $N$.
  \begin{Proof}
    The first claim follows because the direct $\sim$-path from $x$ to $y$ participates in the infimum in the definition of $d_{M/\mathord{\sim}}$. The second claim follows from the first.
  \end{Proof}
\end{Remark}

The first claim from \cref{Rem:QuotientMetric} allows us to directly construct non-expansive maps into quotient spaces.
\begin{Corollary} \label{Cor:QuotientInducedMap} Let $f : M \rightarrow N$ be non-expansive and let $\sim$ be an equivalence relation on $N$. Then the map $g : M \rightarrow N/\mathord{\sim}, \; x \mapsto [f(x)]$ is also non-expansive.
  \begin{Proof}
    For any $x, y \in M$, we have:
    \[ d_{N/\mathord{\sim}}(g(x), g(y)) = d_{N/\mathord{\sim}}([f(x)], [f(y)]) \leq d_N(f(x), f(y)) \leq d_M(x, y) \]
    where the first inequality follows from \cref{Rem:QuotientMetric} and the second from the non-expansiveness of $f$.
  \end{Proof}
\end{Corollary}

Similar to the category $\SET$ of Sets or the category $\Top$ of topological spaces, coequalizers can be defined as quotients in $\EPMet$.

\begin{Lemma} \cite[Lemma~2.2]{Spivak2009METRICRO} \label{Lem:EPMetCoeq} The category $\EPMet$ has coequalizers.\\
  The coequalizer of two parallel morphisms $A \overset{f}{\underset{g}{\rightrightarrows}} B$ in $\EPMet$ is given by $\coeq{f,g}=B/\mathord{\sim}$, where $\sim$ is the equivalence relation generated by $f(a) \sim g(a) \quad \forall a \in A$ and the canonical surjection $\pi : B \rightarrow B/\mathord{\sim}$.
  \begin{Proof}
    Let $(C,d_C)$ be an extended pseudo-metric space and $\pi' : B \rightarrow C$ be another non-expansive map with $\pi' \circ f = \pi' \circ g$. Then $\pi'$ is constant on the equivalence generated by $f(a) \sim g(a) \quad \forall a \in A$. The universal property of the coequalizer then follows from the universal property of the quotient $B/\mathord{\sim}$.
  \end{Proof}
\end{Lemma}

\begin{Corollary} \cite[Lemma~2.2]{Spivak2009METRICRO} The category $\EPMet$ is cocomplete.\\
  This follows from the well known fact that arbitrary colimits may be constructed from arbitrary coproducts and coequalizers. The construction can be found in \cite[Theorem~3.4.12]{riehl2017category}.
\end{Corollary}

\subsection{The Metric Realization} \label{Subsection:MetricRealization}
We can now define the metric realization similar to the geometric realization from \cref{Subsection:GeometricRealization} as a left Kan extension. For this, we define a functor analogous to the functor $G : \Delta \rightarrow \Top$ from \cref{Def:G}. This functor is defined using the metric $n$-simplices.
\begin{Definition} \label{Def:PseudoMetricSimplicies} \;
  \begin{itemize}
  \item For $n \in \mathbb{N}_0$ and $a \in [0,\infty)$, the \defind{metric $n$-simplex of size $a$} is the extended pseudo-metric space
    \[ \Delta^{n,a} = (\Delta^{n},d^{n,a}_1) \]
    where $\Delta^{n}$ is the same underlying set as for the topological $n$-simplex from \cref{Def:TopSimplex} and $d_1^{n,a}$ is the \defind{$\ell_1$ metric}
    \begin{align*}
      d_1^{n,a}& : \Delta^{n,a} \times \Delta^{n,a} \rightarrow [0,\infty)\\
      d_1^{n,a}&(x,y) =  a \cdot\norm{x - y}_1^n \qquad \text{where } \norm{ v }_1^n = \sum_{0 \le i \le n} | v_i |,
    \end{align*}
    also known as \defind{Manhattan metric}, scaled by $a$.
  \end{itemize}
\end{Definition}

This definition is inspired by Spivak \cite[Section~3]{Spivak2009METRICRO}. However, Spivak defines the underlying sets of the metric $n$-simplices as
$$\{ (x_0, \ldots, x_n) \subseteq \mathbb{R}_{\ge 0}^{n+1} \, | \, x_0 + \cdots + x_n = -\log(a) \}$$
While we define them as
$$\Delta^{n} = \{ (x_0, \ldots, x_n) \subseteq \mathbb{R}_{\ge 0}^{n+1} \, | \, x_0 + \cdots + x_n = 1 \}.$$
Essentially, the underlying sets of Spivak's metric simplices scale with the parameter $a$, while we use the same underlying sets and instead scale the $\ell_1$-metric.

This implies that our metric $n$-simplex of size 0 contains infinitely many points, all with distance 0 from each other, while Spivak's metric $n$-simplex of size 0 only contains one point. Our definition avoids the need for rescaling the points when we later define the functor in \cref{Def:T}. This will make computations easier in \cref{Subsection:ClassicalMetricRealization} and resolves the division by 0 error that results from Spivak's scaling factor $\frac{\log(b)}{\log(a)}$, where $a$ is allowed to be 1 \cite[Section~3]{Spivak2009METRICRO}.

We will discuss Spivak's use of $-\log(a)$ later after \cref{Def:MetReSpivak}.

Spivak also incorrectly uses the $\ell_2$, or Euclidean metric. However, we are forced to choose the $\ell_1$ metric to make the degeneracy maps non-expansive. The following counterexample shows that this property breaks down for all other $\ell_p$ metrics with $p \in (1,\infty)$.

\begin{Remark} \label{Rem:L1} For every $\ell_p$ metric 
  \begin{align*}
    d_p^n& : \Delta^{n} \times \Delta^{n} \rightarrow [0,\infty)\\
    d_p^n&(x,y) =  \norm{x - y}_p^n \qquad \text{where } \norm{ v }_p^n = ( \sum_{0 \le i \le n} ( v_i )^p )^{1/p}
  \end{align*}
  with $p \in (1,\infty)$, the degeneracy maps $s^{n}_i$ are not non-expansive for $n \ge 1$.
  \begin{Proof}
    Consider $x :=(0, \ldots, \frac{1}{2}, \frac{1}{2} , \ldots, 0) \in \Delta^{n+1}$ where $\frac{1}{2}$ is at the $i$th and $(i+1)$th entry. Then the norm of $x$ is
    
    $$\norm{x}_p^{n+1} = \left(\left(\frac{1}{2}\right)^p+\left(\frac{1}{2}\right)^p\right)^{1/p} = 2^{(1/p) - 1}.$$
    The norm of $s^{n,a}_i(x)$ is
    $$\norm{s^{n,a}_i(x)}_p^{n} = \left(\left(\frac{1}{2}+\frac{1}{2}\right)^p\right)^{1/p} = 1.$$
    We have $2^{(1/p) - 1} > 1$ for $p \in (1,\infty)$, thus $s^{n,a}_i$ is not non-expansive.
  \end{Proof}
\end{Remark}

\begin{Lemma} \label{Lem:FaceDegNonExpansive} The face $f^{n}_i$ and degeneracy maps $s^{n}_i$ from \cref{Def:TopSimplex} define non-expansive maps
    \begin{align*}
      &f^{n,a} : \Delta^{n-1,a} \rightarrow  \Delta^{n,a}, \quad x \mapsto f^n(x)\\
      &s^{n,a} : \Delta^{n+1,a} \rightarrow  \Delta^{n,a}, \quad x \mapsto s^n(x)
    \end{align*}
    for $n \in \mathbb{N}, \, a \in [0,\infty), \, i \in \{0, \ldots, n\}$.
    \begin{Proof}
      The metric $d_1^{n,a}$ from \cref{Def:PseudoMetricSimplicies} is induced by the norm $a \cdot \norm{ - }_1^n$, so it suffices to check non-expansiveness with respect to this norm. Since this norm is obtained by scaling $\norm{ - }_1^n$ with the constant factor $a$, it suffices to check non-expansiveness with respect to $\norm{ - }_1^n$.
    \begin{enumerate}
    \item Let $n \in \mathbb{N}, \, i \in \{0, \ldots, n\}$. The face maps are distance preserving:
      \begin{align*}
        \norm{ f^{n}_i(x) }_1^{n} =& \norm{ (x_0,\ldots,0,\ldots,x_{n-1}) }_1^{n}\\
        =& |x_0| + \cdots + |0| + \cdots + |x_{n-1}|\\
        =& |x_0| + \cdots + |x_{n-1}| = \norm{ x }^{n-1}_1
      \end{align*}
      \item Let $n \in \mathbb{N}, \, i \in \{0, \ldots, n\}$. The degeneracy maps are distance preserving:
        \begin{align*}
          \norm{ s^{n}_i(x) }_1^{n+1} =& \norm{ (x_0,\ldots,x_i+x_{i+1},\ldots,x_{n}) }_1^{n}\\
          =& |x_0| + \cdots + |x_i+x_{i+1}| + \cdots + |x_{n-1}|\\
          \stackrel{(*)}{=}& |x_0| + \cdots + |x_i| + |x_{i+1}| + \cdots + |x_{n}| = \norm{ x }^{n}_1
        \end{align*}
        where $(*)$ is only an equality because all coordinates are greater than or equal to zero by \cref{Def:TopSimplex}, otherwise we would only have $|x_i+x_{i+1}| \le |x_i| + |x_{i+1}|$.
    \end{enumerate}
  \end{Proof}
\end{Lemma}

With the previous lemmas, we can directly assemble the metric $n$-simplices of size $a$ into the functor $T : \Delta \times ([0,\infty), \geq) \rightarrow \EPMet$ that will be used later to construct the metric realization as a left Kan extension.
\begin{Lemma} \label{Def:T} The pseudo-metric $n$-simplices and the face and degeneracy maps define a functor $T : \Delta \times ([0,\infty), \geq) \rightarrow \EPMet$ with
  \begin{align*}
    &T([n],a) = (\Delta^{n,a},d^{n,a}_1)\\
    &T(f : [n] \rightarrow [m],* : a \geq b) = G(f),
  \end{align*}
  where the maps $G(f) : \Delta^n \rightarrow \Delta^m$ are defined in \cref{Def:G}.
  \begin{Proof}
    For $T$ to be well-defined, the maps $T(f,* : a \geq b)$ must produce non-expansive maps. This is the case because by \cref{Rem:SimpCoSimpRelations}, every morphism $f  : [n] \rightarrow [m]$ can be expressed as a composite of the face and degeneracy morphisms. These are mapped by $G : \Delta \rightarrow \Top$ to the composite map $G(f) : \Delta^n \rightarrow \Delta^m$ of the face and degeneracy maps. As shown in \cref{Lem:FaceDegNonExpansive} these maps are norm-preserving, and so is their composite. Then $G(f) : \Delta^n \rightarrow \Delta^m$ can be considered a non-expansive map in $\EPMet$.
  \end{Proof}
\end{Lemma}

We now want to use \cref{Cor:SimplicialKanExt} to derive the metric realization as a left Kan extension of $T$ along the Yoneda embedding. This requires verifying a few properties of the functor $T$ first.
\begin{Lemma} \label{Lemma:TProperties} Let $n \in \mathbb{N}$.
  \begin{enumerate}
  \item $T(\id{[n]}, * : a \geq b) : T([n],a) \rightarrow T([n],b)$ is an epimorphism for all $\infty \neq a \ge b$.
  \item For any $a$ with a cover $(a_i \in [0,\infty))_{i \in I}$ and $(f_i : T([n], a_i) \rightarrow E)_{i\in I}$ with
    \begin{align*}
      f_i \circ T([n],* : a_i \land a_j \le a_i) = f_j \circ T([n],* : a_i \land a_j \le a_j), \numberthis{eq:Compatibility}
    \end{align*}
    there is a unique $f : T([n], a) \rightarrow E$ such that $f_i = f \circ T(\id{[n]}, * : a \geq a_i)$.
  \end{enumerate}
  \begin{Proof}
      Because $T(\id{[n]}, * : a \geq b) = G(\id{[n]}) = \id{\Delta^{n}}$, we can see that $T(\id{[n]}, * : a \geq b) : \Delta^{n,a} \rightarrow \Delta^{n,b}$ is the identity of the set $\Delta^{n}$ underlying both $\Delta^{n,a}$ and $\Delta^{n,b}$.
    \begin{enumerate}
    \item This map is obviously an epimorphism.
    \item Let $(a_i \in [0,\infty))_{i \in I}$ be a cover of $a \in [0,\infty)$.\smallbreak
      In the poset $([0,\infty), \ge)$, this amounts to $a = \min_{i \in I} a_i$ where $\min_{j \in \emptyset} a_j = \infty$\\
      Then $I$ must be nonempty and there is at least one $i \in I$ with $a = a_i$.

      Now let $(f_i : T([n], a_i) \rightarrow E)_{i\in I}$ satisfy condition \eqref{eq:Compatibility}. As $T(\id{[n]}, * : a \geq b)$ is the identity, the condition \cref{eq:Compatibility} implies that the morphisms $(f_i)_{i\in I}$ all have the same underlying non-expansive map $f : \Delta^n \rightarrow E$. The non-expansive morphism $f : T([n], a) \rightarrow E$ defined by this underlying map is the desired unique element.
    \end{enumerate}
  \end{Proof}
\end{Lemma}

\begin{Definition} \label{Def:MetRe} The \defind{metric realization} $\MetRe = \Lan{y}{T} : \SNSet \rightarrow \EPMet$ is the left Kan extension of $T$ along the Yoneda embedding $y : \Delta \times ([0,\infty), \geq) \rightarrow \SNSet$.\\
  \begin{center}
    \begin{tikzcd}
      \Delta \times ([0,\infty), \geq) \arrow[rr, "T"] \arrow[rd, "y"'] & {} \arrow[d, "\eta", Rightarrow, shift right] & \EPMet \\
      & \SNSet \arrow[ru, "\Lan{y}{T} = \MetRe"', dashed]  &
    \end{tikzcd}
  \end{center}
  Its left adjoint is called the \defind{singular nerve} $$\Sing = \Hom{\EPMet}{T(-,-)}{-} : \EPMet \rightarrow \SNSet.$$
\end{Definition}
The existence of the metric realization and the adjunction from \cref{Def:MetRe} are guaranteed by \cref{Cor:SimplicialKanExt}, whose assumptions have been shown in \cref{Lemma:TProperties}.

This construction is inspired by but differs from Spivak's original definition in \cite[Section~3]{Spivak2009METRICRO}. We will explain these differences in \cref{Sec:SpivakMetRe}.

Similar to \cref{Lem:GeomCoeq} for the geometric realization, we can compute the action of this functor in terms of coequalizers.
\begin{Proposition} \label{Prop:MetRe} The metric realization $\MetRe : \SNSet \rightarrow \EPMet$ assigns
  \begin{itemize}
  \item to a simplicial normed set $S$ the extended pseudo-metric space
    \[ \MetRe(S) = \coprod_{\substack{n \in \mathbb{N}_0,\;a \in [0,\infty)\\s \in S([n],a)}}\Delta^{n,a}/\mathord{\sim} \]
    where
    \[ \iota_{m,b,S(f,* : b \geq a)(s)}(x) \sim \iota_{n,a,s}(T(f, * : b \geq a)(x))  \] \bigbreak
    for all morphisms $(f, * : b \geq a) : ([m],b) \rightarrow ([n],a)$ in $\Delta \times ([0,\infty), \geq)$, $s \in S([n],a)$, and $x \in \Delta^{m,b}$,
  \item to a morphism of simplicial normed sets $\alpha : S \Rightarrow S'$ the non-expansive map
    $$\MetRe(\alpha) : \MetRe(S) \rightarrow \MetRe(S'), \qquad [\iota_{n,a,s}(x)] \mapsto [\iota_{n,a,\alpha_{[n],a}(s)}(x)]$$
  \end{itemize}
  \begin{Proof}
    The proof is similar to \cref{Lem:GeomCoeq}.
    By \cref{Lem:CoequalizerFormula}, in this case with \cref{Rem:CoequalizerFormula}, the left Kan extension $\Lan{y}{T}(S)$ can be expressed as a coequalizer
    $$\coeq{ \coprod_{\substack{(f,* : b \geq a) : ([m],b) \rightarrow ([n],a)\\ s \in S([n],a) }} \Delta^{m,b}  \overset{\iota_{n,a,s} \circ T(f,* : b \geq a)}{\underset{\iota_{m,b,S(f,* : b \geq a)(s)}}{\rightrightarrows}} \coprod_{\substack{m \in \mathbb{N}_0, \;b \in [0,\infty)\\ s \in S([m],b)}} \Delta^{m,b} },$$
    which can be expressed as the quotient (\cref{Lem:EPMetCoeq}) in the statement above.
    
    \Cref{Lem:CoequalizerFormula} also provides the desired expression for the action on morphisms. We must only replace the coequalizer surjection $\pi$ with the equivalence class notation $[-]$.
  \end{Proof}
\end{Proposition}

\subsection{The Classical Metric Realization} \label{Subsection:ClassicalMetricRealization}

Unfortunately, the metric realization from \cref{Def:MetRe} gives little insight into the space constructed. Even when it is expressed as the coequalizer from \cref{Prop:MetRe}, simplices of every size participate in the formula. The reason for this is that it is constructed in terms of the sheaf-theoretic variant of valued sets.

We can obtain the classical perspective if we compose with the equivalence from \cref{Thm:EquivCValSVal} that translates between the classical and the sheaf-theoretic perspective:

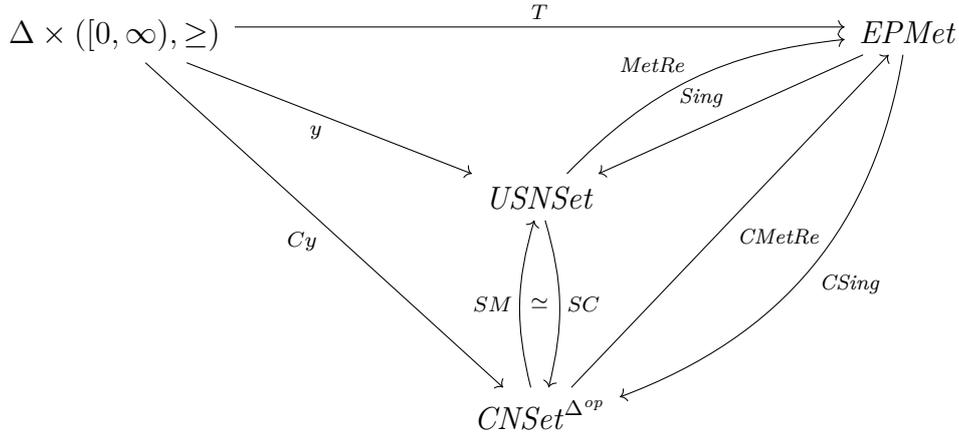
\begin{center}
  \begin{tikzcd}
    \Delta \times ([0,\infty), \ge) \arrow[rrrrrr, "T", shift left] \arrow[rrrdd, "y"'] \arrow[rrrddddd, "Cy"'] &  &  &                                                                                                        &  &  & \EPMet \arrow[llldd, "\Sing"'] \arrow[lllddddd, "\CSing", bend left] \\
    &  &  &                                                                                                        &  &  &                                                                                  \\
    &  &  & \SNSet \arrow[rrruu, "\MetRe", bend left=20] \arrow[ddd, "\simeq\,"', "SC",, bend left=15]   &  &  &                                                                                  \\
    &  &  &                                                                                                        &  &  &                                                                                  \\
    &  &  &                                                                                                        &  &  &                                                                                  \\
    &  &  & \CNSet^{\Delta^\op} \arrow[rrruuuuu, "\CMetRe"'] \arrow[uuu, "SM", bend left=15] &  &  &                                                                                 
  \end{tikzcd}.
\end{center}

Here we obtain the classical metric realization $\CMetRe$ by composing the metric realization $\MetRe$ from \cref{Def:MetRe} with the functor $\SM$ from \cref{Thm:EquivCValSVal} that translates simplicial classical valued sets into sheaf-theoretic uncurried simplicial valued sets. We also obtain the classical singular nerve $\CSing$ and the classical Yoneda embedding $\Cy$ by composing the functor $\Sing$ from \cref{Def:MetRe} and the Yoneda embedding $y$ with the functor $\SM$ from \cref{Thm:EquivCValSVal} that translates sheaf-theoretic uncurried simplicial valued sets into simplicial classical valued sets.

We can then compute the actions of the functors $\CMetRe$, $\CSing$, $\Cy$ in terms of simplicial classical valued sets. These actions are easier to understand than the actions of their sheaf-theoretic counterparts $\MetRe$, $\Sing$, $\y$.
\bigbreak
We begin by computing the actions of the classical metric realization. The key insight to understand the following proof is that the equivalence relation $\sim$ from \cref{Lem:GeomCoeq} fulfills two distinct roles:
\begin{enumerate}
\item When we allow $b \ge a$ but fix the morphism $f = \id{[m]}$, the equivalence relation $\sim$ reduces to
  $$\iota_{m,b,S(\id{[m]},*)(s)}(x) \sim \iota_{m,a,s}(T(f,* : b \ge a)(x)) = \iota_{m,a,s}(G(\id{[m]})(x)) = \iota_{m,a,s}(x).$$
  In that case, a simplex $\Delta^{m,b}$ in the component indexed by $S(\id{[m]},*)(s)$ is identified with a smaller simplex $\Delta^{m,a}$ indexed by $s$.
\item When we fix $a = b$ but allow any morphism $f : [m] \rightarrow [n]$, the equivalence relation $\sim$ reduces to
  $$\iota_{m,a,S(\id{[m]},*)(s)}(x) \sim \iota_{m,a,s}(T(f,* : b \ge a)(x)) = \iota_{m,a,s}(G(f)(x)),$$
  which resembles the equivalence relation for the geometric realization from \cref{Lem:GeomCoeq} and is responsible for gluing simplices along their faces and eliminating degeneracies.
\end{enumerate}

Whenever two simplices of different sizes are glued as described in 1., the resulting space is equivalent to a space constructed using a coproduct with the same components but missing the larger simplex. So we could remove all simplices $\Delta^{n,b}$ from the coproduct in \cref{Lem:GeomCoeq} for which there is a metric $n$-simplex $\Delta^{n,a}$ of smaller size $a \le b$ but equal dimension that is glued to it as in 1.

If the uncurried simplicial normed set $S$ in \cref{Prop:MetRe} has been constructed by the functor $\M$ from \cref{Def:M} from a simplicial classical normed set, then $S(\id{[m]},* : b \ge a)$ is the identity map. Then 1. will always identify a component indexed by $s$ and $b$ with a component that is also indexed by $s$ and a smaller size $a$.

Only the smallest size $a$ available for a given element $s$ will remain. This smallest size is exactly the norm of the element $s$ in the original classical normed set, because by \cref{Def:M}, $s$ will only be a member of all sets $S([m],b)$ where $b$ is greater than or equal to the norm of $s$ in the original classical normed set.

We formally capture this intuition in the following proposition. In the proof, we will construct an isomorphism $\phi$ from the quotient in \cref{Prop:MetRe} to a simpler quotient that only contains the smallest available simplices as components. The map $\phi$ sends every representative of the same point from simplices of different sizes to the smallest simplex available. The inverse map $\psi$ injects the remaining smallest representative back into the formula from \cref{Prop:MetRe}.

To reduce notational clutter in the following proof, we introduce notation to eliminate redundant information.
\begin{Notation} \label{Not:Simplex} Let $S$ be a simplicial classical normed set.
  \begin{itemize}
  \item For an element $s \in S_n$, we denote an $n$-simplex with size $\norm{s}_{S_n}$ as $$\Delta^s := \Delta^{n,\norm{s}_{S_n}}.$$ Recall that $\norm{s}_{S_n}$ is the value map, in this case called a norm, of a classical normed set.
  \item We denote the coproduct ranging over all elements in $S$ as
    $$\coprod_{s \in S}\Delta^s := \coprod_{\substack{n \in \mathbb{N}_0\\s \in S_n}}\Delta^{n,\norm{s}_{S_n}}$$
  \item and we denote the corresponding canonical injections as
    $$\iota_s := \iota_{n,s} : \Delta^s \rightarrow \coprod_{s \in S}\Delta^s.$$
  \end{itemize}
\end{Notation}

\begin{Proposition} \label{Prop:CMetRe} The classical metric realization $$\CMetRe := \MetRe \circ \SM : \CNSet^{\Delta^\op} \rightarrow \EPMet$$ assigns
  \begin{itemize}
  \item to a simplicial classical normed set $S : \Delta^\op \rightarrow \CNSet$ the extended pseudo-metric space
    \[ \CMetRe(S) = \coprod_{s \in S}\Delta^s/\mathord{\sim_C} \]
    with the equivalence relation $\sim_C$ generated by
    \[ \iota_{S(f)(s)}(x) \sim_C \iota_{s}(G(f)(x))  \] \bigbreak
    for all $f : [m] \rightarrow [n], \; s \in S_n, \; x \in \Delta^s$,

  \item to a simplicial map $\alpha : S \Rightarrow S'$ the non-expansive map
    \begin{align*}
      \CMetRe(\alpha) : \CMetRe(S) \qquad &\longrightarrow \qquad \CMetRe(S')\\
                  [\iota_{s}(x)] \qquad &\longmapsto \qquad [\iota_{n,\alpha_n(s)}(x)].
    \end{align*}
  \end{itemize}
  \begin{Proof}
    We first apply \cref{Prop:MetRe} and then simplify as much as possible, including the generator of the equivalence relation:
    \begin{align*}
       &\MetRe \circ \SM(S)\\
      =& (\coprod_{\substack{n \in \mathbb{N}_0,\;a \in [0,\infty)\\s \in \M(S_n)^{\leq a}}} \Delta^{n,a})/\mathord{\sim} &\text{ where }& \iota_{m,b,\SM(S)(f,*)(s)}(x) \sim \iota_{n,a,s}(T(f, *)(x))\\
      =& (\coprod_{\substack{n \in \mathbb{N}_0,\;a \in [0,\infty)\\s \in \M(S_n)^{\leq a}}} \Delta^{n,a})/\mathord{\sim} &\text{ where }& \iota_{m,b,S(f)(s)}(x) \sim \iota_{n,a,s}(G(f)(x))\\
    \end{align*}
    where the simplification of the generator of the equivalence relation $\sim$ is due to
    $$\SM(S)(f,*)(s) = \M(S(f))_{b} \circ \M(S_m)_{a \leq b}(s) = \M(S(f))_{b} (s) = S(f)(s).$$
    Now we construct bijections
    \begin{center}
      \begin{tikzcd}
        \coprod_{\substack{n \in \mathbb{N}_0,\;a \in [0,\infty)\\s \in \M(S_n)^{\leq a}}} \Delta^{n,a}/\mathord{\sim}  \arrow[rrr, "\widetilde{\phi}", shift left] &  & & \coprod_{\substack{n \in \mathbb{N}_0\\s \in S_n}}\Delta^{n,\norm{s}_{S_n}}/\mathord{\sim_C} \arrow[lll, "\widetilde{\psi}", shift left]
      \end{tikzcd}
    \end{center}
    using the universal property of the quotient space from \cref{Lem:EPMetQuot}. For this, consider the non-expansive maps
    \begin{align*}
      \phi : \coprod_{\substack{n \in \mathbb{N}_0,\;a \in [0,\infty)\\s \in \M(S_n)^{\leq a}}} \Delta^{n,a} \qquad &\longrightarrow \qquad \coprod_{s \in S}\Delta^s/\mathord{\sim_C}\\
      \iota_{n,a,s}(x) \qquad  &\longmapsto \qquad  [\iota_{s}(x)]\\
      \psi : \coprod_{s \in S}\Delta^s \qquad &\longrightarrow \qquad \coprod_{\substack{n \in \mathbb{N}_0,\;a \in [0,\infty)\\s \in \M(S_n)^{\leq a}}} \Delta^{n,a} /\mathord{\sim}\\
      \iota_{s}(x) \qquad &\longmapsto \qquad [\iota_{n,\norm{s},s}(x)].
    \end{align*}
    Note that here the mapping $\iota_{n,a,s}(x) \longmapsto \iota_{s}(x)$ is non-expansive because $$s \in \M(S_n)^{\leq a} = \{ s \in S_n | \norm{s} \leq a \}.$$
    By \cref{Cor:QuotientInducedMap}, $\phi$ and $\psi$ are non-expansive.
    These maps are constant on the equivalence classes.
    For $\psi$, we check that the images of pairs of the generators of $\sim_C$ under $\psi$ are identified by $\sim$:
    \begin{align*}
      \psi(\iota_{S(f)(s)}(x)) &= [\iota_{m,\norm{S(f)(s)},S(f)(s)}(x)]\\
                       &\stackrel{(\sim)}{=} [\iota_{n,\norm{S(f)(s)},s}(T(f, \id{\norm{S(f)(s)}})(x))]\\
                       &\stackrel{\text{(def $T$)}}{=} [\iota_{n,\norm{S(f)(s)},s}(G(f)(x))] \stackrel{\text{(def $\psi$)}}{=} \psi(\iota_{s}(G(f)(x))).
    \end{align*}
    For $\phi$ we check that the image of pairs of the generator of $\sim$ under $\phi$ is identified by $\sim_C$:
    \begin{align*}
      \phi(\iota_{m,b,S(f)(s)}(x)) &=  [\iota_{S(f)(s)}(x)]\\
                          &\stackrel{(\sim_C)}{=} [\iota_{s}(G(f)(x))]\\
                          &\stackrel{\text{(def $T$)}}{=} [\iota_{s}(T(f, * : b \geq a)(x))] \stackrel{\text{(def $\phi$)}}{=} \phi(\iota_{n,a,s}(T(f, *: b \geq a)(x))),
    \end{align*}
    By the universal property of the quotient space, we obtain $\widetilde{\phi},\widetilde{\psi}$ from $\phi,\psi$, respectively.\smallbreak
    To see that $\widetilde{\phi}$ and $\widetilde{\psi}$ are mutually inverse, we compute:
    \[
      \widetilde{\phi} \circ \widetilde{\psi}([\iota_{s}(x)]) \stackrel{\text{(def $\widetilde{\psi}$)}}{=} \widetilde{\phi}([\iota_{n,\norm{s},s}(x)]) \stackrel{\text{(def $\widetilde{\phi}$)}}{=} [\iota_{s}(x)].
    \]
    \begin{align*}
      \widetilde{\psi} \circ \widetilde{\phi}([\iota_{n,a,s}(x)]) &\stackrel{\text{(def $\widetilde{\phi}$)}}{=} \widetilde{\psi}([\iota_{s}(x)])\\
                                                   &\stackrel{\text{(def $\widetilde{\psi}$)}}{=} [\iota_{n,\norm{s},s}(x)]\\
                                                   &\stackrel{(\sim)}{=} [\iota_{n,\norm{s},s}(T(\id{[n]}, * : a \geq \norm{s})(x))]\\
                                                   &\stackrel{\text{(def $T$)}}{=} [\iota_{n,a,s}(G(f)(x))] \stackrel{\text{(def $G$)}}{=} [\iota_{n,a,s}(G(f)(x))],\\
    \end{align*}
    where $a \geq \norm{s}$ is due to the fact that $s \in \M(S_n)^{\leq a} = \{ \, s \in S_n \; | \; \norm{s} \leq a \, \}$.
    \bigbreak
    For the action on morphisms, we must precompose with $\widetilde{\psi}$ to translate an equivalence class back into the original representation, send it through the original action on a morphism $\alpha : S \Rightarrow S'$ and then postcompose with $\widetilde{\phi}$ to translate back into the new representation. This gives
    \begin{align*}
         \widetilde{\phi} \circ (\MetRe \circ \SM(\alpha)) \circ \widetilde{\psi}([\iota_{s}(x)])
      =& \widetilde{\phi} \circ (\MetRe \circ \SM(\alpha))([\iota_{n,\norm{s},s}(x)])\\
      =& \widetilde{\phi}([\iota_{n,\norm{s},\SM(\alpha)_{n,\norm{s}}(s)}(x)])\\
      =& \widetilde{\phi}([\iota_{n,\norm{s},\M(\alpha_n)_{\norm{s}}(s)}(x)])\\
      =& \widetilde{\phi}([\iota_{n,\norm{s},\alpha_n(s)}(x)])= [\iota_{\alpha_n(s)}(x)].\\
    \end{align*}
  \end{Proof}
\end{Proposition}

We now compute the classical singular nerve $\CSing : \EPMet \rightarrow \CNSet^{\Delta^\op}$ that is left adjoint to the classical metric realization. We can directly simplify the action of this functor on an extended pseudo-metric space $E$. Then $\CSing$ constructs a simplicial classical normed set $S$, where the individual sets $S_n$ contain equivalence classes of non-expansive maps $f : \Delta^{n,a} \rightarrow E$ that map metric $n$-simplices of varying sizes $a$ into $E$. This direct simplification is still difficult to work with. However, we can identify the quotient set $S_n$ with the set that contains all Lipschitz continuous maps $f : \Delta^{n,1} \rightarrow E$ without the need for equivalence classes. This also allows us to elegantly describe the norm $\norm{f}_{S_n}$ as the best Lipschitz constant of $f$.

Before we prove this, we recall some terminology on Lipschitz continuity.
\begin{Definition}\label{Def:Lipschitz} Let $M$ and $N$ be extended pseudo-metric spaces and let $f : M \rightarrow N$ be a map.
  \begin{enumerate}
  \item A constant $c \in [0,\infty)$ is called a \defind{Lipschitz constant} for $f$ if
    \[ d_N(f(x),f(y)) \leq cd_M(x,y) \qquad \text{ for all } x,y \in M. \]
  \item The map $f$ is called \defind{Lipschitz continuous} if it has a Lipschitz constant.
  \item The smallest Lipschitz constant for a Lipschitz continuous map $f$ is called the \defind{best Lipschitz constant}.
  \item The \defind{set of Lipschitz continuous maps}
    \[ \Lip(M,N) := \{ \, f : M \rightarrow N \, | \, f \text{ is Lipschitz continuous} \,\} \]
    becomes a normed set with the norm that assigns the best Lipschitz constant
    \[ \norm{f} := \inf\{ \, a \, | \, a \text{ is a Lipschitz constant for } f \, \}. \]
  \end{enumerate}
\end{Definition}
The intuition here is that Lipschitz maps are essentially non-expansive maps where we are first allowed to stretch $M$ by some factor $c \in(0,\infty]$. Note that even though the extended pseudo-metric might assign $\infty$, we do not allow $\infty$ as a Lipschitz constant here. Otherwise all maps would be Lipschitz continuous. Also note that every Lipschitz continuous map has a best Lipschitz constant given by the infimum over all Lipschitz constants, which always exists in $[0,\infty)$ and is a Lipschitz constant itself.

\begin{Proposition} \label{Prop:CSing} The right adjoint to the functor $\CMetRe$ is the classical singular nerve $$\CSing := \SC \circ \Hom{\EPMet}{T(-,-)}{-} : \EPMet \rightarrow \CNSet^{\Delta^\op}.$$ It assigns
  \begin{itemize}
    \item to an extended pseudo-metric space $M$ the simplicial classical normed set:
      \begin{align*}
        &\CSing(M)_n = \Lip(\Delta^{n,1},M) \numberthis{eq:CSingObjObj}\\
        &\CSing(M)(f : [n] \rightarrow [m])(\sigma : \Delta^{m,1} \rightarrow M) = \sigma \circ G(f), \numberthis{eq:CSingObjMorph}
      \end{align*}
  \item to a non-expansive map $g : M \rightarrow N$ the morphism of simplicial classical normed sets
    \begin{align*}
      \CSing(g : M \rightarrow N)_n(\sigma) &= g \circ \sigma. \numberthis{eq:CSingMorph}
    \end{align*}
  \end{itemize}

  \begin{Proof}
    We begin by computing \eqref{eq:CSingObjObj}:
    \begin{align*}
      \CSing(M)_n &= (\SC(\Hom{\EPMet}{T(-,-)}{M}))_n\\
                        &= \C(\Hom{\EPMet}{T([n],-)}{M})\\
                        &=: (X,\norm{-}_X),
    \end{align*}
    where $X$ can be rewritten as
    \begin{align*}
      X &= \coprod_{a\in[0,\infty)}\Hom{\EPMet}{\Delta^{n,a}}{M}^{=a}\\
        &= \coprod_{a\in[0,\infty)} \left\{ \sigma : \Delta^{n,a} \rightarrow M \text{ non-expansive }   \; \Big| \; \parbox{6cm}{no non-expansive $\sigma' : \Delta^{n,b} \rightarrow M$ with $\sigma' = \sigma$ for $b < a$ exists} \; \right\}\\
        &= \coprod_{a\in[0,\infty)} \left\{ \sigma : \Delta^{n,a} \rightarrow M \text{ non-expansive }   \; \Big| \; \parbox{6cm}{there are $x,y \in \Delta^{n,a}$ \\with $d_M(\sigma(x),\sigma(y)) = d^{n,a}_1(x,y)$} \; \right\}\\
        &= \coprod_{a\in[0,\infty)} \left\{ \sigma : \Delta^{n,a} \rightarrow M \text{ non-expansive }        \; \Big| \; \parbox{3.5cm}{the best Lipschitz\\ constant of $\sigma$ is $1$ } \right\}\ \\
        &\simeq \coprod_{a\in[0,\infty)} \left\{ \sigma : \Delta^{n,1} \rightarrow M \text{ Lipschitz continuous } \; \Big| \; \parbox{3.5cm}{the best Lipschitz\\ constant of $\sigma$ is $a$ } \right\}\ \numberthis{eq:coprod} \\
        &\simeq \Lip(\Delta^{n,1},M).
    \end{align*}
    From this computation we see that $X$ can be rewritten as a coproduct where each simplex $\sigma : \Delta^{n,1} \rightarrow M$ only resides in the component indexed by its best Lipschitz constant \eqref{eq:coprod}. The norm $\norm{-}_X$ assigns each element the index of its component (\cref{Def:C}). Thus this norm assigns each map the best Lipschitz constant and coincides with the norm of $\Lip(\Delta^{n,1},M)$ from \cref{Def:Lipschitz}.\bigbreak
    We now establish equation \eqref{eq:CSingObjMorph}:
    \begin{align*}
      \CSing(M)(f)(\iota_a(\sigma)) &= (\SC(\Hom{\EPMet}{T(-,-)}{M}))(f)(\iota_a(\sigma))\\
                                 &= \C(\Hom{\EPMet}{T(f,-)}{M})(\iota_a(\sigma))\\
                                 &= \uparrow^a_{\CSing(M)_m} \Hom{\EPMet}{T(f,* : a \geq a)}{M}(\sigma)\\
                                 &= \uparrow^a_{\CSing(M)_m} (\sigma \circ T(f,*)) = \uparrow^a_{\CSing(M)_m} (\sigma \circ G(f)).
    \end{align*}
    Because we rewrote the sets $\CSing(M)_n$ as the isomorphic set $\Lip(\Delta^{n,1},M)$, we can remove the inclusions $\iota_a$ and the component selection map $\uparrow^a_{\CSing(M)_m}$ to obtain $\CSing(M)(f)(\sigma) = \sigma \circ T(f,*)$. \bigbreak
    
    We conclude by computing the action of $\CSing$ on morphisms \eqref{eq:CSingMorph}:
    \begin{align*}
      \CSing(g : M \rightarrow N)_n(\iota_a(\sigma)) &= (\SC(\Hom{\EPMet}{T(-,-)}{g}))_n(\iota_a(\sigma))\\
                                          &= \C(\Hom{\EPMet}{T([n],-)}{g})(\iota_a(\sigma))\\
                                          &= \uparrow^a_{\CSing(N)_n} \circ \Hom{\EPMet}{T([n],a)}{g}(\sigma)\\
                                          &= \uparrow^a_{\CSing(N)_n} (g \circ \sigma).
    \end{align*}
    As in the previous argument, we can drop $\iota_a$ and $\uparrow^a_{\CSing(N)_n}$ to obtain $$\CSing(g)_n(\sigma) = g \circ \sigma.$$
  \end{Proof}
\end{Proposition}

Finally, we compute the classical Yoneda embedding $\Cy : \Delta \times ([0,\infty), \geq) \rightarrow \CNSet^{\Delta^{\op}}$. Here, we can identify the underlying set of the normed set $y([n],a)$ with the set assigned by the ordinary Yoneda embedding $y([n])$. The norm of $y([n],a)$ simply assigns every element the value $a$. Parts of the proof are analogous to the proof of \cref{Prop:CMetRe}.

\begin{Proposition} \label{Prop:ClassicalYoneda} The \defind{classical Yoneda embedding} $$\Cy := \SC \circ y : \Delta \times ([0,\infty), \geq) \rightarrow \CNSet^{\Delta^{\op}}$$ assigns
  \begin{itemize}
  \item to an object $([m],a)$ the simplicial classical normed set:
    \begin{align*}
      &\Cy([m],a)_n = (\Hom{\Delta}{[n]}{[m]}, \norm{-}) \qquad \text{where } \norm{f} = a \numberthis{eq:CyObj}\\
      &\Cy([m],a)(f : [n] \rightarrow [k])(h) = h \circ f, \numberthis{eq:CyObjMorph}
    \end{align*}
    
  \item to a morphism $(g, *) : ([m],a) \rightarrow ([n],b)$ the morphism of simplicial classical normed sets
    \begin{align*}
      &\Cy(g,*) : \Hom{\Delta}{-}{[m]} \rightarrow \Hom{\Delta}{-}{[n]}\\
      &\Cy(g,*)_k(h : [k] \rightarrow [m]) = g \circ h. \numberthis{eq:CyMorph}
    \end{align*}
  \end{itemize}
  \begin{Proof} We begin with the proof of \eqref{eq:CyObj}:
    \begin{align*}
      \Cy([m],a)_n &= (\SC(y([m],a)))_n\\
                  &= \C(y([m],a)([n],-))\\
                  &=: (X,\norm{-}_X),
    \end{align*}
    where
    \begin{align*}
      X &= \coprod_{b\in[0,\infty)}y([m],a)([n],b)^{=b}\\
        &\simeq \coprod_{b\in[0,\infty)}\Hom{\Delta\times([0,\infty), \geq)}{([n],b)}{([m],a)}^{=b}\\
        &\stackrel{(*)}{\simeq} \Hom{\Delta\times([0,\infty), \geq)}{([n],a)}{([m],a)}\\
        &\simeq \Hom{\Delta}{[n]}{[m]}.\\
    \end{align*}
    The isomorphism $(*)$ follows, because components $\Hom{\Delta\times([0,\infty), \geq)}{([n],b)}{([m],a)}^{=b}$ of the coproduct are empty except for $b = a$ due to \cref{Def:^=}. The norm of a classical normed set constructed by $\C$ assigns each element the index of its component, thus the norm is the constant map $\norm{f}_X = a$.\\
    
    The computations for \eqref{eq:CyObjMorph} and \eqref{eq:CyMorph} are analogous to the computations for \eqref{eq:CSingObjMorph} and \eqref{eq:CSingMorph} in \cref{Prop:CMetRe}.
  \end{Proof}
\end{Proposition}

\subsection{Spivak's Metric Realization} \label{Sec:SpivakMetRe}

As mentioned at the beginning of \cref{Sec:EPMetAndMetRe}, Spivak originally defined the metric realization in terms of fuzzy sets. To define Spivak's metric realization, we must reparametrize the functor $$T : \Delta \times ([0,\infty), \ge) \rightarrow \EPMet$$ as $$T' : \Delta \times ((0,1], \le) \rightarrow \EPMet.$$

We can do this by precomposing with the isomorphism $i : ([0,\infty],\le) \rightarrow ([0,1], \ge)$ from \cref{Example:MorphismsOfLocales}, which, in our case, is just $-\log$, as we omit the bottom elements, see \cref{Rem:SimpValSet}.

\begin{Definition} \label{Def:MetReSpivak} \defind{Spivak's metric realization} $\MetRe_{-\log} : \SFuz \rightarrow \EPMet$ is the left Kan extension of $T \circ (\id{\Delta} \times (-\log))$ along the Yoneda embedding.
  \begin{center}
    \begin{tikzcd}
      \Delta \times ((0,1],\le) \arrow[rr, "T \circ (\id{\Delta} \times (-\log))"] \arrow[rd, "y'"'] & {} \arrow[d, "\eta", Rightarrow, shift right] & \EPMet \\
      & \SFuz \arrow[ru, "\MetRe_{-\log} := \Lan{y}{T \circ (\id{\Delta} \times (-\log))}"', dashed]  &
    \end{tikzcd}
  \end{center}
\end{Definition}

Similarly to \cref{Def:MetRe}, the existence is guaranteed by \cref{Cor:SimplicialKanExt}. The properties for $T \circ (\id{\Delta} \times (-\log))$ can be verified analogously to \cref{Lemma:TProperties}.

Except for the differences explained in the paragraph after \cref{Def:PseudoMetricSimplicies}, the functor $\MetRe_{-\log}$ coincides with the functor $\mathit{Re} : \text{sFuz} \rightarrow \text{UM}$ from \cite[Section~3]{Spivak2009METRICRO}.

Just as our metric realization in \cref{Prop:MetRe}, we can understand this functor better by composing it with the functor $\SM : \CVal{[0,1],\le}^{\Delta^\op} \rightarrow \SVal{[0,1],\le}$ from \cref{Thm:EquivCValSVal} to express it in terms of classical fuzzy sets.

\begin{Proposition} The classical variant of Spivak's metric realization is defined as $\CMetRe_{-\log} := \MetRe_{-\log} \circ \SM : \CVal{[0,1],\le}^{\Delta^\op} \rightarrow \EPMet$. \smallbreak It assigns to a simplicial classical fuzzy set $S : \Delta^\op \rightarrow \CFuz$ the extended pseudo-metric space
  $$\CMetRe_{-\log}(S) = \coprod_{s \in S}\Delta^{s,-\log{\mu_{S_n}(s)}}/\mathord{\sim_C},$$
  with the equivalence $\sim_C$ generated by
    \[ \iota_{S(f)(s)}(x) \sim_C \iota_{s}(G(x))  \] \bigbreak
    for all $f : [m] \rightarrow [n], \; s \in S_n, \; x \in \Delta^{s,-\log{\mu_{S_n}(s)}}$.
    
  Here $\mu_{S_n}$ is the membership strength map of the classical fuzzy set $S_n = S([n])$.
  \begin{Proof}
    The proof is analogous to the proof of \cref{Prop:MetRe}.
  \end{Proof}
\end{Proposition}

We can compare the classical variant of Spivak's metric realization and our classical metric realization from \cref{Prop:MetRe} by relating simplicial classical sets to simplicial classical normed sets. This is archived via the isomorphism $i$ that we used to reparameterize the functor in \cref{Def:MetReSpivak}.

\begin{Lemma} \label{Lem:SpivakRelationship} Let $S : \Delta^\op \rightarrow \CFuz$ be a classical fuzzy set. Then we have
  $$\CMetRe_{-\log}(S) = \CMetRe(\CVal{i}(S)),$$
  where $i$ is the isomorphism of locales from \cref{Example:MorphismsOfLocales}.
  \begin{Proof}
    By \cref{Def:IsoCVal}, we have $\norm{s}_{\CVal{i}(S)_n} = \mu_{S_n}(s)$, and thus
    \begin{align*}
      \CMetRe_{-\log}(S) &= \coprod_{s \in S_n}\Delta^{n,-\log{\mu_{S_n}(s)}}/\mathord{\sim_C}\\
                        &= \coprod_{s \in \CVal{i}(S)_n}\Delta^{n,\norm{s}_{\CVal{i}(S)_n}}/\mathord{\sim_C} = \CMetRe(\CVal{i}(S)).
    \end{align*}
    Note that by \cref{Not:Simplex} $\Delta^{n,\norm{s}_{\CVal{i}(S)_n}}$ is denoted $\Delta^{s}$ in \cref{Prop:MetRe}.
  \end{Proof}
\end{Lemma}

Ultimately, the difference between Spivak's and our metric realization is only how we interpret the result but mathematically immaterial.

When we realize a normed set using our metric realization from \cref{Def:MetRe}, the norm of each element dictates the size of the corresponding simplex in the metric realization.

When we realize a fuzzy set using Spivak's metric realization from \cref{Def:MetReSpivak}, the membership strength dictates ``closeness''. Two points connected by a 1-simplex will be closer if the membership strength is larger. At membership strength 1, both points will be identified. If we reduce the membership strength, then the two points will be increasingly farther apart the closer we approach 0.

The choice of the negative logarithm or the interval $(0,1]$ is not canonical. Any other isomorphism of locales $\phi : ([0,\infty], \ge) \rightarrow (\mathcal{L},\preceq)$ would grant us yet another parametrization $\MetRe_{\phi}(S)$ of our metric realization from \cref{Def:MetRe} with $$\CMetRe_{\phi}(S) = \CMetRe(\CVal{\phi}(S)).$$

All computations from \cref{Subsection:ClassicalMetricRealization} could be done with every other parametrization as well. However, this would lead to more complicated formulas, because the reparametrization by $\phi$ causes $\phi$ to appear in the coequalizer. Our parametrization in terms of normed sets is the only one which avoids this, so it could be argued to be the most canonical choice, or at least the simplest choice.

\subsection{The Finite Metric Realization} \label{Sec:FinMetRe}

McInnes et al. \cite[Theorem 2]{UMAP} construct a finite variant of Spivak's metric realization from \cref{Def:MetReSpivak}. We will attempt to reproduce this result. The finite metric realization we construct in this section could equivalently be defined in terms of normed sets or any other parametrization obtained through another isomorphic locale as explained after \cref{Lem:SpivakRelationship}. We intend to discuss the claimed relationship between this functor and the UMAP algorithm in \cref{Subsection:Discussion}, so we will use the same parametrization as McInnes et al.

For the finite metric realization, McInnes et al. introduce finite variants of the relevant categories.

The first finite variant is the category of finite extended pseudo-metric spaces, introduced in the paragraph just after \cite[Definition 6]{UMAP}. Their definition does not precisely state that finiteness in this case refers to finiteness of the underlying set, but this definition seems to be compatible with the other results from McInnes et al. \cite{UMAP}, so we will use this definition.

\begin{Definition} \label{Def:FinEpMet} The \defind{category of finite extended pseudo-metric spaces} \defind{$\Fin{\EPMet}$} is the full subcategory of \defind{$\EPMet$} where the underlying set is finite.
  
\end{Definition}

When we restrict ourselves to $\Fin{\EPMet}$, we lose infinite cocompleteness, which was required to show the existence of Spivak's (\cref{Def:MetReSpivak}) and our (\cref{Def:MetRe}) metric realization.

\begin{Remark} \label{Rem:FinEPMetCocompleteness} $\Fin{\EPMet}$ is finitely cocomplete, but not cocomplete.
  
  It is easy to see that coequalizers can still be constructed in $\Fin{\EPMet}$, because an equivalence relation can only decrease the number of points but not increase it. However, coproducts can only be constructed in $\Fin{\EPMet}$ if the index set for the components is finite. An obvious counterexample is the coproduct $\coprod_{n \in \mathbb{N}} \{ * \}$ where $\{ * \}$ is the finite extended pseudo-metric space that contains a single point with distance of 0 to itself.

  Some colimits over infinite diagrams might still exist if the coequalizer identifies enough points such that only finitely many equivalence classes remain in the colimit object. However, the standard construction from \cite[Theorem~3.4.12]{riehl2017category} of colimits from coequalizers is only guaranteed to work if the diagram category is finite.
\end{Remark}

McInnes et al. also introduce a finite variant of the category $\SFuz$ from \cref{Def:SimpValSetsSpecialCases} (1) in the paragraph just before \cite[Definition 7]{UMAP}. Here McInnes et al. only state that this category is defined as the subcategory of objects in $\SFuz$ which are ``bounded'', but do not define what makes an object in $\SFuz$ ``bounded''. The following is our interpretation of this definition, again, with the goal of obtaining similar results to \cite{UMAP}.

\begin{Definition} \label{Def:FinSFuz} \,
  \begin{itemize}
  \item A fuzzy set $X : ((0,1],\le)^{\op} \rightarrow \SET$ is called finite, if there exists an upper bound $B_X$ such that all sets $X^{\ge a}$ for $a \in (0,1]$ have a cardinality less than $B_X$.
    
    The \defind{category of finite fuzzy sets} \defind{$\Fin{\Fuz}$} is the full subcategory of $\Fuz$ that has finite fuzzy sets as objects.
    
  \item An uncurried simplicial fuzzy set $S : (\Delta \times ((0,1], \le))^{\op} \rightarrow \SET$ is called finite if  $S([n],-) : (0,1], \le)^\op \rightarrow \SET$ is finite for every $n \in \mathbb{N}_0$.

    The \defind{category of finite uncurried simplicial fuzzy sets} \defind{$\Fin{\SFuz}$} is the full subcategory of $\SFuz$ that has finite uncurried simplicial fuzzy sets as objects.
  \end{itemize}
\end{Definition}

We will later need a key property of finite uncurried simplicial fuzzy sets to verify the existence of McInnes et al's finite metric realization in \cref{Def:FinMetRe}: For some fuzzy sets $X$, we cannot see all elements in $X$ by looking at a set $X^{\ge a}$ for sufficiently small $a$, as shown in \cref{Example:MActsOnFuzzy}. However, for finite classical fuzzy sets, this is always the case.

We will formally capture this in \cref{Lem:FinSFuzIso}. For this, we first introduce definitions for the smallest admissible upper bound $L_X$ of the cardinalities of the sets $X^{\ge a}$ and the largest value $l_X$ for which the sets $X^{\ge a}$ with $a \le l_X$ do not grow anymore.

\begin{Definition} \label{Def:FinSFuzBounds} Let $X : ((0,1],\le)^{\op} \rightarrow \SET$ be a finite fuzzy set. We denote
  \begin{itemize}
  \item $L_X := \max \{ \, | X^{\ge a} | \quad | \quad a \in (0,1] \, \} \in \mathbb{N}_0,$
  \item $l_X := \max \{ a \, | \, a \in (0,1], \; |X^{\ge a}| = L_X \} \in (0,1].$
  \end{itemize}
\end{Definition}

For finite fuzzy sets $X$, the set $X^{\le l_X}$ contains all information about $X$, because all sets $X^{\le a}$ for $a \le l_X$ are isomorphic to $X^{\le l_X}$.

\begin{Lemma} \label{Lem:FinSFuzIso} Let $X : ((0,1],\le)^{\op} \rightarrow \SET$ be a finite fuzzy set. \smallbreak Then $X(* : a \le l_X) : X(l_X) \rightarrow X(a)$ is an isomorphism for all $a \le l_X$.
  \begin{Proof}
    Thus, by \cref{Def:SheafValuedSets}, $X(* : a \le l_X)$ is injective. By \cref{Def:FinSFuzBounds}, the set $X(a)$ has a cardinality of at most $L_X$. Because $X(l_X)$ has exactly the cardinality $L_X$, the injection $X(* : a \le l_X) : X(l_X) \rightarrow X(a)$ is an isomorphism. 
  \end{Proof}
\end{Lemma}

We have now defined all relevant finite versions of the relevant categories. The next step to reproduce McInnes et al.'s finite metric realization is to define analogous finite variants of the Yoneda embedding and the functor $T$ from \cref{Def:T}.

\begin{Remark} The Yoneda embedding $y : \Delta \times ((0,1], \le) \rightarrow \SET^{(\Delta \times ((0,1], \le))^\op}$ takes values in $\Fin{\SFuz}$ and thus defines a functor $y : \Delta \times ((0,1], \le) \rightarrow \Fin{\SFuz}$.

  The hom sets $\Hom{\Delta\times((0,1], \le)}{([m],b)}{([n],a)}$ consist of pairs $(f : [m] \rightarrow [n], * : b \le a)$ where $*$ is an element of a singleton set. There are only finitely many maps $f : [m] \rightarrow [n]$ for any $m$ and $n$ in the simplex category, because $[m]$ and $[n]$ are finite sets.
 
\end{Remark}

McInnes et al. directly define a functor $F : \Delta \times I \rightarrow \Fin{\SFuz}$ in \cite[Appendix B, Theorem 2]{UMAP} to construct the finite metric realization as a left Kan extension of $F$. We will attempt to decompose this functor into finite metric simplices to make it more similar to the analogous functor in \cref{Def:T} which we used to construct the metric realization.

\begin{Definition} \label{Def:FiniteMetricSimplices} \;
  For $n \in \mathbb{N}_0$ and $a \in [0,\infty)$, the \defind{finite metric $n$-simplex of size $a$} is the finite extended pseudo-metric space
  \[ \Fin{\Delta^{n,a}} = ([n],d_a), \]
  where the metric $d_a : [n] \times [n] \rightarrow [0,\infty)$ is the discrete metric scaled by $a$,
  \[ d_a(i,j) :=       \begin{cases}
    a & i \neq j\\
    0 & i = j.
  \end{cases} \]
\end{Definition}

\begin{Lemma} \cite[Theorem 2]{UMAP} \label{Def:F} The finite metric $n$-simplices define a functor $F : \Delta \times ((0,1], \le) \rightarrow \Fin{\EPMet}$ by
  \begin{align*}
    &F([n],a) = \Fin{\Delta^{n,-\log(a)}}\\
    &F(f,* : a \le b) = f.
  \end{align*}
where $F(f,* : a \le b)$ is non-expansive because the metric $d_{-\log(a)}$ produces larger distances than the metric $d_{-\log(b)}$, because we have $-\log(a) \ge -\log(b)$ for $a \le b$.
\end{Lemma}

Our definition of $F$ differs slightly from McInnes et al. \cite[Appendix B, Theorem 2]{UMAP}. They define the underlying set of $F([n],a)$ as some set with the appropriate number of points, while we specifically map to objects in the simplex category, which are sets with the same number of points. Our variant allows us to reuse the face and degeneracy maps in $\Delta$ as face and degeneracy maps for the finite metric $n$-simplices, so we do not have to define them separately.

We can now define McInnes et al.'s finite metric realization as a Kan extension of $F$. The proofs for Spivak's (\cref{Def:MetRe}) and our (\cref{Def:MetRe}) metric realization require cocompleteness of the target category. This property is lost in the case of $\Fin{\MetRe}$, as discussed in \cref{Rem:FinEPMetCocompleteness}. However, we can still show the existence of this Kan extension by ensuring that the necessary colimits for the colimit formula \cref{Thm:ColimitFormula} exist.

\begin{Proposition} \cite[Appendix B, Theorem 2]{UMAP} \label{Def:FinMetRe} \defind{McInnes et al's finite metric realization} $$\Fin{\MetRe_{-\log}} = \Lan{y}{F} : \Fin{\SFuz} \rightarrow \Fin{\EPMet}$$ is the left Kan extension of $F$ along $y : \Delta \times ((0,1], \le) \rightarrow \Fin{\SFuz}$\\
  \begin{center}
    \begin{tikzcd}
      \Delta \times ((0,1], \le) \arrow[rr, "F"] \arrow[rd, "y"'] & {} \arrow[d, "\eta", Rightarrow, shift right] & \Fin{\EPMet} \\
      & \Fin{\SFuz} \arrow[ru, "\Lan{y}{F} = \Fin{\MetRe_{-\log}}"', dashed]  &
    \end{tikzcd}.
  \end{center}
  It is left adjoint to the finite singular nerve $$\Fin{\Sing_{-\log}} = \Hom{\Fin{\EPMet}}{F(-,-)}{-} : \Fin{\EPMet} \rightarrow \Fin{\SFuz}.$$
  \begin{Proof}
    We first show that this Kan extension exists by showing that the required colimits
    \[ \colim{FP^{S} : y \downarrow S \longrightarrow \Delta \times ((0,1], \le) \longrightarrow \Fin{\EPMet}} \]
    
    in \cref{Thm:ColimitFormula} exist. We will construct these colimits in $\EPMet$ and then show that the resulting colimit object is also a colimit in $\Fin{\EPMet}$ by \cref{Lem:ColimitSubcat}.
    
    As $\EPMet$ is cocomplete, we can rewrite the colimits $\colimNotScaling{FP^{S}}$ with \cref{Lem:CoequalizerFormula}, where we consider $F$ as a functor $F : \Delta \times ((0,1], \le) \rightarrow \EPMet$. We obtain a coequalizer in $\EPMet$ that can be expressed as the quotient
    \begin{align*}
      \coprod_{\substack{n \in \mathbb{N}_0, \; a \in [0,\infty)\\s \in S([n],a)}}\Fin{\Delta}^{n,-\log(a)}/\mathord{\sim}, \numberthis{eq:FinMetReQuotient}
    \end{align*}
    where the equivalence relation $\sim$ is generated by
    \[ \iota_{m,b,S(f,* : b \le a)(s)}(x) \sim \iota_{n,a,s}(F(f, * : b \le a)(x)), \]
    with the canonical inclusion morphisms
    \[ \iota_{n,a,s} : \Fin{\Delta}^{n,a} \rightarrow  \coprod_{\substack{n \in \mathbb{N}_0, \; a \in [0,\infty)\\s \in S([n],a)}} \Fin{\Delta}^{n,-\log(a)}. \]
    
    The coproduct used in this expression to form the colimit object may have infinitely many components, so it is not necessarily in $\Fin{\EPMet}$. However, by \cref{Lem:ColimitSubcat}, if this colimit object we formed in $\EPMet$ lies in $\Fin{\EPMet}$, the object is also a colimit of $FP^{S}$ in the full subcategory $\Fin{\EPMet}$.
    
    To show that the object from \eqref{eq:FinMetReQuotient} lies in $\Fin{\EPMet}$, we show that the number of equivalence classes is finite.

    For this, consider the finite subset
    $$U \; := \quad \coprod_{\substack{s \in S([0],l)}} \Fin{\Delta}^{0,-\log(l)} \qquad \subseteq \qquad \coprod_{\substack{n \in \mathbb{N}_0, \; a \in [0,\infty)\\s \in S([n],a)}} \Fin{\Delta}^{n,-\log(a)} \quad =: \; M,$$
    where $l := l_{S([0],-)}$ is the value below which all sets $S([0],a)$ are isomorphic as shown in \cref{Lem:FinSFuzIso}.
    
    The subset $U \subseteq M$ is finite because the metric $0$-simplices of any size only contain a single point and because the index set $S([0],l)$ is finite by \cref{Def:FinSFuz}.

    We prove that $M /\mathord{\sim}$ is finite by showing that every point $p \in M$ is identified with a point in $U$.

    Let $p \in M$. Then $p$ is of the form $p = \iota_{n,a,s}(i)$. We can identify $p$ with a point $\iota_{0,a,s'}(0)$ indexed by an element $s' \in S([0],a)$ by
    \begin{align*}
      \iota_{n,a,s}(i) = \iota_{n,a,s}(f(0)) = \iota_{n,a,s}(F(f, * : a \le a)(0)) \sim \iota_{0,a,S(f,*)(s)}(0),
    \end{align*}
    where $f : [0] \rightarrow [n]$ is the morphism that sends 0 to $i$ and $s' := S(f,*)(s)$.
    By case distinction, we show that $\iota_{0,a,s'}(0)$ can be identified with a point in $U$:
    \begin{itemize}
      \item If $l \le a$, we have
      \begin{align*}
        \iota_{0,a,s'}(0) = \iota_{0,a,s'}(F(\id{0},* : l \le a)(0)) \sim \iota_{0,l,S(f,*)(s')}(0).
      \end{align*}
      \item If $a \le l$, by \cref{Lem:FinSFuzIso}, $S(\id{0},* : a \le l)$ is an isomorphism and thus there exists an element $s' \in S([0], l)$ with $S(\id{0},* : a \le l)(s'') = s'$. Then we have
      \begin{align*}
        &\iota_{0,a,s'}(0) = \iota_{0,a,S(\id{0},* : a \le l)(s'')}(0) \sim \iota_{0,l,s''}(F(\id{0},* : a \le l)(0)) = \iota_{0,l,s''}(0).
      \end{align*}
    \end{itemize}
    \bigbreak
    For the adjoint, we can apply \cref{Prop:KanExtYonedaSubcat}. The required properties of $F$ are easy to verify similarly to \cref{Lemma:TProperties}.
  \end{Proof}
\end{Proposition}

We can describe this functor and its adjoint in terms of classical fuzzy sets analogously to our metric realization in \cref{Subsection:ClassicalMetricRealization}.

This requires a finite variant of the category $\CFuz$, as well as finite variants of the functors $\SC : \SFuz \rightarrow \CFuz^{\Delta^\op}$ and $\SM : \CFuz^{\Delta^\op} \rightarrow \SFuz$. For the category of finite classical fuzzy sets, we demand that the underlying sets are finite.

\begin{Definition} \label{Def:FinClassical} \;
  \begin{itemize}
  \item A \defind{finite classical fuzzy set} is a classical fuzzy set $(X,\mu_X)$ where $X$ is finite. The \defind{category of finite classical fuzzy sets} $\Fin{\CFuz}$ is the full subcategory of $\CFuz$ with finite classical fuzzy sets as objects.
  \item A \defind{simplicial finite classical fuzzy set} is a simplicial object in $\Fin{\CFuz}$. The \defind{category of simplicial finite classical fuzzy sets} $\Fin{\CFuz}^{\Delta^\op}$ is the category of simplicial objects in $\Fin{\CFuz}$.
  \end{itemize}
\end{Definition}

The next step is to construct analogous finite variants of the equivalence from \cref{Thm:FinEquivCM} and the equivalence from \cref{Thm:EquivCValSVal}.

Instead of constructing these equivalences from scratch, we can show that the established equivalences from \cref{Thm:EquivCM} and \cref{Thm:EquivCValSVal} can be restricted to the finite variants of the involved categories.

\begin{Theorem} \label{Thm:FinEquivCM} The equivalence of categories
  \begin{center}
    \begin{tikzcd}[column sep=5em] 
      \CFuz \arrow[r, "\M", shift left] & \Fuz \arrow[l, "\C", shift left]
    \end{tikzcd}
  \end{center}
  from \cref{Thm:EquivCM} can be restricted to the equivalence
  \begin{center}
    \begin{tikzcd}[column sep=5em] 
      \Fin{\CFuz} \arrow[r, "\Fin{\M}", shift left] & \Fin{\Fuz} \arrow[l, "\Fin{\C}", shift left]
    \end{tikzcd}.
  \end{center}
  where $\Fin{\M}$ and $\Fin{\C}$ are the respective restrictions and corestrictions of $\M$ and $\C$.
  
  \begin{Proof}
    We verify that the functors $\M$ from \cref{Def:M} and $\C$ from \cref{Def:C} can be restricted and corestricted to $\Fin{\CFuz}$ and $\Fin{\Fuz}$.
    \begin{itemize}
    \item Let $(X,\mu_X)$ be a finite classical fuzzy set. Then $\M(X,\mu_X)$ is a finite fuzzy set, because the sets $\M(X,\mu_X)^{\ge a} = X$ are finite for every $a \in (0,1]$.
    \item Let $Y : \Delta \times ((0,1], \le)^\op \rightarrow \SET$ be a finite fuzzy set. By \cref{Def:C}, the underlying set of $\C(Y)$ is given by
      $$\coprod_{a \in (0,1]} Y^{=a}$$
      By \cref{Lem:FinSFuzIso}, all restriction maps $Y(* : a \le l_Y) : Y(l_Y) \rightarrow Y(a)$ for $a \le l_Y$ are isomorphisms. Thus, by \cref{Def:^=}, the sets $Y^{=a}$ are empty for $a \le l_Y$.
      Then the underlying set $\C(Y)$ can be rewritten to
      $$\coprod_{a \in (0,1]} Y^{=a} \cong \coprod_{a \in (l_Y,1]} Y^{=a} \cong Y^{\ge l_Y}.$$
      As $Y^{\ge l_Y}$ is finite by assumption, the underlying set of $\C(Y)$ is finite and thus $\C(Y)$ is a finite classical fuzzy set.
    \end{itemize}
  \end{Proof}
\end{Theorem}

\begin{Lemma} \label{Cor:RestrictedEquiv} The equivalence of categories
  \begin{center}
    \begin{tikzcd}[column sep=5em] 
      \CFuz^{\Delta^\op} \arrow[r, "\SM", shift left] & \SFuz \arrow[l, "\SC", shift left]
    \end{tikzcd}
  \end{center}
  from \cref{Thm:EquivCValSVal} for $\mathcal{L} = ([0,1],\le)$ can be restricted to the equivalence
  \begin{center}
    \begin{tikzcd}[column sep=5em] 
      \Fin{\CFuz^{\Delta^\op}} \arrow[r, "\Fin{\SM}", shift left] & \Fin{\SFuz} \arrow[l, "\Fin{\SC}", shift left]
    \end{tikzcd}.
  \end{center}
  where $\Fin{\SM}$ and $\Fin{\SC}$ are the respective restrictions and corestrictions of $\SM$ and $\SC$.
  \begin{Proof}
    The equivalence from \cref{Thm:EquivCValSVal} was obtained by composing the equivalence from \cref{Thm:EquivCM} with the isomorphism of categories from \cref{Lem:Curry}.

    We showed that the equivalence from \cref{Thm:EquivCM} can be restricted to the desired categories in \cref{Thm:FinEquivCM}.

    It is easy to verify that the isomorphism of categories \cref{Lem:Curry} can also be restricted to the desired finite categories.
  \end{Proof}
\end{Lemma}

With the restricted equivalence from \cref{Cor:RestrictedEquiv}, we can express and compute analogous results to \cref{Prop:CMetRe} and \cref{Prop:CSing} for McInnes et al's finite metric realization from \cref{Def:FinMetRe}.

\begin{Proposition} The classical variant of McInnes et al's finite realization is the composite
  $$\Fin{\CMetRe_{-\log}} \; := \; \Fin{\MetRe_{-\log}} \circ \Fin{\M} \; : \; \Fin{\CFuz^{\Delta^\op}} \longrightarrow \Fin{\EPMet}.$$
  It assigns to a simplicial finite classical fuzzy set $S : \Delta^\op \rightarrow \SET$ the finite extended pseudo metric space
  $$\CMetRe_{-\log}(S) = \coprod_{s \in S}\Fin{\Delta}^{s,-\log(\mu_{S_n}(s))}/\mathord{\sim_F},$$
  with the equivalence $\sim_F$ generated by
  \[ \iota_{S(f)(s)}(i) \sim_F \iota_{s}(F(f)(i))  \] \bigbreak
  for all $f : [m] \rightarrow [n], \; s \in S_n, \; i \in \Fin{\Delta}^{s,-\log(\mu_{S_n}(s))}$.
  \begin{Proof}
    Analogous to the proof of \cref{Prop:CMetRe}.
  \end{Proof}
\end{Proposition}

\begin{Proposition} \label{Prop:CSing} The classical variant of McInnes et al's finite singular nerve is the composite
  $$\Fin{\CSing_{-\log}} \; := \; \Fin{\C} \circ \Fin{\Sing_{-\log}} \; : \; \Fin{\EPMet}  \longrightarrow \Fin{\CFuz^{\Delta^\op}}.$$
  It assigns to an extended pseudo-metric space $M$ the simplicial classical fuzzy set:
  \begin{align*}
    &\Fin{\CSing_{-\log}}(M)_n = (\Lip(\Fin{\Delta}^{n,1},M),\mu) \qquad \text{where } \mu(\sigma) = \exp(-\norm{\sigma})\\
    &\Fin{\CSing_{-\log}}(M)(f : [n] \rightarrow [m])(\sigma : \Fin{\Delta}^{m,1} \rightarrow M) = \sigma \circ F(f),
  \end{align*}
  where $\norm{\sigma}$ denotes the best Lipschitz constant of the map $\sigma : \Fin{\Delta}^{n,1} \rightarrow M$.
  \begin{Proof}
    Analogous to the proof of \cref{Prop:CSing}. In this case, we obtain the membership strength $\exp(-\norm{\sigma})$ due to
    \begin{align*}
      \Fin{\CSing_{-\log}}(M)_n &= (\Fin{\SC}(\Hom{\Fin{\EPMet}}{F(-,-)}{M}))_n\\
                        &= \Fin{\C}(\Hom{\Fin{\EPMet}}{F([n],-)}{M})\\
                        &=: (X,\mu_X),
    \end{align*}
    where $X$ can be rewritten as
    \begin{align*}
         & \coprod_{a\in[0,\infty)}\Hom{\Fin{\EPMet}}{\Fin{\Delta}^{n,-\log(a)}}{M}^{=a}\\
        \simeq& \coprod_{a\in[0,\infty)} \left\{ \sigma : \Fin{\Delta}^{n,1} \rightarrow M \text{ Lipschitz cont. } \; \Big| \; \parbox{4.5cm}{the best Lipschitz\\ constant of $\sigma$ is $-\log(a)$ } \right\} \\
        \simeq& \Lip(\Delta^{n,1},M).
    \end{align*}
    In this case, maps $\sigma : \Fin{\Delta}^{n,1} \rightarrow M$ with the best Lipschitz constant $-\log(a)$ reside in the component indexed by $a$. By definition of $\Fin{\C}$ as a restriction of $\C$, the membership strength $\mu_X$ assigns each element the index of its component. Thus a map $\sigma : \Fin{\Delta}^{n,1} \rightarrow M$ with the best Lipschitz constant $c = -\log(exp(-c))$ will be assigned the membership strength $\exp(-c)$. 
  \end{Proof}
\end{Proposition}

\Cref{Prop:CSing} reveals that the set $\Fin{\CSing_{-\log}}(M)_1$ collects all finite distances in $M$.

\begin{Remark} \label{Rem:OneSkeleton} Let $M$ be a finite extended pseudo metric space.

  The elements of $\Fin{\CSing_{-\log}}(M)_1$ are maps $\sigma : \Fin{\Delta}^{m,1} \rightarrow M$ with $\sigma(1) = x$ and $\sigma(0) = y$, where $\mu(\sigma) = \exp(-d_M(x,y))$ for every pair $x,y \in M$ with $d_M(x,y) \neq \infty$.
  
\end{Remark}


\clearpage
\section{The Theory Behind UMAP?} \label{Sec:UMAP}

In this section we will comment on the relationship between UMAP and the finite metric realization as claimed by \cite{UMAP}.

We begin by describing the general idea of dimensionality reduction algorithms in \cref{Subsection:DimensionReduction}.

As the UMAP algorithm uses undirected weighted graphs as a data structure in the intermediate computational steps, we introduce them in \cref{Subsection:WeightedGraphs} and discuss their relationship to fuzzy sets. We then describe the algorithm in \cref{Subsection:UMAPAlgo} and discuss the claimed relationship to the finite metric realization in \cref{Subsection:Discussion}.

\subsection{Dimensionality Reduction Algorithms} \label{Subsection:DimensionReduction}

The UMAP algorithm is a dimensionality reduction algorithm. An overview of topological dimensionality reduction algorithms can be found in \cite{vandermaaten2009dimensionality}. In general, a dimensionality reduction algorithm accepts an input dataset $X = \{ x_1, \ldots, x_k \} \subseteq \mathbb{R}^D$ and possibly additional parameters. It outputs a dataset $Y = \{ y_1, \ldots, y_k \} \subseteq \mathbb{R}^d$ where $d < D$, while attempting to preserve properties of interest. What it means to preserve properties of interest is not always formally stated and depends on the application.

For example, one could demand that clusters of points in $X$ should result in clusters of the corresponding points in $Y$.

One could also demand that topological features are preserved. For example, if the input points in $X \subseteq \mathbb{R}^{100}$ are distributed roughly along a circle $S^1$ embedded in $\mathbb{R}^{100}$, one could demand that the algorithm produces a set of output points $Y \subseteq \mathbb{R}^2$ also distributed along a circle $S^1$, embedded in $\mathbb{R}^2$.

Additionally, we could demand the preservation of geometric features. For example, if the input points are distributed along an ellipse in $\mathbb{R}^{100}$, one might demand that the output points are also distributed along an ellipse while preserving the eccentricity of the original ellipse.

Most algorithms only preserve these conditions if we already know that the input dataset $X$ satisfies specific properties necessary for the specific algorithm.

\subsection{Weighted Graphs and Fuzzy Unions} \label{Subsection:WeightedGraphs}

A graph consists of a set of vertices, where each pair of vertices may or may not be connected by an edge. In our case these edges are not directed. The edges of a graph can be equipped with additional information, which is usually referred to as weighting the edges.

\begin{Definition} An \defind{undirected $W$-weighted graph $(V,E,w_E)$} consists of
  \begin{itemize}
    \item a set $V$ called vertices,
    \item a subset $E \subseteq \{ \{ u,v \} \, | \, u,v \in V, u \neq v \}$,
    \item and a weight map $w_E : E \rightarrow W$ with values in a set $W$.
  \end{itemize}
\end{Definition}

Weighted graphs are commonly used in many algorithms. For example, a route finding algorithm may use graphs to represent a network of roads where the vertices are intersections, the edges are roads, and the weights correspond to the lengths of the roads. Weighted graphs can also relate to non-geometric data. In a social network, a graph could use the vertices to represent users and the edges to model interactions between them, where the weights correspond to the number of interactions.

The UMAP algorithm uses $(0,1]$-weighted graphs in intermediate computational steps. These graphs relate to classical fuzzy sets.

\begin{Remark} \label{Rem:Graph} Any undirected $(0,1]$-weighted graph $(V,E,w_E)$ defines a classical fuzzy set $(E,w_E)$.
\end{Remark}

One step of the UMAP algorithm requires forming the union of such graphs. This union operation is inspired by the unions of classical fuzzy sets. To define a union of classical fuzzy sets $(A,\mu_1)$ and $(A,\mu_2)$ with the same underlying set, we must decide a way to compute the membership strength of an element $x$ from the possibly different membership strengths in $\mu_1(x)$ and $\mu_2(x)$. We also want the union of fuzzy sets to satisfy properties similar to the union of ordinary sets such as symmetry or associativity. One way to define such unions is via $T$-conorms.
\begin{Definition} \label{Def:Conorm} \cite[Section~3.4]{FuzzyLogic} A \defind{$T$-conorm}, also called \defind{$S$-norm}, is a map $c : [0,1] \times [0,1] \rightarrow [0,1]$ satisfying the following axioms for $x,y,z \in [0,1]$:
  \begin{itemize}
  \item $c(x,y) = c(y,x)$ (symmetry).
  \item $c(x,y) \le c(z,y) \text{ for } x \le z$ (monotonicity).
  \item $c(x,c(y,z)) = c(c(x,y),z)$ (associativity).
  \item $c(x,0) = x$ (boundary condition).
  \end{itemize}
\end{Definition}

\begin{Example} \cite[Section~3.4]{FuzzyLogic} \; \label{Examples:Conorm}
  \begin{itemize}
  \item The maximum map $\max : [0,1] \times [0,1] \rightarrow [0,1]$ is a T-conorm.
  \item The algebraic sum $P : [0,1] \times [0,1] \rightarrow [0,1]$ is given by
    $$P(x,y) = x + y - xy.$$
    This conorm is referred to as the probabilistic $T$-conorm by McInnes et al. \cite{UMAP}.
  \end{itemize}
\end{Example}

A union of fuzzy sets $(A,\mu_1)$ and $(A,\mu_2)$ with the same underlying set $A$ can then be defined by assigning each element $x \in A$ the membership strength $c(\mu_1(x), \mu_2(x))$, where $c : [0,1] \times [0,1] \rightarrow [0,1]$ is any conorm.

\begin{Definition} \cite[Section~3.4]{FuzzyLogic} \label{Def:FuzzySetUnion} The \defind{binary union of classical fuzzy sets} $(A,\mu_1)$ and $(A,\mu_2)$ with identical underlying sets $A$, with respect to the conorm $c : [0,1] \times [0,1] \rightarrow [0,1]$ is defined by
  \[ (A,\mu_1) \cup_c (A,\mu_2) := (A, \mu_{c}), \]
  where the membership strength map $\mu_{c} : A \rightarrow (0,1]$ is given by
  \[ \mu_c(x) = c(\mu_1(x), \mu_2(x)). \]
\end{Definition}

The reason why $T$-conorms are defined as in \cref{Def:Conorm} is that the properties of T-conorms result in the following properties of the union of fuzzy sets.

\begin{Remark} Let $A,B,C$ be fuzzy sets with identical underlying sets and let $c$ be a $T$-conorm. Then we have
  \begin{align*}
    A \cup_c B &= B \cup_c A && \text{(symmetry)}\\
    A \cup_c (B \cup_c C) &= (A \cup_c B) \cup_c C && \text{(associativity)}\\
    A \cup_c \emptyset &= A && \text{(identity element)}
  \end{align*}
\end{Remark}

The monotonicity condition ensures that this union is compatible with classical fuzzy subsets. However, we will not need classical fuzzy subsets in this thesis.

We can define a similar union operation for undirected $(0,1]$-weighted graphs.

\begin{Definition} \label{Def:GraphUnion} The \defind{binary union of two undirected $(0,1]$-weighted graphs} with identical vertices and edges $G = (V,E,w_1)$ and $G' = (V,E,w_2)$ with respect to the conorm $c : [0,1] \times [0,1] \rightarrow [0,1]$ is defined by
  \[ (V,E,w_1) \cup_c (V,E,w_2) := (V, E, w_c), \]
  where the weight map $w_c : E \rightarrow (0,1]$ is given by
  \[ w_c(e) = c(w_1(e), w_2(e)). \]
\end{Definition}

This union of undirected $(0,1]$-weighted graphs coincides with the union of the classical fuzzy sets induced by graphs from \cref{Rem:Graph}.

\subsection{The UMAP Algorithm} \label{Subsection:UMAPAlgo}

The algorithm UMAP is motivated and described in \cite[Section~3]{UMAP}. A pseudocode implementation is provided in \cite[Section~4.1]{UMAP}. 

The UMAP algorithm is a type of dimensionality reduction algorithm that mostly seeks to preserve topological structure. Given an input dataset $X  \subseteq \mathbb{R}^D$ that has been sampled from a Riemannian manifold $\mathcal{M}$ embedded in  $\mathbb{R}^D$, the output dataset $Y \subseteq \mathbb{R}^d$ should preserve the topological structure of $\mathcal{M}$ as much as possible. What it means to preserve the structure of $\mathcal{M}$ is not formally defined.

In this section, we sketch the computational steps of the algorithm. Afterwards, we will discuss the claims regarding the preservation of structure throughout each step of the algorithm in \cref{Subsection:Discussion}.

To run the algorithm, we must provide the input dataset $X = \{ x_1, \ldots, x_N \} \subseteq \mathbb{R}^D$ and a target dimension $d < D$. In addition, we must select parameters $k \in \mathbb{N}$ and $n\mathit{-epochs} \in \mathbb{N}$ \cite[Algorithm 1]{UMAP}. UMAP then performs multiple steps to compute the output set $Y = \{ y_1, \ldots, y_N \} \subseteq \mathbb{R}^d$.

The main procedure of the UMAP algorithm is described in \cite[Algorithm~1]{UMAP}. It consists of the following steps, some of which are implemented as subroutines in \cite[Algorithm~2-5]{UMAP}.

\begin{enumerate}
  \item \cite[Algorithm 2]{UMAP} For each input point $x_i \in X$, we first construct a local undirected $(0,1]$-weighted graph $G_i$ where
  \begin{itemize}
  \item the vertices of $G_i$ are $X$,
  \item the edges of $G_i$ are all edges from $x_i$ to every other $x_j \in X$,
  \item the weight of an edge between $x_i$ and another vertex $x_j \in X$ is set to $$\exp(-d_\mathcal{M}(x_i,x_j)-\rho) \in (0,1],$$ where $\rho$ is the distance between $x_i$ and the nearest point and $d_\mathcal{M}$ is an approximation of the metric on $\mathcal{M}$ described in \cite[Section~2.1]{UMAP} in terms of the approximated distances to the $k$ nearest neighbors. 
  \end{itemize}
  
\item \label{Step:Union} \cite[Algorithm 1]{UMAP} We then form the union $G := \bigcup_{1\le i \le N,P} G_i$ over all local graphs with respect to the probabilistic conorm $P$ from \cref{Examples:Conorm}.
  
  The input dataset is finite, so this union can be expressed as $G_1 \cup_P \cdots \cup_P G_N$, with the binary union from \cref{Def:GraphUnion}.
  
\item \cite[Algorithm 4]{UMAP} We then use spectral embedding to embed the vertices of $G$ into $\mathbb{R}^{d}$ with respect to the weights. This yields a set of coordinates $Y_0$ for the vertices of $G$.

  Spectral embedding is a well-known algorithm that can be used to embed the vertices of a weighted graph into any $\mathbb{R}^n$ for an $n \in \mathbb{N}$ of our choice such that the weights of the edges translate roughly to the distance of the points in $\mathbb{R}^n$ for some metric. The vertices are not required to have coordinates to run the spectral embedding algorithm, because the algorithm treats the vertices purely as labels and generates coordinates.

  In the case of UMAP, this means that the spectral embedding step disregards the original coordinates of the vertices $X$ in $G$ and generates entirely new coordinates while attempting to place vertices connected with an edge of weight close to 1 near each other and vertices connected with an edge of weight closer to 0 or not connected at all further apart.
 
\item \label{Step:OptimizeEmbedding} \cite[Algorithm 5]{UMAP} The coordinates in $Y_0$ are then optimized further by stochastic gradient descent for $n\mathit{-epochs}$ iterations.

  For this, a new graph $G_{Y_0}$ is constructed. This graph has the vertex set $Y_0$ and the edges have weights that depend on the Euclidean distance of the endpoints in the embedding $Y_0$.
  We then compute one gradient descent step, which attempts to minimize dissimilarity between the weights of $G$ and $G_{Y_0}$ by varying randomly chosen coordinates of $Y_0$. This yields new coordinates $Y_1$ and a new graph $G_{Y_1}$.
  We then compute the next step, which attempts to minimize cross entropy between $G$ and $G_{Y_1}$ and so on.
  We stop this procedure after $n\mathit{-epochs}$ iterations. The final output of the algorithm is the dataset $Y := Y_{n\mathit{-epochs}}$.
\end{enumerate}

In theory, the spectral embedding in step 3 could be omitted by initializing step 4 with random coordinates $Y_0$. In essence, step 3 serves the purpose of generating good initial values for faster convergence and better numerical stability of the iterative approximation in step 4 \cite[Section~3.2]{UMAP}.

\subsection{Discussion} \label{Subsection:Discussion}

As mentioned in \cref{Subsection:UMAPAlgo}, UMAP seeks to preserve the topological structure of a Riemannian manifold $\mathcal{M}$ from which we assume the input is sampled. There are no formal theorems to justify this claim in the article \cite{UMAP} as stated in \cite[Section~7]{UMAP}. However, McInnes et al. informally justify the algorithm with the following claims about intermediate steps of the algorithm. We comment on these claims and relate to our work:

\begin{enumerate}
\item \label{Claim:Rescaling} McInnes et al. claim that when the input dataset $X = \{ x_1, \ldots, x_N \} \subseteq \mathbb{R}^D$ is assumed to be sampled from a Riemannian manifold $\mathcal{M}$ embedded in $\mathbb{R}^D$, the metric can be locally rescaled such that the data, even if not sampled uniformly from $\mathcal{M}$, appears uniformly sampled with respect to the locally rescaled metric \cite[Section~2.1]{UMAP}.

  Lemma~1 in \cite[Section~2.1]{UMAP} and the related \cite[Appendix~A]{UMAP} do not make a formal argument in terms of probability distributions over $\mathcal{M}$. Further work is necessary to verify this claim.
  
\item \label{Claim:Graph} Definition 9 in \cite{UMAP} defines the fuzzy topological representation of a dataset $X$ as $\bigcup_{0 \le i \le N} \text{FinSing}(X,d_i)$, where $d_i$ corresponds to the local metric around $X_i$ from claim \ref{Claim:Rescaling}.

  In \cite[Section~3.1]{UMAP}, McInnes et al. claim that, when considered as a fuzzy set, the local graphs $G_i$ constructed in \cref{Subsection:UMAPAlgo}, step 1 correspond to the 1-skeletons of $\text{FinSing}(X,d_i)$ from \cite[Definition 9]{UMAP} and that the union of the local graphs $G$ corresponds to the 1-skeleton of $$\bigcup_{0 \le i \le N} \text{FinSing}(X,d_i).$$

  Here the extended pseudo metrics $d_i : X \times X \rightarrow [0,\infty]$ are given by
  $$d_i(x_j,x_k) =
  \begin{cases}
    d_\mathcal{M}(x_j,x_k) - \rho& \text{if } $i = j$ \text{ or } $i = k$\\
    \infty & \text{else}
  \end{cases},
  $$
  where $\rho$ is the distance of $x_i$ to the nearest point and $d_\mathcal{M}$ is approximated as described in \cite[Section~2.1]{UMAP}.

  We have verified the construction of the finite singular nerve in \cref{Def:FinMetRe}. By expressing this functor in terms of classical fuzzy sets in \cref{Prop:CSing}, we established that the classical fuzzy set $\Fin{\CSing}(X,d_i)_1$ contains an element for each pair of points $x,y$ of finite distance with membership strength $\exp(-d_i(x,y))$, see \cref{Rem:OneSkeleton}. Thus, by \cref{Rem:Graph}, the local graphs $G_i$ correspond to the classical fuzzy set $\Fin{\CSing}(X,d_i)_1$, so this claim appears to be correct.

  Further work is necessary to extend the union of classical fuzzy sets to simplicial classical fuzzy sets and verify the correspondence between $G$ and $\bigcup_{0 \le i \le N} \Fin{\CSing}(X,d_i).$

\item In \cite[Section~3.1]{UMAP}, McInnes et al. claim ``Intuitively one can think of the weight of an edge as akin to the probability that the given edge exists. Section 2 demonstrates why this construction faithfully captures the topology of the data.''

  This claim is later used to theoretically justify the choice for the probabilistic $T$-conorm for the union in step \ref{Step:Union} \cite[Section~4.1]{UMAP}. It is also used to justify the choice of cross entropy in step \ref{Step:OptimizeEmbedding} \cite[Section~3.2]{UMAP}, as cross entropy is a measure of divergence of probability distributions.
  
  The article \cite{UMAP} does not provide a probability-theoretic argument for this claim. The article \cite{UMAP} also does not define a probability theoretic event that corresponds to an edge existing. We are thus not yet convinced that the weights of edges of the graph $G$ should be thought of as probabilities.

  Further work is necessary to formally capture this claim and verify it. If a justified interpretation as probabilities can be found, it should also be verified that this interpretation is compatible with claim \ref{Claim:Graph}.

\item McInnes et al. claim that if one accepts the claim that $G$ meaningfully captures the topological structure of $\mathcal{M}$, the iterative procedure from step $\ref{Step:OptimizeEmbedding}$ produces lower-dimensional output coordinates $Y$ that match the topology of $\mathcal{M}$ as close as possible. Further work is necessary to verify this claim.
  
\end{enumerate}

In this thesis, we verified the metric realizations of several variants of fuzzy sets and showed how they can be realized as left Kan extensions along Yoneda embeddings. Together with the explicit descriptions of these Kan extensions, this gives a precise description of the mathematical framework that describes UMAP according to \cite{UMAP}. However, this thesis does not and cannot establish if this framework explains the efficiency of UMAP or gives any additional insights into this algorithm.


\section*{Acknowledgement}

I would like to thank my advisor Prof. Dr. Meusburger for her excellent
guidance. Working with Prof. Dr. Meusburger did feel like a privilege.

I would also like to thank Prof. Dr. Creutzig for serving as second
examiner and investing the time to read my thesis.

\urlstyle{same}
\printbibliography

@misc{UMAP,
      title={UMAP: Uniform Manifold Approximation and Projection for Dimension Reduction}, 
      author={Leland McInnes and John Healy and James Melville},
      year={2020},
      eprint={1802.03426},
      archivePrefix={arXiv},
      primaryClass={stat.ML},
}

@electronic{Spivak2009METRICRO,
  title={METRIC REALIZATION OF FUZZY SIMPLICIAL SETS, unpublished draft},
  author={David I. Spivak},
  url={http://www.dspivak.net/metric_realization090922.pdf},
  urldate={2025-09-16}
}

@article{BarrFuzzy,
  title={Fuzzy Set Theory and Topos Theory},
  volume={29},
  number={4},
  journal={Canadian Mathematical Bulletin},
  author={Barr, Michael},
  year={1986},
  pages={501–508}
}

@book{riehl2017category,
  title={Category Theory in Context},
  author={Riehl, E.},
  series={Aurora: Dover Modern Math Originals},
  year={2017},
  publisher={Dover Publications}
}

@book{Riehl_2014,
  place={Cambridge},
  series={New Mathematical Monographs},
  title={Categorical Homotopy Theory},
  publisher={Cambridge University Press},
  author={Riehl, Emily}, year={2014},
  collection={New Mathematical Monographs}
}

@misc{barth,
      title={Fuzzy simplicial sets and their application to geometric data analysis}, 
      author={Lukas Silvester Barth and Fatemeh and Fahimi and Parvaneh Joharinad and Jürgen Jost and Janis Keck and Thomas Jan Mikhail},
      year={2024},
      eprint={2406.11154},
      archivePrefix={arXiv},
      primaryClass={math.AT},
}

@electronic{maclane1992sheaves,
  address = {New York, NY},
  author = {Mac Lane, Saunders and Moerdijk, Ieke},
  description = {Sheaves in Geometry and Logic - Springer},
  publisher = {Springer New York},
  refid = {853260961},
  title = {Sheaves in Geometry and Logic a First Introduction to Topos Theory},
  year = 1992
}

@book{Richter_2020,
  place={Cambridge},
  series={Cambridge Studies in Advanced Mathematics},
  title={From Categories to Homotopy Theory},
  publisher={Cambridge University Press},
  author={Richter, Birgit}, year={2020},
  collection={Cambridge Studies in Mathematics}
}

@article{vandermaaten2009dimensionality,
  author = {Van Der Maaten, Laurens and Postma, Eric and Van den Herik, Jaap},
  journal = {J Mach Learn Res},
  pages = {66-71},
  title = {Dimensionality reduction: a comparative review},
  volume = 10,
  year = 2009
}

@book{FuzzyLogic,
  title={Fuzzy Sets and Fuzzy Logic: Theory and Applications},
  publisher={Prentice Hall PTR},
  author={George J. Klir and Bo Yuan}, year={1995},
  collection={Cambridge Studies in Mathematics}
}
\addcontentsline{toc}{section}{Bibliography}
\printindex
\addcontentsline{toc}{section}{Index} \label{Sec:Index}
\newpage
\subsection*{Category Translation Table}

\begin{center}
  \resizebox{\textwidth}{!}{
    \begin{tabular}{ m{6.7cm} | m{5.4cm} | m{1.6cm} | m{1.7cm} | m{2.3cm} } 
      Full name in this Thesis & Notation in this Thesis&Notation in \cite{BarrFuzzy} &Notation in \cite{Spivak2009METRICRO} &Notation in \cite{UMAP} \\
            \hline
      sets                           & $\SET$             & - & Sets  & Sets\\
      topological spaces             & $\Top$             & - & -  & Top\\
      metric spaces                  & $\Met$             & - & -  & -  \\
      extended pseudo metric spaces  & $\EPMet$           & - & UM & EPMet \\
      \hline
      classical $\mathcal{L}$-valued sets & $\CVal{\mathcal{L}}$             &  $\text{Fuz}_0(\mathcal{L})$        & -  & -\\
      classical fuzzy sets            & $\CFuz := \CVal{[0,1], \le}$  &  - & - & -\\
      classical normed sets           & $\CNSet := \CVal{[0,\infty], \ge}$ &  - & - & -\\
      \hline
      sheaves on $\mathcal{L}$ & $\Sheaf{\mathcal{L}}$ & Sh($\mathcal{L}$) & Shv($\mathcal{L}$) & - \\
      \hline
      $\mathcal{L}$-valued sets & $\Val{\mathcal{L}}$                 & Mon($\mathcal{L}$) & - & - \\
      fuzzy sets      & $\Fuz := \Val{[0,1], \le}$  & -        & Fuz & Fuzz \\
      normed sets     & $\NSet := \Val{[0,\infty], \ge}$ & -       & - & - \\
      \hline
      Simplex category &$\Delta$ & - & $\Delta$ & $\Delta$\\
      \hline
      simplicial classical $\mathcal{L}$-valued sets & $\CVal{\mathcal{L}}^{\Delta^\op}$ & - & - & - \\
      simplicial classical fuzzy sets           & $\CFuz^{\Delta^\op}$          & - & sFuz & sFuzz \\
      simplicial classical normed sets          & $\CNSet^{\Delta^\op}$         & - & - & - \\
      \hline
      uncurried simplicial $\mathcal{L}$-Valued sets & $\SVal{\mathcal{L}}$                  & - & - & - \\
      uncurried simplicial fuzzy sets      & $\SFuz := \SVal{[0,1], \le}$  & - & sFuz & sFuzz \\
      uncurried simplicial normed sets     & $\SNSet := \SVal{[0,\infty], \ge}$ & - & - & - \\
      \hline
      finite extended pseudo-metric spaces  & $\Fin{\EPMet}$           & - & - & FinEPMet \\
      finite uncurried simplicial normed sets  & $\Fin{\SFuz}$           & - & - & Fin-sFuzz \\
    \end{tabular}}
\end{center}


\end{document}